\documentclass[10pt,twocolumn,letterpaper]{article}

\usepackage[pagenumbers]{cvpr} %
\usepackage[frozencache,cachedir=minted-cache]{minted}

\usepackage[accsupp]{axessibility}

\definecolor{cvprblue}{rgb}{0.21,0.49,0.74}
\usepackage{multirow}
\usepackage{algpseudocode}
\usepackage{algorithm}
\usepackage{tcolorbox}
\usepackage{amssymb}%
\usepackage{pifont}%
\newcommand{\cmark}{\ding{51}}%
\newcommand{\xmark}{\ding{55}}%
\usepackage[pagebackref,breaklinks,colorlinks,citecolor=cvprblue]{hyperref}

\def\paperID{5747} %
\def\confName{CVPR}
\def\confYear{2025}

\title{Reference-Based 3D-Aware Image Editing with Triplanes}

\author{Bahri Batuhan Bilecen$^1$ \quad Yigit Yalin$^1$ \quad Ning Yu$^2$ \quad Aysegul Dundar$^1$\\
Bilkent University$^1$ \quad Netflix Eyeline Studios$^2$\\
{\tt\footnotesize \{batuhan.bilecen@, yigit.yalin@ug., adundar@cs.\}bilkent.edu.tr, ning.yu@scanlinevfx.com}
}
\begin{document}
\twocolumn[{%
\renewcommand\twocolumn[1][]{#1}%
\maketitle
\begin{center}
    \centering
    \captionsetup{type=figure}
\vspace{-0.75cm}
    \includegraphics[width=0.99\textwidth]{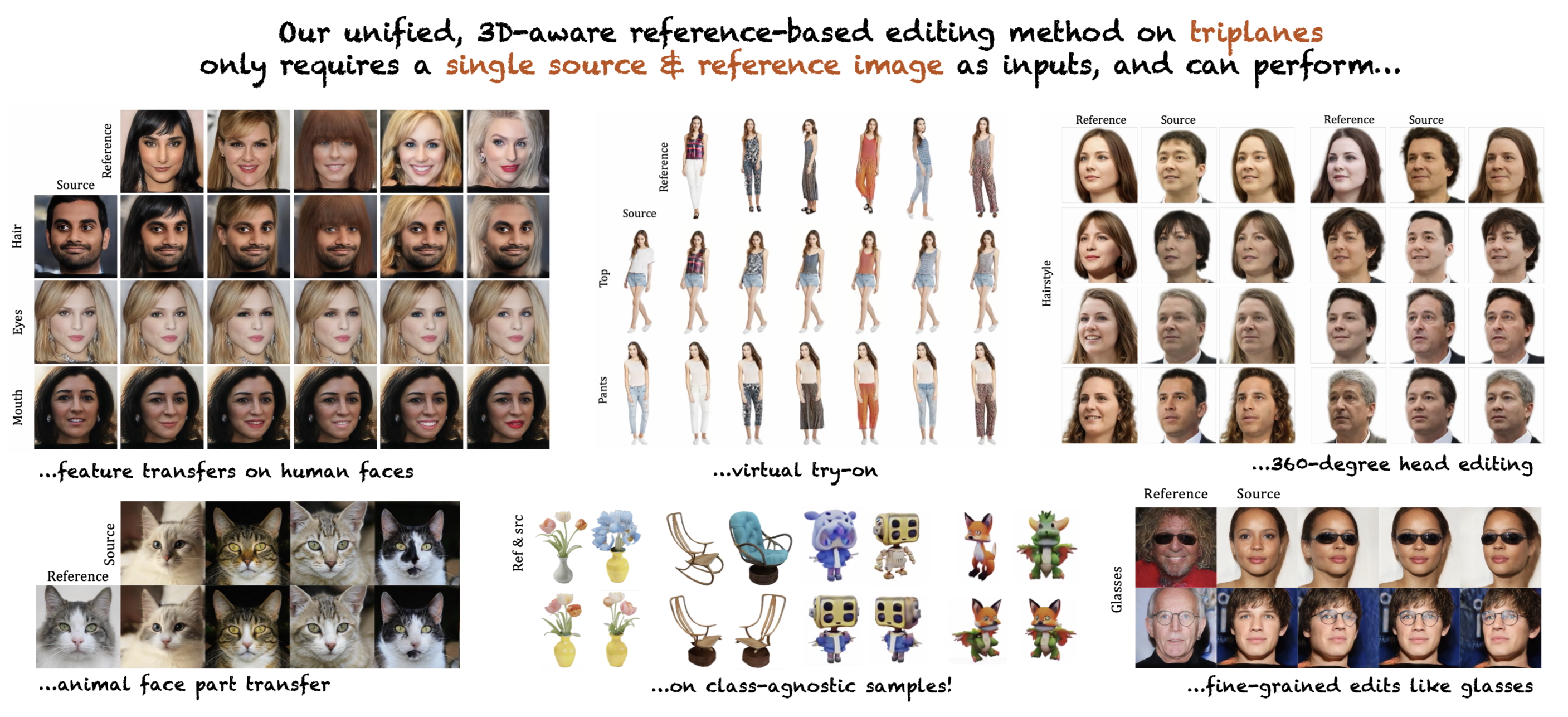}
    \captionof{figure}{Our approach excels in refer ence-based edits, faithfully reproducing the copied reference parts with a \textbf{single} source and reference image. Leveraging 3D-aware triplanes, our edits are versatile and 3D consistent, allowing for rendering from various viewpoints. We show results on human faces, heads, bodies, and extending beyond to animal faces and class-agnostic samples.}
    \label{fig:teaser}
\end{center}%
}]

\begin{abstract}
Generative Adversarial Networks (GANs) have emerged as powerful tools for high-quality image generation and real image editing by manipulating their latent spaces. Recent advancements in GANs include 3D-aware models such as EG3D, which feature efficient triplane-based architectures capable of reconstructing 3D geometry from single images. However, limited attention has been given to providing an integrated framework for 3D-aware, high-quality, reference-based image editing. This study addresses this gap by exploring and demonstrating the effectiveness of the triplane space for advanced reference-based edits. Our novel approach integrates encoding, automatic localization, spatial disentanglement of triplane features, and fusion learning to achieve the desired edits. We demonstrate how our approach excels across diverse domains, including human faces, 360-degree heads, animal faces, partially stylized edits like cartoon faces, full-body clothing edits, and edits on class-agnostic samples. Our method shows state-of-the-art performance over relevant latent direction, text, and image-guided 2D and 3D-aware diffusion and GAN methods, both qualitatively and quantitatively.\footnote{{\tt\small \href{https://three-bee.github.io/triplane_edit/}{three-bee.github.io/triplane\_edit}}}
\end{abstract}
\vspace{-0.5cm}

\section{Introduction}

\newcommand{\interpfigt}[1]{\includegraphics[trim=0 0 0cm 0, clip, width=2.3cm]{#1}}

In recent years, the high-fidelity image synthesis performance has been profoundly transformed by the emergence of Generative Adversarial Networks (GANs) \cite{goodfellow2014generative,karras2019style,karras2020analyzing}. Through adversarial training, they learn to map random distributions to actual data observations, enabling the generation of photo-realistic images from latent codes. The evolution of GANs from 2D into 3D-aware~\cite{chan2021pi,chan2022efficient,gu2021stylenerf,or2022stylesdf} has further boosted this capability, integrating hybrid 3D representations and Neural Radiance Fields (NeRF)~\cite{mildenhall2021nerf} into style-based generators, yielding unparalleled success in crafting highly realistic 3D portraits.

The application of GANs extends beyond generation, venturing into real image editing through GAN inversion~\cite{alaluf2021restyle,richardson2021encoding,tov2021designing,yuan2023make,yildirim2023diverse} and manipulation of the latent space~\cite{shen2020interpreting}. A critical challenge in reference-based image editing via this latent space is striking the optimal balance between retaining essential elements from the input image and incorporating desired attributes from the reference image. This balance is crucial to avoid overshadowing the input image's essence in the editing process, thereby maintaining its identity while still transferring the reference's attributes. This presents itself as a complex problem of disentanglement and fusion, requiring the identification and integration of relevant feature components from both the reference and input images within the latent space to produce the edited output.

\begin{figure}[t!]
    \centering
    \Large
    \scalebox{0.47}{
\setlength\tabcolsep{1pt}
    \begin{tabular}{cccccccc}
    & Src & Ref & \textbf{Ours} & \cite{barbershop} & \cite{paintbyexample} & \cite{infedit} & \cite{noiseclr} \\
    \rotatebox{90}{~~~Hair}& 
    \interpfigt{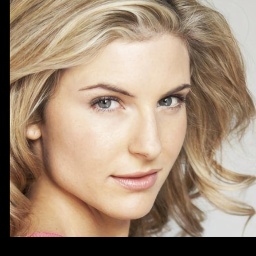}&
    \interpfigt{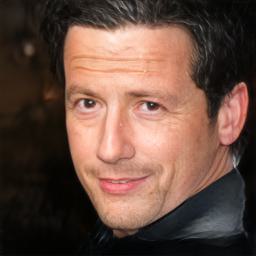}&
    \interpfigt{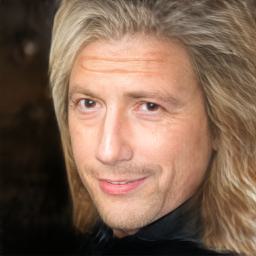}&
    \interpfigt{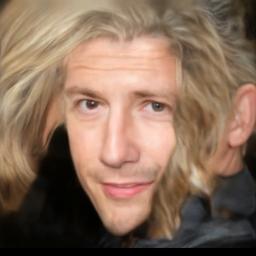}&
    \interpfigt{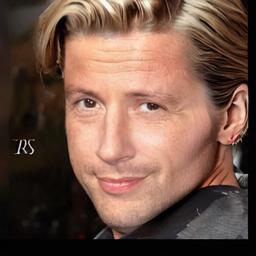}&
    \interpfigt{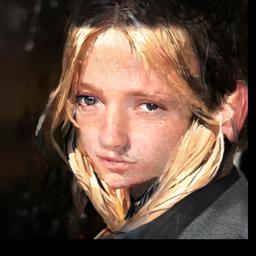}&
    \interpfigt{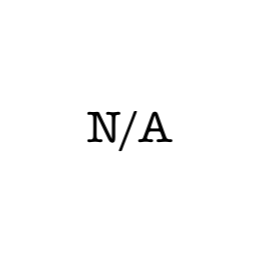}
    \\
    \rotatebox{90}{~~~Glasses}& 
    \interpfigt{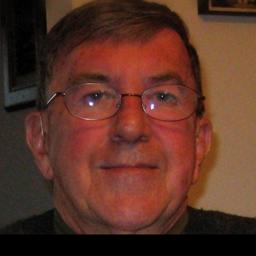}&
    \interpfigt{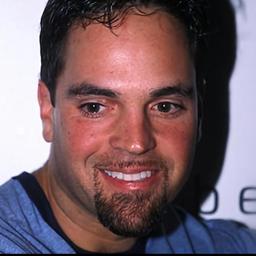}&
    \interpfigt{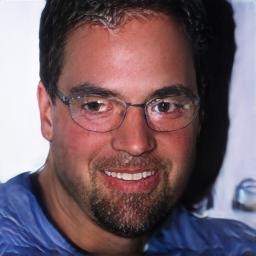}&
    \interpfigt{Figures/NA.png}&
    \interpfigt{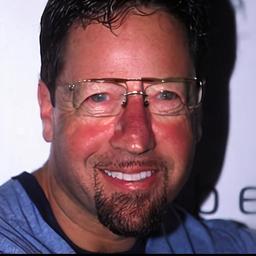}&
    \interpfigt{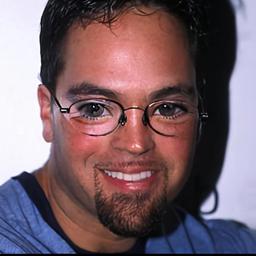}&
    \interpfigt{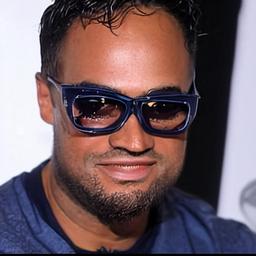}
    \end{tabular}
    }
    \caption{Current methods struggle with 3D consistency~\cite{barbershop,hairclipv2,paintbyexample}, faithfulness to the reference~\cite{kafri2021stylefusion,infedit,ledits++}, and visual artifacts~\cite{infedit, paintbyexample}. Our method provides 3D-consistent, reference-based edits from single images, independent of camera poses. N/A indicates the model is incapable of such edits.}
    \vspace{-0.5cm}
    \label{fig:first_comparison}
\end{figure}

Despite the advancements in image editing, there is a noticeable gap in the development of reference-based, 3D-aware image editing techniques. Existing methods in 3D-aware image editing lack reference-based capabilities~\cite{yuan2023make,lei2023diffusiongan3d,Fruehstueck2023VIVE3D}, while current GAN and diffusion reference-based editing approaches do not support 3D-awareness~\cite{barbershop,hairclipv2, paintbyexample} and/or facilitate local editing~\cite{shen2020interpreting,patashnik2021styleclip,Bobkov_2024_CVPR,noiseclr,ledits++} (\cref{fig:first_comparison,fig:full_comparison}). Addressing this gap, our work focuses on pioneering reference-based, 3D-aware image editing, where we learn spatial disentanglement and fusion within triplane latent spaces~\cite{chan2022efficient,dong2023ag3d}. 
Our primary motivation stems from discovering that triplanes can be manipulated for editing purposes akin to the 2D image domain but offer distinct advantages. Triplanes not only facilitate 3D editing but also alleviate alignment issues inherent in 2D image space. For example, transferring eyeglasses from one person to another in 2D necessitates precise alignment in the image space. Conversely, in the triplane space, stitching is facilitated as images with varying camera parameters can be projected onto the same canonical triplane space. However, ensuring seamless boundaries requires careful attention.
To accomplish reference-based image editing, our framework localizes parts within triplanes using masked residual gradients and fuses them using the encoders we train.

Our contributions span the following:
\begin{enumerate}
\item We are at the forefront of conceptualizing reference-based 3D-aware image editing as an integrated framework by leveraging the power of triplanes. Our approach includes encoding triplane features, spatial disentanglement with automatic localization of features, and fusion learning for desired image editing.
\item Our work establishes new benchmarks for quantitative and qualitative assessment in reference-based image editing and surpasses the 12 most recent and relevant baseline editing methods. This advancement is quantified by significant improvements in FID for quality and masked pixel-wise metrics for source preservation.
\item We extend our framework based on~\cite{chan2022efficient} towards 360-degree human heads~\cite{An_2023_CVPR}, animal faces~\cite{chan2022efficient}, cross-domain edits with cartoon portraits~\cite{song2022diffusion}, full-body~\cite{dong2023ag3d} clothing edits, and edits on class-agnostic samples~\cite{hong2024lrm,xu2024instantmesh,lan2024ln3diff}, showcasing its versatility and robustness across different triplane domains as shown in~\cref{fig:teaser}. This not only proves the effectiveness of our method but also broadens its potential in creative editing contexts.
\end{enumerate}
    
\section{Related Works}

\textbf{StyleGAN inversion-based editing.} It has been shown that well-trained GAN models organize their latent space in a semantically meaningful way that enables edits via latent vector arithmetic. 
Especially for StyleGAN models~\cite{karras2020analyzing}, many methods are proposed to find interpretable directions in unsupervised ways such as GANSpace~\cite{harkonen2020ganspace}, StyleFlow~\cite{abdal2021styleflow}, StyleSpace~\cite{wu2021stylespace}, StyleCLIP~\cite{patashnik2021styleclip} and supervised ways such as InterFaceGAN~\cite{shen2020interpreting}, and recently E3DGE~\cite{Lan_2023_CVPR}, Barbershop~\cite{barbershop}, HairCLIPv2~\cite{hairclipv2}, and SFE~\cite{Bobkov_2024_CVPR}. 
These methods are combined with real image inversion methods so that an image is projected onto StyleGAN's latent space ($\mathcal{W}, \mathcal{W^+}, \mathcal{F}$, etc.) and is edited~\cite{richardson2021encoding, pehlivan2023styleres}. 
Recently, EG3D~\cite{chan2022efficient} augmented StyleGAN architecture with triplanes that provide efficient 3D-aware representations.
For this new domain, new inversion methods are proposed~\cite{xie2023high, yin20233d, yuan2023make, Fruehstueck2023VIVE3D}; however, previous editing methods~\cite{shen2020interpreting, patashnik2021styleclip} are re-used from 2D StyleGAN literature, which all lack reference-based editing.

\noindent \textbf{Reference-based image-to-image translation.} We refer to the models that are end-to-end trained for editing applications as image-to-image translation methods. 
These methods are trained to change selected attributes of images while preserving the content~\cite{dundar2020panoptic, liu2022partial, starganv2,shen2017learning,xiao2018elegant,zhang2018generative,li2021image,hou2022guidedstyle, lyu2021sogan}, rather than using the latent codes of StyleGAN.
Specifically, they include an encoder for the reference image and another encoder or shared one for the source image~\cite{zhu2017multimodal, li2021image, zhu2020sean, huang2018multimodal,starganv2, yang2021l2m, lyu2023dran}. %
In these models, the reference style can be sampled from a normal distribution, or it can be encoded from a reference image~\cite{li2021image, dalva2022vecgan, dalva2023image}. 
One major limitation of these works is that they require labeled datasets for each attribute~\cite{liu2015deep}; hence, models like VecGAN~\cite{dalva2023image} and HisD \cite{li2021image} can only achieve a handful of edits (hair color, smile, eyeglasses, bangs, etc.). %
They also cannot provide editing using images taken from vastly different camera poses and lack faithfulness to the reference.

\noindent\textbf{Text and image-conditioned diffusion models for image editing.}
There is an abundance of diffusion-based image editing methods~\cite{ledits++, infedit, paintbyexample, noiseclr, ObjectStitch, ding2023diffusionrig}, leveraging the prior of Stable Diffusion (SD) denoiser~\cite{LDM} and achieving diverse edits. Recently, methods like ControlNet~\cite{ControlNet}, IP~\cite{ye2023ip-adapter}, and T2I~\cite{t2i_adapter} have made editing more controllable. Still, all these SD-based methods lack 3D-aware, faithful-to-reference image editing, especially in the face and body domain. LEDITS++~\cite{ledits++} and InfEdit~\cite{infedit} utilize text prompts and are incapable of image reference-based editing. Paint by Example~\cite{paintbyexample} fine-tunes the denoiser with masked images to carry information from reference images. Nevertheless,~\cite{paintbyexample} is not capable of choosing which region to copy, harming the controllability - more noticeably when camera poses are different. NoiseCLR~\cite{noiseclr} attempts to find editing directions unsupervised but fails for fundamental face-edit attributes like hair color and style changes. There are 3D-aware denoisers~\cite{Zero1to3,liu2024syncdreamer,DreamComposer} and edit methods~\cite{pandey2024diffusionhandles, erkoc2024preditor3d, zhang2025ditscene}, but they also become impractical in our context, as they are either fine-tuned with general object datasets~\cite{objaverse,objaverseXL} failing to perform well in the face domain or lack faithful reference-based editing altogether. There are also methods fusing GAN \& diffusion for adding stylization and diversity~\cite{song2022diffusion,DiffPortrait3D,lei2023diffusiongan3d,bai20243dpe}, but controllable reference-based edits are not achieved in those, either.

\section{Method}
Our main motivation lies in that triplanes can be stitched and blended for editing like in the 2D image domain. 
However, achieving satisfying results requires carefully designed steps, which will be detailed in this section.

\subsection{Localizing parts in the triplane space} 
\label{sec:localizingtriplane}

\begin{figure}[ht!]
    \centering
    \includegraphics[width=.9\linewidth]{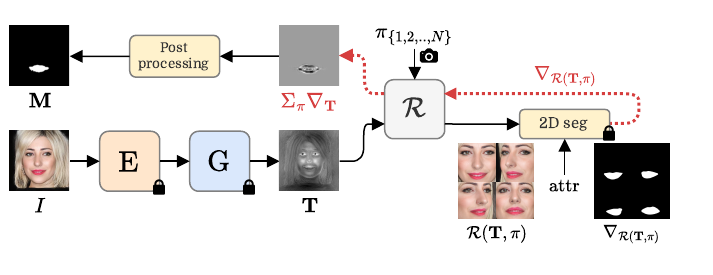}
    \caption{Triplane part localization stage, where $\mathbf{E}$, $\mathbf{G}$, and $\mathcal{R}$ are encoder~\cite{yuan2023make,bhattarai2024triplanenet}, generator~\cite{chan2022efficient}, and neural volumetric renderer, respectively. For the 2D segmentation model $\mathcal{S}_\text{2D}$, we use state-of-the-art off-the-shelf segmentation models~\cite{faceparsingpytorch}. Images other than the input image are zoomed in for visualization purposes.}
    \label{fig:mask-extract}
\end{figure}

The most direct way to perform a region transfer from reference to source image would be to mask and copy \& paste the region of interest. However, this process becomes intricate in hybrid 3D representations like triplanes, due to the absence of a conventional method for identifying the regions to be masked.

\definecolor{mygray}{RGB}{30, 30, 30}
\definecolor{myred}{RGB}{255, 179, 156}
\definecolor{myyellow}{RGB}{255, 228, 161}

\begin{algorithm}
\footnotesize
\caption{Triplane localization and masking}
\begin{algorithmic}[1]
\Require Generator \textbf{G}, encoder \textbf{E}, renderer $\mathcal{R}$, image $I$, extrinsic matrices $\pi_{1,2,\ldots,N}$, segmentation net $\mathcal{S}_\text{2D}$, post-processing params ($\epsilon, \gamma$)
\Ensure Triplane mask $\textbf{M}$

\begin{tcolorbox}[colback=myred!40!white, colframe=white, arc=3pt, outer arc=3pt, boxrule=0pt, parbox=false]
\State $\mathbf{T}, \nabla_{\textbf{T}}, \leftarrow \mathbf{G}(\mathbf{E}(I)), 0$ \Comment{\textit{Initialize triplane and gradients}}
\For{each $\pi_i$ in $\pi_{1,2,\ldots,N}$} \Comment{\textit{Triplane localization}}
    \State $I_{\pi_i} \leftarrow \mathcal{R}(\textbf{T},\pi_i)$
    \State $\nabla_{\mathcal{R}(\textbf{T},\pi_i)} \leftarrow \mathcal{S}_{\text{2D}}(I_{\pi_i}, \mathbf{attr})$
    \State  $\nabla_{\textbf{T}} \leftarrow \nabla_{\textbf{T}} + \mathbf{autograd}(\mathcal{R}, \nabla_{\mathcal{R}(\textbf{T},\pi_i)})$ 
\EndFor
\end{tcolorbox}

\begin{tcolorbox}[colback=myyellow!40!white, colframe=white, arc=3pt, outer arc=3pt, boxrule=0pt, parbox=false]
\For{each $c$ in $\nabla_{\mathbf{T}}$}\Comment{\textit{Post-processing}}
    \State $\nabla_{\mathbf{T}_c} \leftarrow \{ x \in \nabla_{\mathbf{T}_c} : |\nabla_{\mathbf{T}_c}(x) - \mu_c| \leq \epsilon \}$
    \State $\nabla_{\mathbf{T}_c} \leftarrow (\nabla_{\mathbf{T}_c} - \min{\nabla_{\mathbf{T}_c}})/(\max\nabla_{\mathbf{T}_c} - \min\nabla_{\mathbf{T}_c})$
    \State $\nabla_{\mathbf{T}_c} \leftarrow \{ x \in \nabla_{\mathbf{T}_c} : 1 \ \mathbf{if} \ \nabla_{\mathbf{T}_c}(x) \geq \gamma \ \mathbf{else} \ 0 \}$ 
\EndFor
\end{tcolorbox}

\State \textbf{return}  $\mathbf{M} \leftarrow \nabla_{\mathbf{T}}$
\end{algorithmic}
\label{alg:alg1}
\end{algorithm}

To address this, we take advantage of the volumetric renderer used in~\cite{chan2022efficient} being fully differentiable, and back-propagate the 2D image domain masks to the 3D hybrid triplane domain to calculate gradients on triplanes (\cref{fig:mask-extract}, \textit{triplane localization} in~\cref{alg:alg1}). 
First, input images are encoded using a pre-trained model \cite{bhattarai2024triplanenet, yuan2023make} to obtain triplane features \textbf{T}. These features are rendered with different camera poses $\pi_{1,2,\ldots,N}$ to create multi-view 2D renderings $\mathcal{R}(\textbf{T},\pi)$. An off-the-shelf segmentation network $\mathcal{S}_\text{2D}$ identifies attributes~\cite{faceparsingpytorch} (e.g., hair, eyes, glasses) in each rendering. The segmentation outputs are assigned as output gradients $\nabla_{\mathcal{R}(\textbf{T},\pi)}$, and are back-propagated to the triplane domain to accumulate input gradients $\Sigma_{\pi}{\nabla_{\textbf{T}}}$, which localizes the triplane mask.

To convert a gradient mask into a binarized one, we perform mean clipping, normalization, and thresholding with parameters $(\epsilon, \gamma)$ (\textit{post-processing} in~\cref{alg:alg1}). These parameters are set once and used across all experiments for each attribute and domain, and are in the Supplementary. This localization is done for source and reference images.

For finer granularity and to avoid copying unwanted attributes, such as copying glasses and not the eyes, $\mathcal{W}^+$ directions~\cite{shen2020interpreting} can be utilized in addition. By computing the difference between triplanes with and without the attribute, $\Delta \textbf{T}_{\text{attr}} = \mathbf{G}(w) - \mathbf{G}(w - w_{\text{attr}})$ and multiplying with gradient mask \textbf{M}, a more precise mask can be created.

\subsection{Implicit fusion by encoding \& decoding} 
\label{sec:implicit}
After finding suitable masks for reference and source triplanes, $\textbf{M}_{\text{ref}}$ and $\textbf{M}_{\text{src}}$ respectively, a naive approach would be to follow~\cref{eqn:manual_fusion}:
\begin{equation}
    \textbf{T}_{\text{tmp}} = \textbf{M}_{\text{ref}}*\textbf{T}_{\text{ref}} + \textbf{M}_{\text{src}}*\textbf{T}_{\text{src}}
    \label{eqn:manual_fusion}
\end{equation}
However,~\cref{fig:development} (V1) reveals only using~\cref{eqn:manual_fusion} distorts the geometry and color consistency, and creates stitching seams around the editing borders. In addition, in some cases, the contents in $\textbf{T}_{\text{ref}}$ and $\textbf{T}_{\text{src}}$ are not enough to ensure the editing looks natural around the region where two masks meet, necessitating the hallucination of additional content to complement the editing. 
For example, we may want to remove the long hair from the source image and replace it with the short hair from the reference image. 
In this case, the pixels that correspond to long hair regions need to be inpainted.

\begin{figure*}[ht!]
    \centering
    \includegraphics[width=0.7\linewidth]{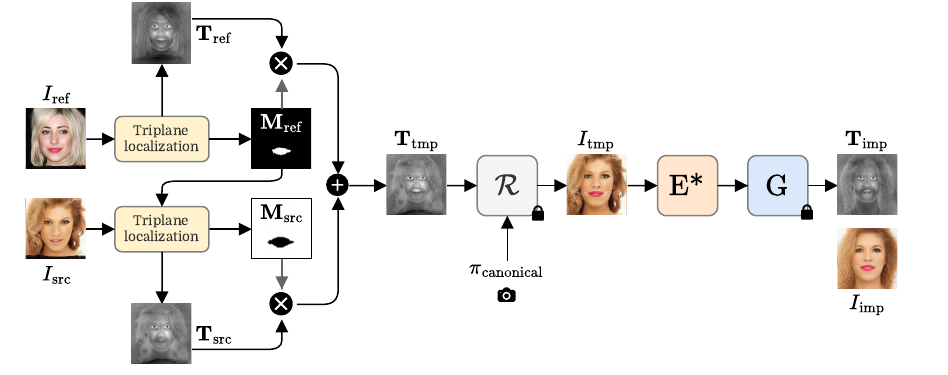}
    \caption{Triplane localization and implicit fusion stages, where $\mathbf{E}^*$ denotes the fine-tuned image encoder that is described in~\cref{sec:fine-tune}. 
    Straightforward stitching in the triplane results in color inconsistency across the boundaries, as shown in $I_{tmp}$ (zoom in for details).
    Leveraging $\mathbf{E}^*$, we aim to attain seamless boundaries and produce outputs with a natural appearance.}
    \label{fig:fusion}
\end{figure*}

Given the above observations and GAN's latent spaces embed natural images, we render the naively fused triplane with the canonical pose $\pi_\text{can}$, re-encode $\mathbf{E}_{\mathcal{W}^+}$, and re-decode via the generator $\mathbf{G}$ to obtain an implicitly fused triplane, shown in~\cref{eqn:implicit_fusion}:

\begin{equation}
   \textbf{T}_{\text{imp}} = \mathbf{G}(\mathbf{E}_{\mathcal{W}^+}(\mathcal{R}(\textbf{T}_{\text{tmp}},\pi_\text{can})))
   \label{eqn:implicit_fusion}
\end{equation}

Note that state-of-the-art image inversion methods for EG3D~\cite{yuan2023make, bhattarai2024triplanenet} employ both low-rate $\mathcal{W}^+$ and high-rate $\mathcal{F}$ features. The latter is crucial for reconstructing fine image details. \textit{However, our objective in this phase is not to achieve perfect image reconstruction. On the contrary, we aim for the encoder to map the image to a latent space with natural reconstructions.} To achieve this, we disable the high-frequency restoration branch of $\mathbf{E}_{\mathcal{W}^+}$ to prevent encoding visible seams. This allows us to project the edited image onto its nearest representation on $\mathbf{G}$'s manifold, implicitly fusing the masked triplanes. 
\cref{fig:fusion} showcases seamless boundaries across the stitches after this operation.

Note that image details are compromised at this stage, as we solely depend on the low-rate $\mathcal{W}^+$ space. To bring the details back, we only employ $\textbf{T}_\text{imp}$ at the transition regions, as depicted in~\cref{eqn:final_fusion}.
For instance, when we transfer the mouth from reference to source, we aim for the triplane features outside the mouth to originate from $\textbf{T}_\text{src}$, the mouth features from $\textbf{T}_\text{ref}$, and features near mouth from $\textbf{T}_\text{imp}$.

\begin{equation}
\begin{split}
   \textbf{T}_\text{f} &= \mathcal{E}( \textbf{M}_\text{ref})*\textbf{T}_\text{ref} + \mathcal{E}(\textbf{M}_\text{src})*\textbf{T}_\text{src} \\ &+ (\mathcal{E}(\textbf{M}_\text{src})-\mathcal{E}( \textbf{M}_\text{ref}))*\textbf{T}_\text{imp}
\end{split}
   \label{eqn:final_fusion}
\end{equation}

Here, $\mathcal{E}$ denotes morphological erosion. We also apply Gaussian blurring with parameters $(\mu, \sigma)$ onto the masks to avoid sharp edges. This step is not illustrated in~\cref{fig:fusion} for brevity, but $(\mu, \sigma)$ are provided in the Supplementary.

Finally, $\textbf{T}_\text{f}$ is rendered from any desired pose $\pi_R$, and the final image is obtained, as illustrated in~\cref{eqn:final_rendering}.
\begin{equation}
   I_\text{edited} = \mathcal{R}(\textbf{T}_\text{f},\pi_R)
   \label{eqn:final_rendering}
\end{equation}

\subsection{Fine-tuning the image encoder} 
\label{sec:fine-tune}
Although $\textbf{T}_\text{imp}$ obtained in~\cref{sec:implicit} helps tremendously even though a pre-trained encoder is used to obtain it,  we notice in some cases where we have skin color inconsistencies, background leakages, and missing high-frequency details around the editing regions (\cref{fig:development} (V2)).
Hence, we fine-tune implicit fusion encoder $\mathbf{E}_{\mathcal{W}^+}$, jointly with the triplane editing pipeline to mitigate the aforementioned effects.

During the fine-tuning phase, we generate renders as ground-truths from various viewpoints of the source and reference images corresponding to different attributes. Then, we employ masked losses to guide $\mathbf{E}_{\mathcal{W}^+}$ in encoding only the visible regions, illustrated in~\cref{fig:training}. For instance, if our objective is to transfer hair, we mask the reference renderings to exclude pixels that do not represent hair while doing the opposite for the source ground truth. To ensure that losses do not affect the boundaries, we dilate the segmentation masks. The objective function is provided in~\cref{eqn:objective}:

\begin{figure*}[t]
    \centering
    \includegraphics[width=0.7\linewidth]{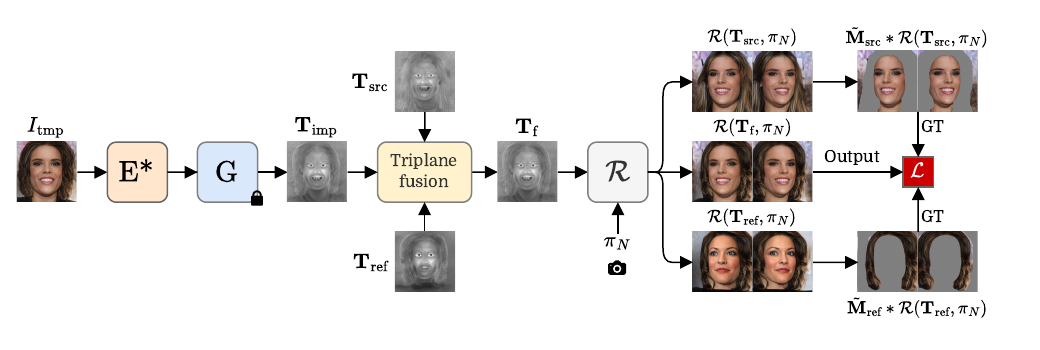}
    \caption{Pipeline for the implicit fusion encoder fine-tuning. We generate masked ground truths for our task by utilizing 2D segmentation networks and via renderings from multiple views. 
    We aim to carry the reference parts in great detail to our source image while preserving the source's identity. Triplane fusion corresponds to~\cref{eqn:final_fusion}.}
    \vspace{-0.2cm}
    \label{fig:training}
\end{figure*}

\begin{equation}
\begin{split}
\min_{\mathbf{E}} \lambda_{\Phi} \mathcal{L}_{\Phi}(\mathcal{D}(\Tilde{\textbf{M}}_\text{ref}) * \mathcal{R}(\textbf{T}_\text{f}, \pi_i), \mathcal{D}(\Tilde{\textbf{M}}_\text{ref}) * \mathcal{R}(\textbf{T}_\text{ref}, \pi_i)) \\+ \lambda_{\text{ID}} \mathcal{L}_{\text{ID}}(\mathcal{D}(\Tilde{\textbf{M}}_\text{src}) * \mathcal{R}(\textbf{T}_\text{f}, \pi_i), \mathcal{D}( \Tilde{\textbf{M}}_\text{src}) * \mathcal{R}(\textbf{T}_\text{src}, \pi_i))
\end{split}
\label{eqn:objective}
\end{equation}

where $\mathcal{L}_{\Phi}$ is the learned perceptual image patch similarity loss (LPIPS)~\cite{zhang2018lpips}, $\mathcal{L}_{\text{ID}}$ is the identity similarity loss~\cite{deng2019arcface}, $\mathcal{D}$ is dilation operation, and $\Tilde{\textbf{M}}_\text{src}$ and $\Tilde{\textbf{M}}_\text{ref}$ are the corresponding 2D segmentation masks for the rendered images $\mathcal{R}(\textbf{T}_\text{src}, \pi_i)$ and $\mathcal{R}(\textbf{T}_\text{ref}, \pi_i)$ with poses $\pi_i$, respectively. 

To achieve identity preservation of the source image, we rely on the ID losses, and to copy the attribute with details, we rely on the LPIPS score from~\cref{eqn:objective}. We omit pixel-wise losses like $\mathcal{L}_{\text{2}}$ since they are highly dependent on the quality of the off-the-shelf 2D segmentation network. During the fine-tuning, we use each subject tuple $(\textbf{T}_\text{src},\textbf{T}_\text{ref})$ multiple times, rendered from $N$ randomly chosen $\pi_i$'s.

Compliant with the common encoder training methodology, we use the same training datasets generators are trained with. We employ Ranger optimizer, which is a combination of Rectified Adam~\cite{liu2019radam} with Lookahead~\cite{zhang2019lookahead}. The learning rate is set to $1e^{-4}$, and fine-tuning is done for $1500$ steps with a batch size of 2 on a single RTX 4090.

\section{Experiments}

\textbf{Evaluation.}
We present metrics evaluating both the reconstruction and editing qualities of our approach. For editing assessment, we employ the Fréchet Inception Distance (FID)~\cite{heusel2017gans}, which evaluates realism by comparing the distribution of target images with that of edited images. 
Specifically, we compute FIDs for adding eyeglasses and hair edits from black to blonde transition using the CelebA~\cite{liu2015deep} dataset. 
For instance, leveraging ground-truth attribute labels, we add eyeglasses to images without them and compute FIDs between the edited and original images that already have eyeglasses. This procedure is similarly applied to hair edits. For reconstruction evaluation, we mask the edited areas and measure the L2 and Structural Similarity Index (SSIM) between the input and edited images. For instance, in the eyeglasses edit, we mask the eyeglasses and measure the alteration in the unedited regions. This dual assessment framework ensures a robust evaluation of both the fidelity and quality of our editing approach.

\noindent\textbf{Baselines.}
We conducted comprehensive comparisons by evaluating our method against a range of latent direction, text and image reference-based, 2D and 3D-aware, GAN, and diffusion-based editing methods. Notably, no existing reference-based editing methods achieve 3D consistency within a single framework.

HiSD and VecGAN++ are reference-based 2D image-to-image methods. InterFaceGAN, StyleCLIP ($\mathcal{W}^+$), SFE ($\mathcal{W}^+/\mathcal{F}$), StyleFusion ($\mathcal{W}$), Barbershop, and HairCLIPv2 ($\mathcal{F}/\mathcal{S}$) operate within the latent spaces of StyleGAN and can be adapted to 3D-aware GANs like EG3D. However, only the last three are reference-based, with two focusing on hair edits. E3DGE is a SDF-based generator, and editing is done via latent directions. For diffusion-based models, LEDITS++ and InfEdit perform text-based editing, while NoiseCLR utilizes pretrained latent edit noise directions. Paint by Example inpaints the masked source image with the reference image but lacks control over which parts of the reference are used.

\begin{table}[t!]
\caption{Quantitative scores on CelebA. (\xmark) indicates the method is not capable of such edits. \textbf{First} and \underline{second best} method are given in \textbf{bold} and \underline{underlined}. Time is measured on Tesla T4.}
\centering
\scriptsize
\setlength\tabcolsep{1pt}
\begin{tabular}{r|ccc|ccc|c}
\toprule
{} & \multicolumn{3}{c|}{\textbf{Eyeglasses}} & \multicolumn{3}{c}{\textbf{Hair}} & \multicolumn{1}{|c}{\textbf{Time}} \\ 
&  {FID $\downarrow$} & $\mathcal{M}_{\text{SSIM}} \uparrow$ & $\mathcal{M}_{\mathcal{L}_2} \downarrow$ & {FID $\downarrow$} & $\mathcal{M}_{\text{SSIM}} \uparrow$ & $\mathcal{M}_{\mathcal{L}_2} \downarrow$ & (s) \\ 
\hline
 HiSD~\cite{li2021image} & 77.56 & 0.9471 & 0.0090 & 94.53 & \textbf{0.9743} & 0.0036 & \underline{1.1} \\ 
VecGAN++~\cite{dalva2022vecgan} & \underline{71.47} & 0.7483 & 0.0630 & 80.47 & 0.9296 & 0.0090 & 2.2  \\ 
Barbershop~\cite{barbershop} & \xmark & \xmark & \xmark & \textbf{62.80} & 0.8756 & 0.0182 & 125 \\
HairCLIPv2~\cite{hairclipv2} & \xmark & \xmark & \xmark & 85.75 & 0.8769 & 0.0173 & 180\\
{StyleFusion}~\cite{kafri2021stylefusion} & \xmark & \xmark & \xmark & 84.67 & 0.8435 & 0.0198 & 2.4 \\
\hline
{InterfaceGAN}~\cite{shen2020interpreting} & 88.13 & 0.9398 & 0.0104 & 80.93 &
0.7888 & 0.0387 & \textbf{0.6} \\ 
{StyleCLIP}~\cite{patashnik2021styleclip} & 80.13 & 0.8421 & 0.0476 & 92.60 & 0.8716 & 0.0196 & \textbf{0.6}\\ 
{SFE}~\cite{Bobkov_2024_CVPR} & 106.1 & 0.9341 & 0.0099 & 89.49 & 0.9355 & 0.0050 & 5.1\\
{E3DGE}~\cite{Lan_2023_CVPR} & \xmark & \xmark & \xmark & 77.86 & 0.8083 & 0.0257 & 1.2\\
\hline
NoiseCLR~\cite{noiseclr} & 107.1 & 0.7958 & 0.0440 &  \xmark & \xmark & \xmark & 17.2\\
LEDITS++~\cite{ledits++} & 115.2 & \underline{0.9645} & \underline{0.0047} & 96.56 & 0.9717 & 0.0025 & 25.9 \\
InfEdit~\cite{infedit} & 90.33 & 0.8338 & 0.1042 & 105.4 & 0.7425 & 0.0613 & 9.1\\
Paint by ex.~\cite{paintbyexample} & 74.18 & 0.8828 & 0.0252 & 82.38 & 0.9229 & 0.0155 & 9.6\\
\hline
\textbf{Ours} & \textbf{66.68} & \textbf{0.9818} & \textbf{0.0021} & \underline{64.59} & \underline{0.9720} & \textbf{0.0029} & 6.0 \\ 
\bottomrule
\end{tabular}
\label{table:full_comparison}
\vspace{-0.5cm}
\end{table}

\begin{figure*}[t]
\centering
\scalebox{0.47}{
\setlength\tabcolsep{0.5pt}
\begin{tabular}{cccccccccccccccc}

\interpfigt{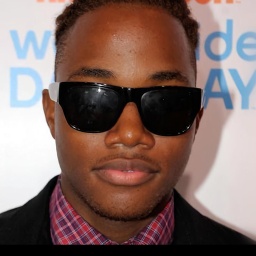}&
\interpfigt{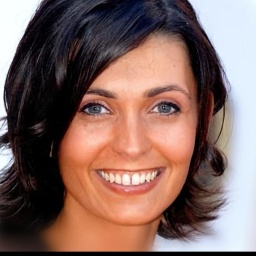}&
\interpfigt{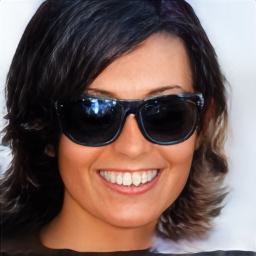}&
\interpfigt{Figures/NA.png}&
\interpfigt{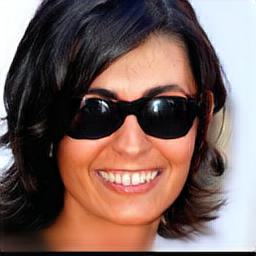}&
\interpfigt{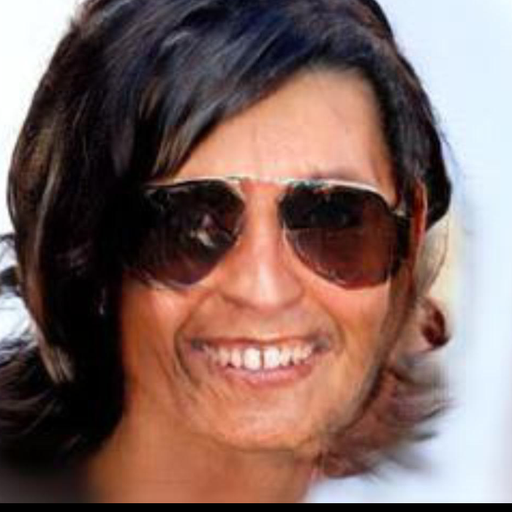}&
\interpfigt{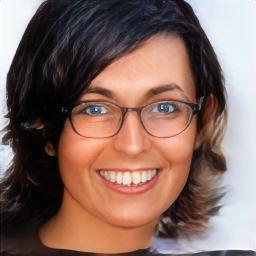}&
\interpfigt{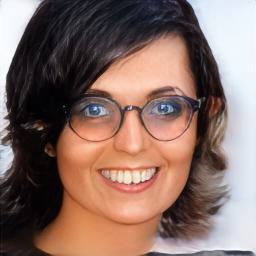}&
\interpfigt{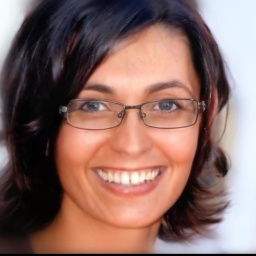}&
\interpfigt{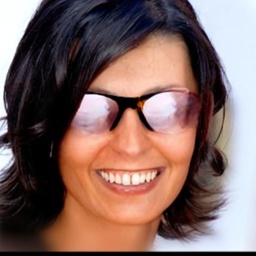}&
\interpfigt{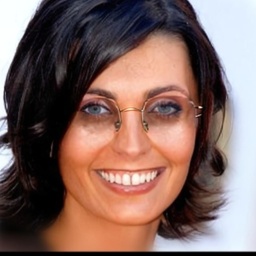}&
\interpfigt{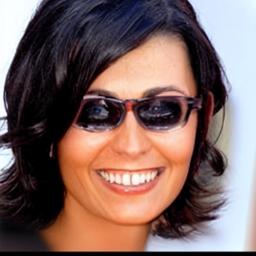}&
\interpfigt{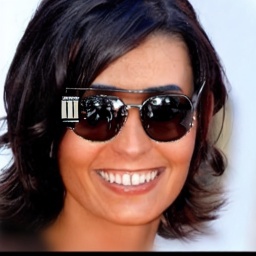}&
\interpfigt{Figures/NA.png}&
\interpfigt{Figures/NA.png}&
\interpfigt{Figures/NA.png}
\\

\interpfigt{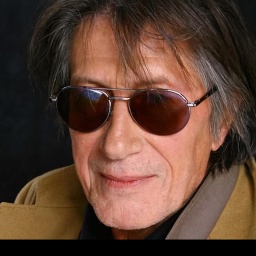}&
\interpfigt{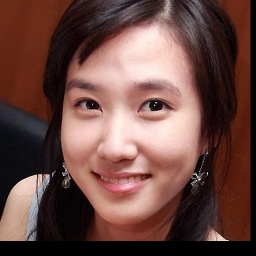}&
\interpfigt{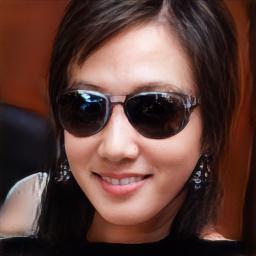}&
\interpfigt{Figures/NA.png}&
\interpfigt{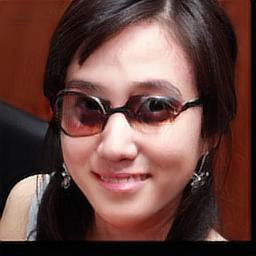}&
\interpfigt{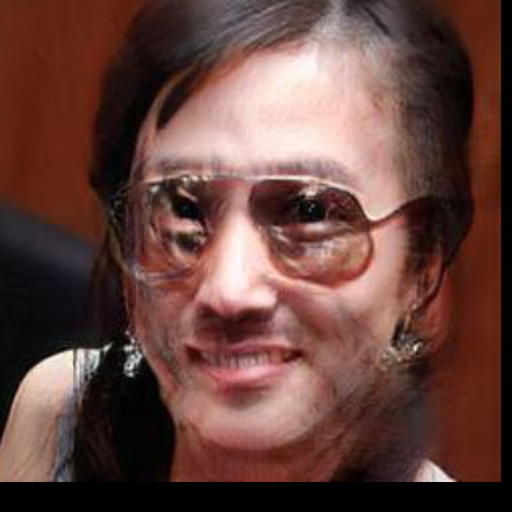}&
\interpfigt{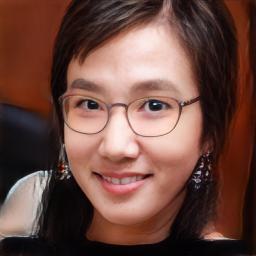}&
\interpfigt{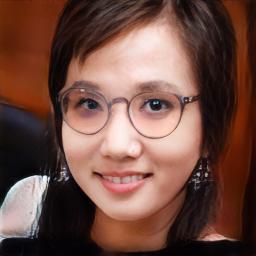}&
\interpfigt{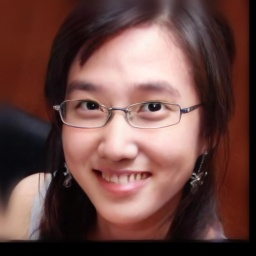}&
\interpfigt{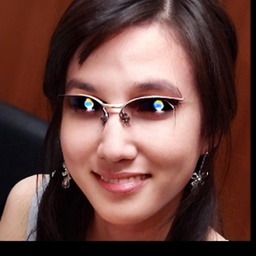}&
\interpfigt{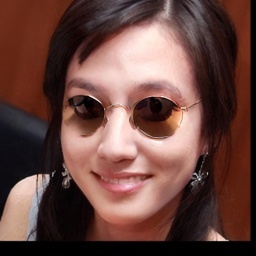}&
\interpfigt{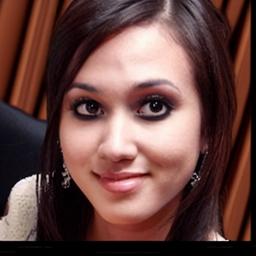}&
\interpfigt{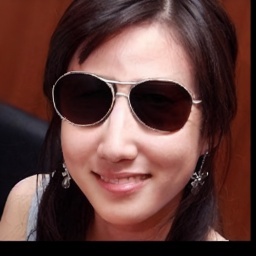}&
\interpfigt{Figures/NA.png}&
\interpfigt{Figures/NA.png}&
\interpfigt{Figures/NA.png}
\\

\interpfigt{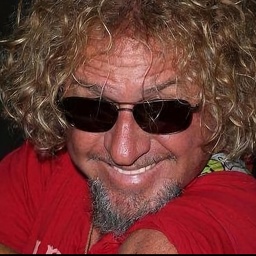}&
\interpfigt{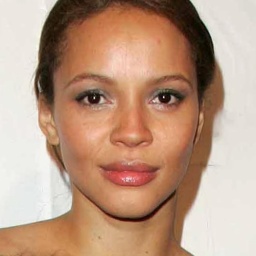}&
\interpfigt{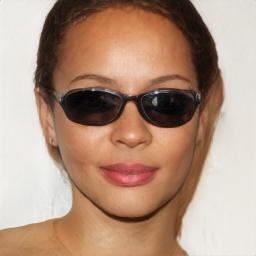}&
\interpfigt{Figures/NA.png}&
\interpfigt{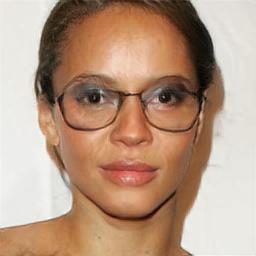}&
\interpfigt{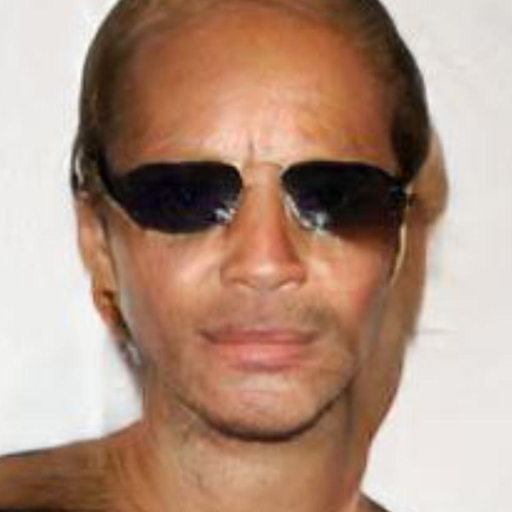}&
\interpfigt{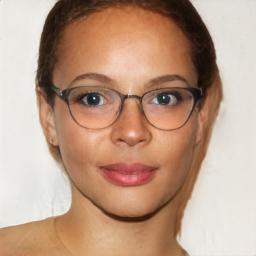}&
\interpfigt{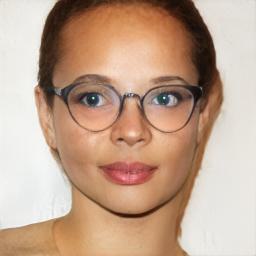}&
\interpfigt{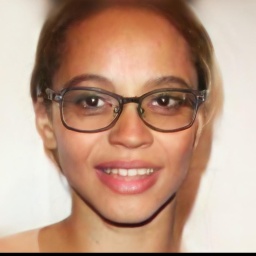}&
\interpfigt{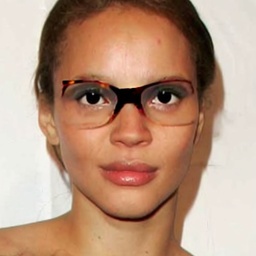}&
\interpfigt{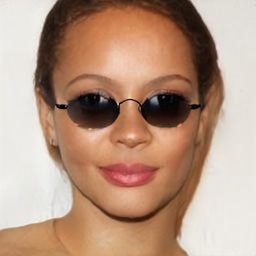}&
\interpfigt{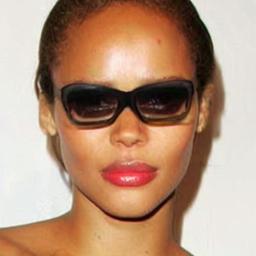}&
\interpfigt{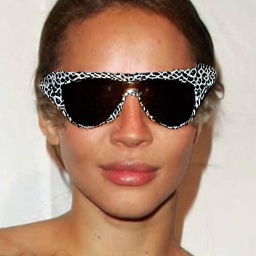}&
\interpfigt{Figures/NA.png}&
\interpfigt{Figures/NA.png}&
\interpfigt{Figures/NA.png}
\\

\interpfigt{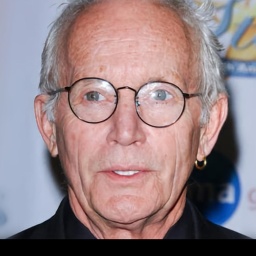}&
\interpfigt{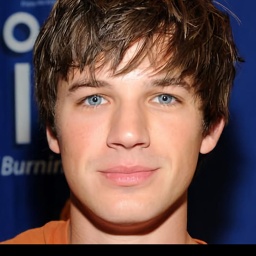}&
\interpfigt{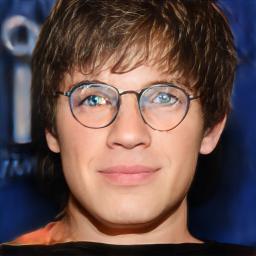}&
\interpfigt{Figures/NA.png}&
\interpfigt{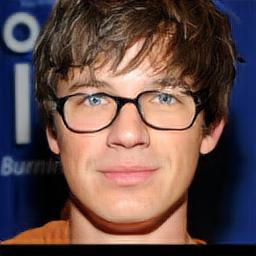}&
\interpfigt{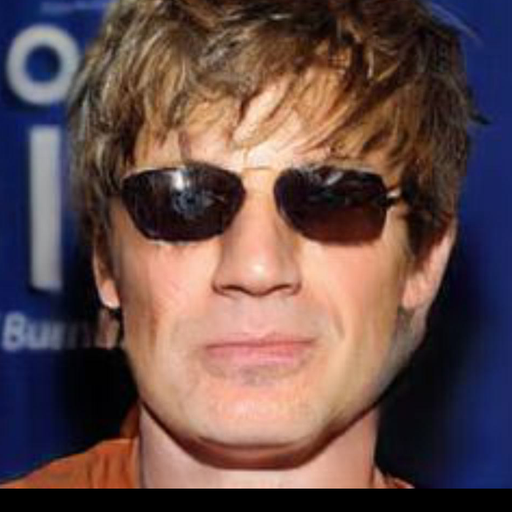}&
\interpfigt{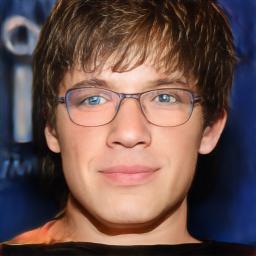}&
\interpfigt{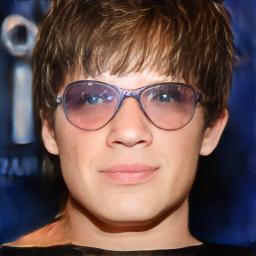}&
\interpfigt{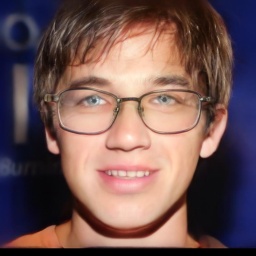}&
\interpfigt{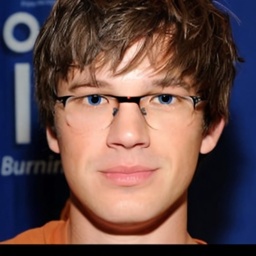}&
\interpfigt{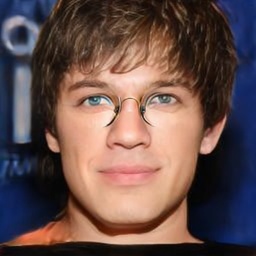}&
\interpfigt{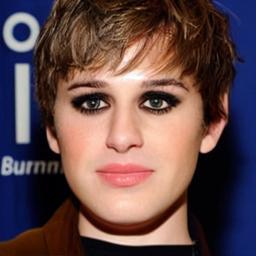}&
\interpfigt{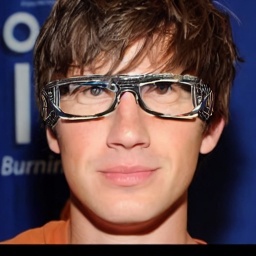}&
\interpfigt{Figures/NA.png}&
\interpfigt{Figures/NA.png}&
\interpfigt{Figures/NA.png}
\\

\interpfigt{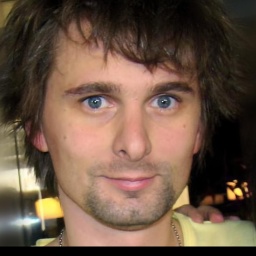}&
\interpfigt{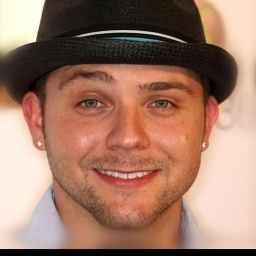}&
\interpfigt{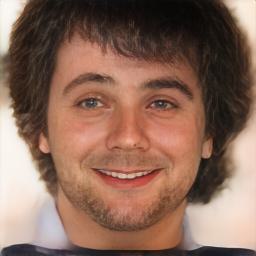}&
\interpfigt{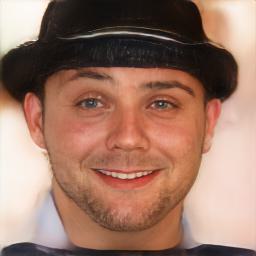}&
\interpfigt{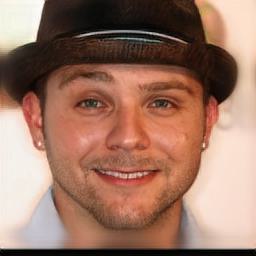}&
\interpfigt{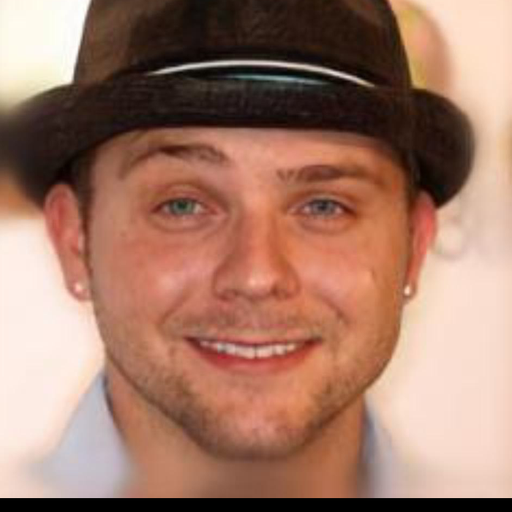}&
\interpfigt{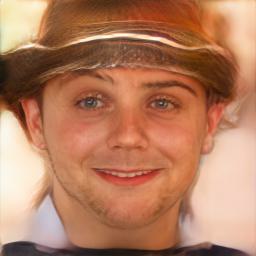}&
\interpfigt{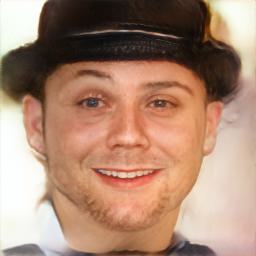}&
\interpfigt{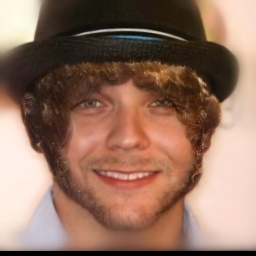}&
\interpfigt{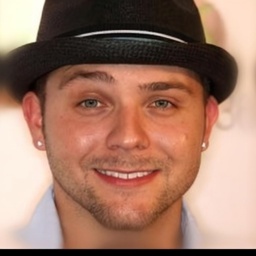}&
\interpfigt{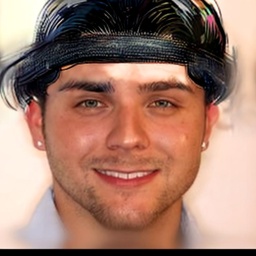}&
\interpfigt{Figures/NA.png}&
\interpfigt{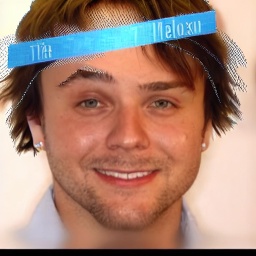}&
\interpfigt{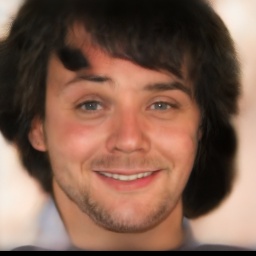}&
\interpfigt{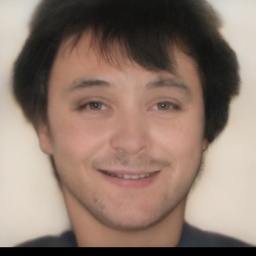}&
\interpfigt{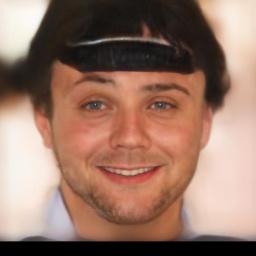}
\\

\interpfigt{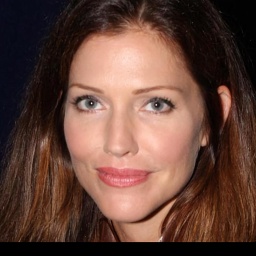}&
\interpfigt{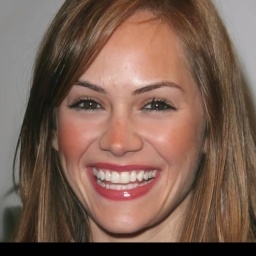}&
\interpfigt{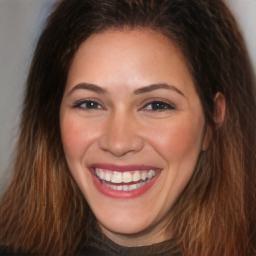}&
\interpfigt{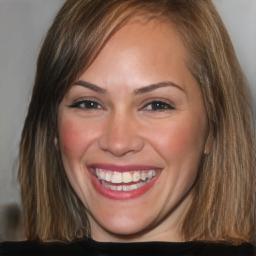}&
\interpfigt{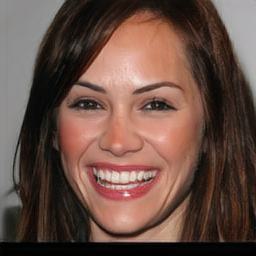}&
\interpfigt{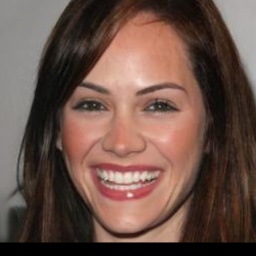}&
\interpfigt{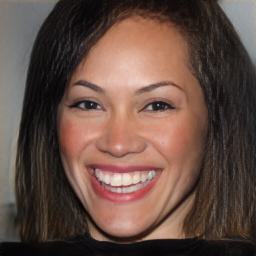}&
\interpfigt{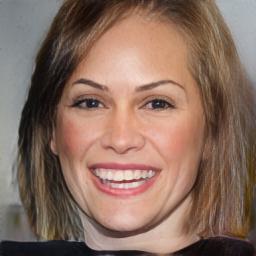}&
\interpfigt{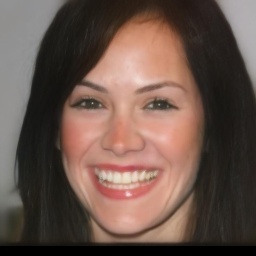}&
\interpfigt{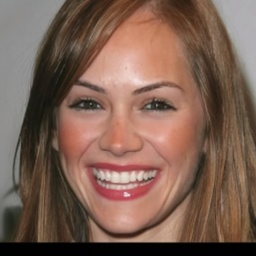}&
\interpfigt{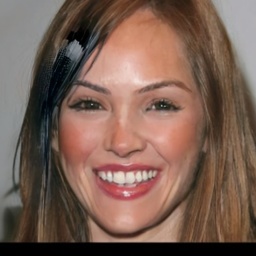}&
\interpfigt{Figures/NA.png}&
\interpfigt{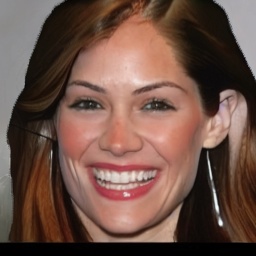}&
\interpfigt{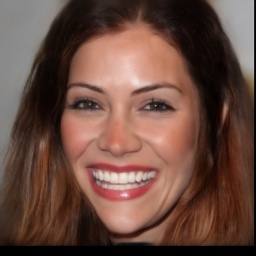}&
\interpfigt{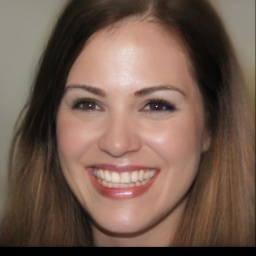}&
\interpfigt{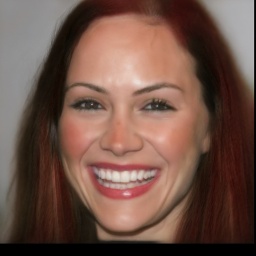}
\\

\interpfigt{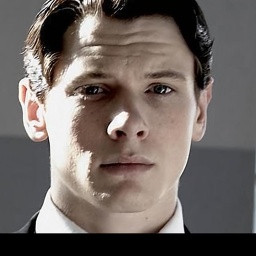}&
\interpfigt{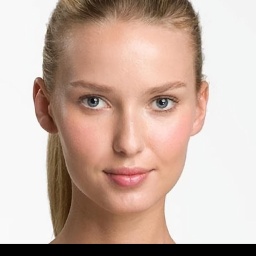}&
\interpfigt{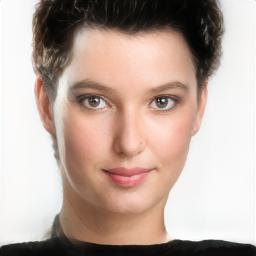}&
\interpfigt{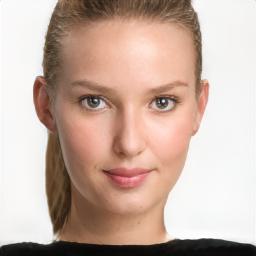}&
\interpfigt{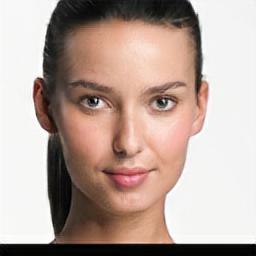}&
\interpfigt{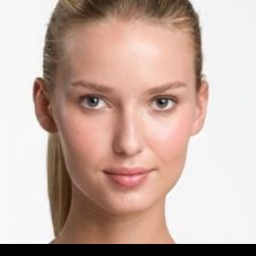}&
\interpfigt{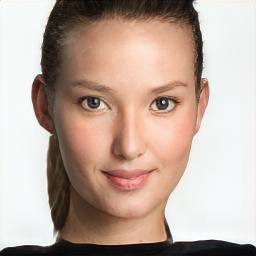}&
\interpfigt{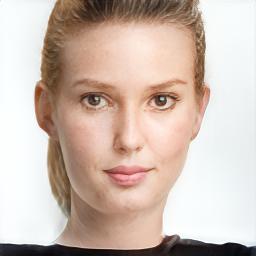}&
\interpfigt{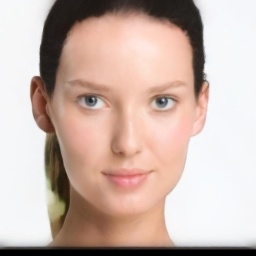}&
\interpfigt{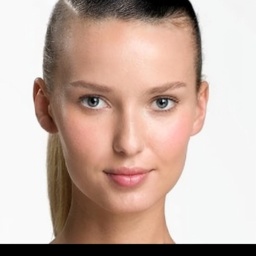}&
\interpfigt{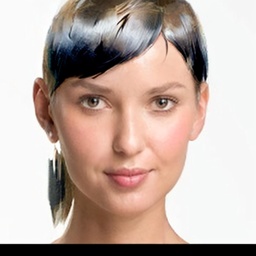}&
\interpfigt{Figures/NA.png}&
\interpfigt{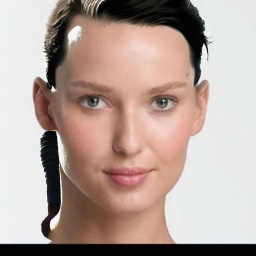}&
\interpfigt{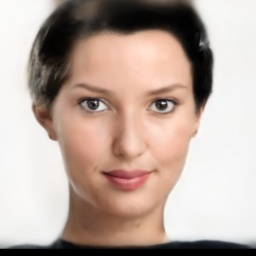}&
\interpfigt{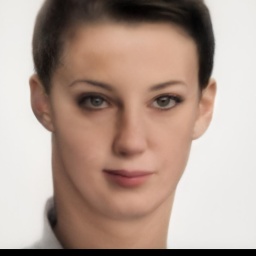}&
\interpfigt{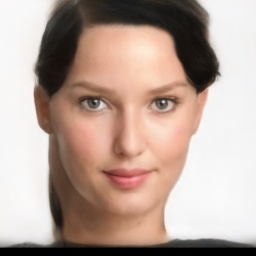}
\\

\interpfigt{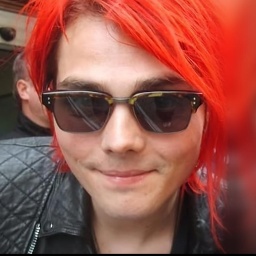}&
\interpfigt{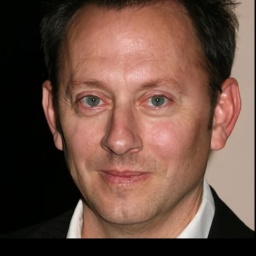}&
\interpfigt{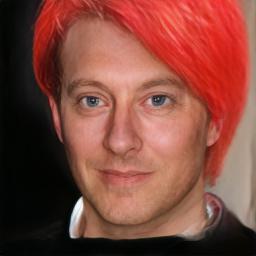}&
\interpfigt{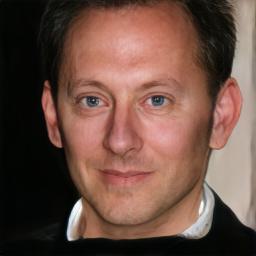}&
\interpfigt{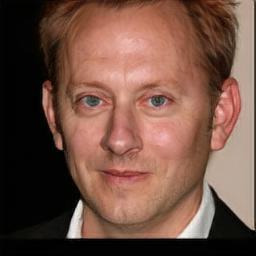}&
\interpfigt{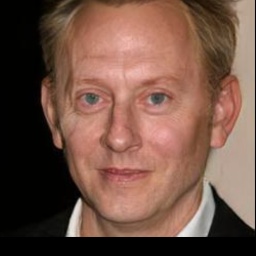}&
\interpfigt{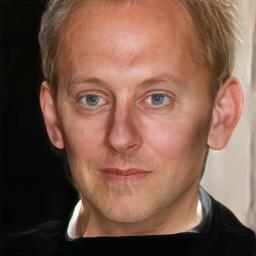}&
\interpfigt{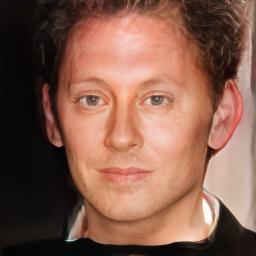}&
\interpfigt{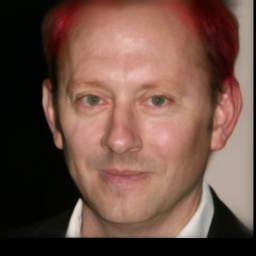}&
\interpfigt{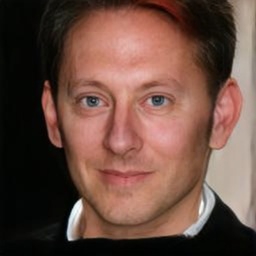}&
\interpfigt{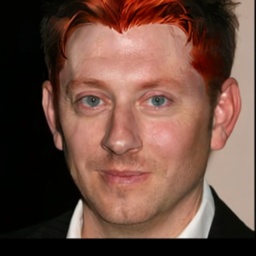}&
\interpfigt{Figures/NA.png}&
\interpfigt{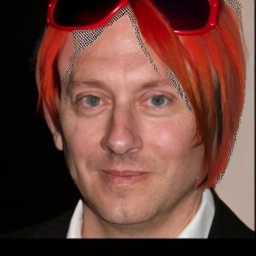}&
\interpfigt{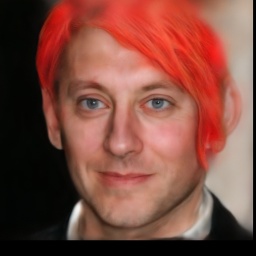}&
\interpfigt{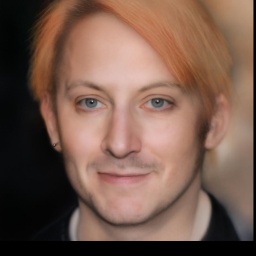}&
\interpfigt{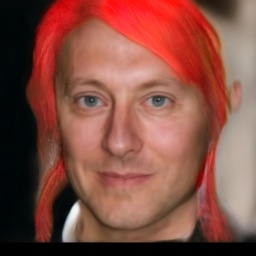}

\\
Reference & Source & \textbf{Ours} & E3DGE & HisD & VecGAN++ & InterFaceGAN  & StyleCLIP & SFE & LEDITS++ & InfEdit & NoiseCLR & Paint by Ex. & Barbershop & StyleFusion & HairCLIPv2\\
\end{tabular}
}
\caption{Comparisons with the competing editing methods for glasses addition and hair edits. Ours, HisD, VecGAN++, Barbershop, StyleFusion, HairCLIPv2, and Paint by Ex. use reference images for editing. InterFaceGAN, StyleCLIP, SFE, E3DGE, and NoiseCLR use previously calculated latent directions. LEDITS++ and InfEdit use text prompts. N/A indicates the model is incapable of such edits.}
\label{fig:full_comparison}
\end{figure*}

\begin{figure*}[t]
\centering
\Large
\scalebox{0.6}{
\setlength\tabcolsep{1pt}
\begin{tabular}{ccccccccccccc}

\rotatebox{90}{~~~Mouth} &
\interpfigt{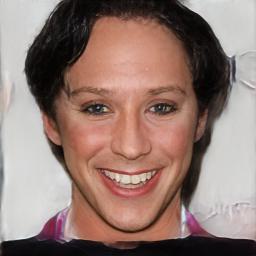}&
\interpfigt{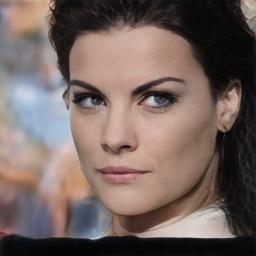}&
\interpfigt{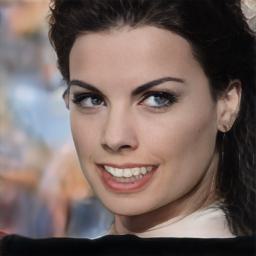}&
\interpfigt{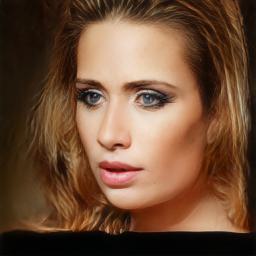}&
\interpfigt{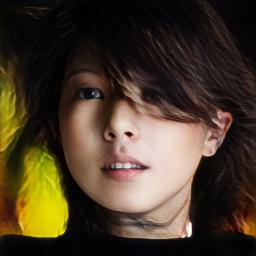}&
\interpfigt{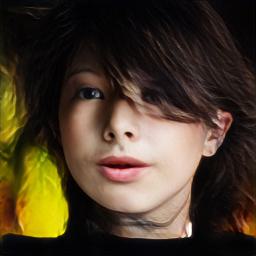}&
\interpfigt{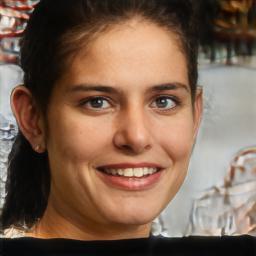}&
\interpfigt{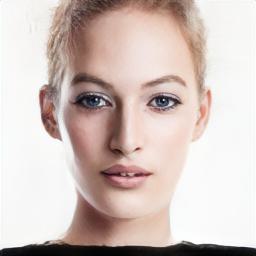}&
\interpfigt{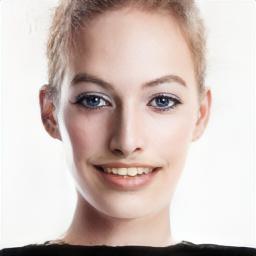}&
\interpfigt{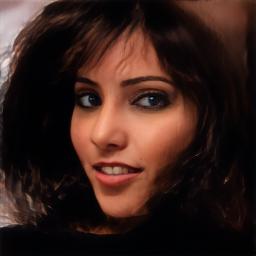}&
\interpfigt{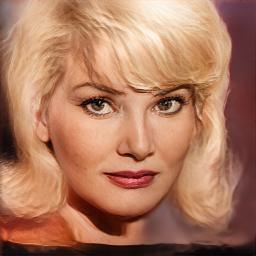}&
\interpfigt{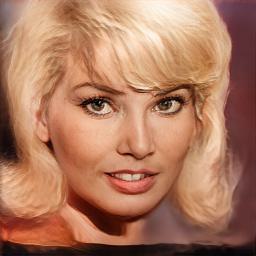}
\\

\rotatebox{90}{~~~Eyes} &
\interpfigt{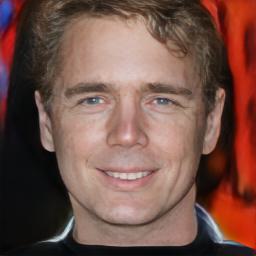}&
\interpfigt{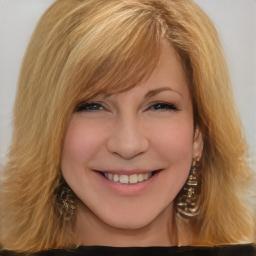}&
\interpfigt{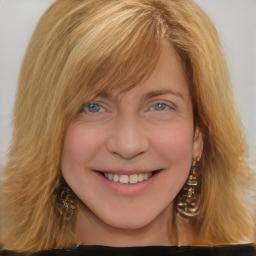}&
\interpfigt{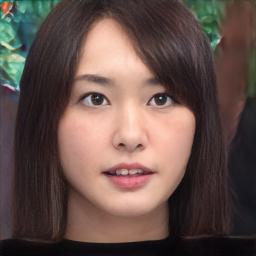}&
\interpfigt{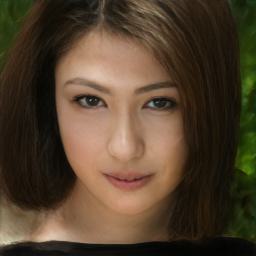}&
\interpfigt{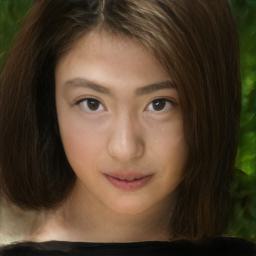}&
\interpfigt{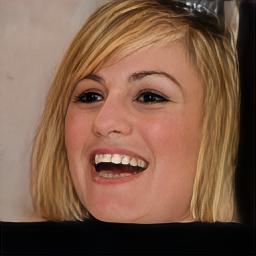}&
\interpfigt{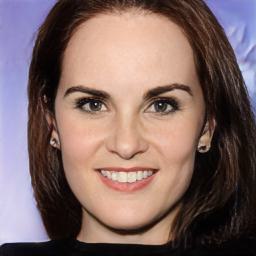}&
\interpfigt{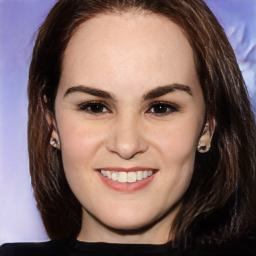}&
\interpfigt{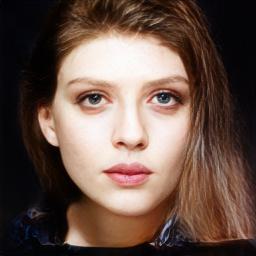}&
\interpfigt{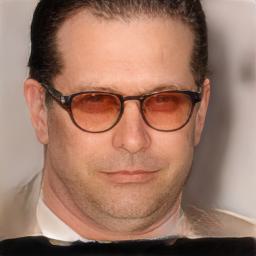}&
\interpfigt{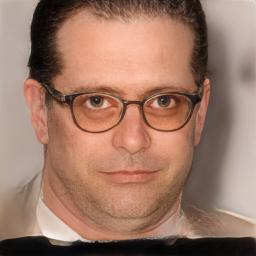}
\\

\rotatebox{90}{~~~Nose} &
\interpfigt{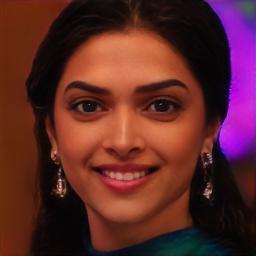}&
\interpfigt{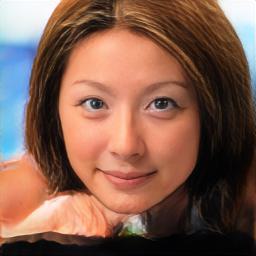}&
\interpfigt{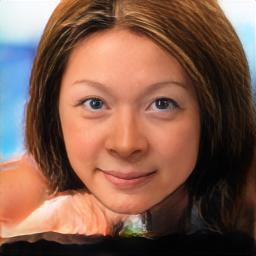}&
\interpfigt{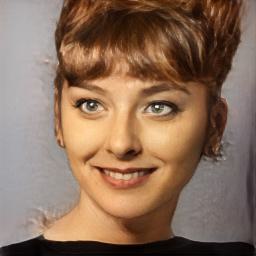}&
\interpfigt{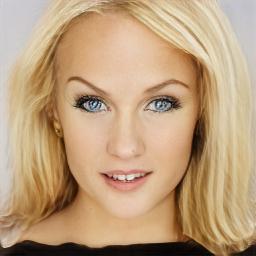}&
\interpfigt{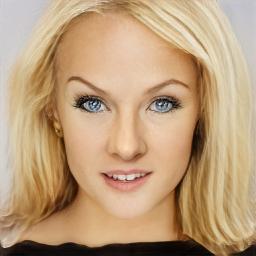}&
\interpfigt{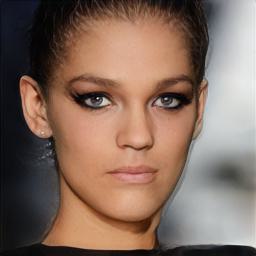}&
\interpfigt{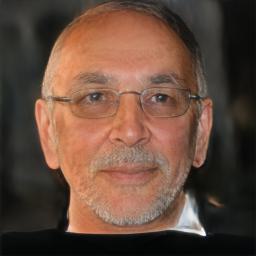}&
\interpfigt{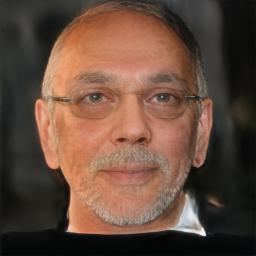}&
\interpfigt{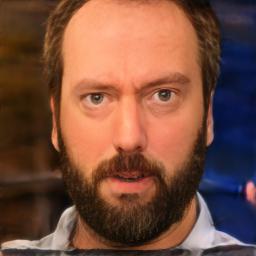}&
\interpfigt{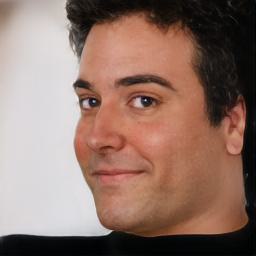}&
\interpfigt{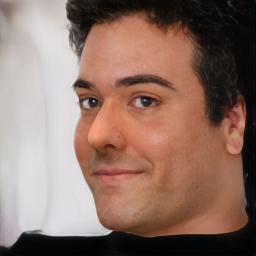}
\\

& Reference & Source & Output &Reference & Source & Output & Reference & Source & Output & Reference & Source & Output\\
\end{tabular}
}
\caption{
Additional editing examples from the CelebA dataset showcasing our method's ability to seamlessly incorporate features such as lips, eyes, and nose from reference to source, despite pose differences and interference like eyeglasses.}
\label{fig:other_editings_face}
\vspace{-0.2cm}
\end{figure*}

\noindent\textbf{Results.} 
We present quantitative and qualitative comparisons with competing methods in~\cref{table:full_comparison} and~\cref{fig:full_comparison}, respectively.
From~\cref{table:full_comparison}, it is evident that our method outperforms competing methods significantly in terms of FID and preserves identity better in the non-edited regions. 

Observing \cref{fig:full_comparison}, our method demonstrates superior performance in hair and glasses edits compared to competing methods. HisD and VecGAN++ struggle with maintaining fidelity to the reference, particularly with glasses, due to their reliance on low-rate latent spaces. While InterFaceGAN, StyleCLIP, and SFE can add glasses and perform some hair transfers, they falter with uncommon edits like hat removal (row 5) and red hair (row 8) due to the limitations of their $\mathcal{W}^+/\mathcal{F}$ spaces and their non-reference-based approach. E3DGE's pre-trained editing directions yield unsatisfactory results. LEDITS++ and InfEdit, being text-conditioned, fail to accurately reflect the original reference in their edits. NoiseCLR does not effectively explore hairstyle directions, and its glasses modifications are entangled with makeup changes (rows 2 and 4). Paint by Example can transfer some features but often produces severe out-of-domain artifacts (rows 1, 5-8). Barbershop and HairCLIPv2, optimized for hairstyle edits, suffer from geometric inconsistencies (row 8) and fail in some hair edit cases (row 5). Finally, StyleFusion’s feature transfer relies on $\mathcal{W}/\mathcal{W}^+$ directions, resulting in the loss of many high-rate details and outputs that do not fully reflect the original images, especially when the features cannot be well-represented in those domains (row 8).

\begin{table}[t!]
\caption{Results of our user study where participants are asked to identify the edited image. Based on this study, we find that our edits are challenging to distinguish. }
\centering
\footnotesize
\setlength\tabcolsep{2pt}
\begin{tabular}{r|ccc|c|cc|c}
\toprule
& \multicolumn{4}{c|}{\textbf{FFHQ}}  & \multicolumn{3}{c}{\textbf{AFHQ}} \\
& Eyes & Nose & Mouth & Overall & Eyes & Nose+Mouth  & Overall \\
\hline
Original  & 40\% & 29\% & 34\% & 34\% & 40\% & 42\% & 41\% \\
Ours (edited) & 49\% & 59\% & 40\% & 49\% & 49\% & 43\% & 46\%\\ 
Undecided & 11\% & 13\% & 26\% & 16\% & 11\% & 15\% & 13\% \\
\bottomrule
\end{tabular}
\label{table:user_study}
\end{table}

\begin{table}[t!]
\caption{Quantitative ablation study of our editing. V1 is the post-processing triplane gradients, V2 is the implicit fusion, and V3 fine-tunes the implicit fusion encoder.}
\centering
\footnotesize
\setlength\tabcolsep{2pt}
\begin{tabular}{ccc|ccc|ccc}
\toprule
& & & \multicolumn{3}{c|}{\textbf{Eyeglasses}} & \multicolumn{3}{c}{\textbf{Hair}} \\
V1 & V2 & V3 & {FID $\downarrow$} & 
$\mathcal{M}_{\text{SSIM}} \uparrow$ & $\mathcal{M}_{\mathcal{L}_2} \downarrow$ & {FID $\downarrow$} & 
$\mathcal{M}_{\text{SSIM}} \uparrow$ & $\mathcal{M}_{\mathcal{L}_2} \downarrow$ \\
\hline
\xmark & \xmark & \xmark & 79.50 & 0.8451 & 0.0323 & 82.42 & 0.8177 & 0.0195 \\
\cmark & \xmark & \xmark & 74.46 & 0.9814 & 0.0022 & 77.30 & 0.9674 & 0.0034 \\
\cmark & \cmark & \xmark & 68.19 & \textbf{0.9822} & \textbf{0.0020} & 67.04 & 0.9691 & 0.0033 \\
\cmark & \cmark & \cmark & \textbf{66.68} & 0.9818 & 0.0021  & \textbf{64.59} & \textbf{0.9720} & \textbf{0.0029}  \\
\bottomrule
\end{tabular}
\label{table:development}
\end{table}

\begin{figure}[t!]
\centering
\Large
\scalebox{0.55}{
\setlength\tabcolsep{1pt}
\begin{tabular}{cccccc}
\interpfigt{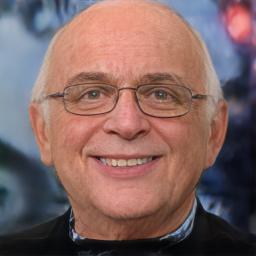}&
\interpfigt{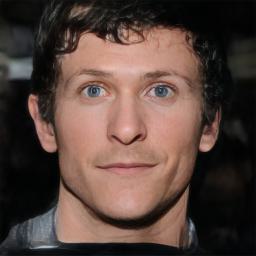}&
\interpfigt{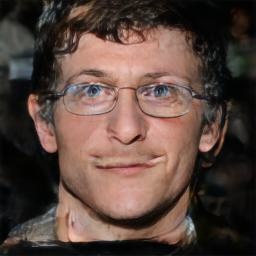}&
\interpfigt{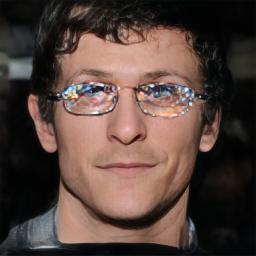}&
\interpfigt{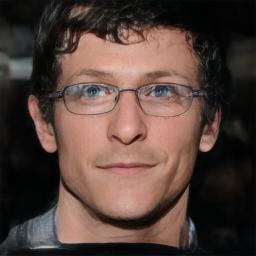}&
\interpfigt{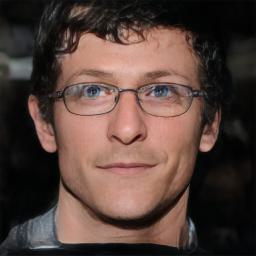} \\

\interpfigt{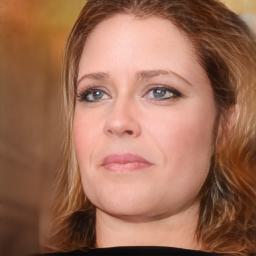}&
\interpfigt{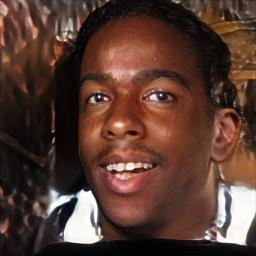}&
\interpfigt{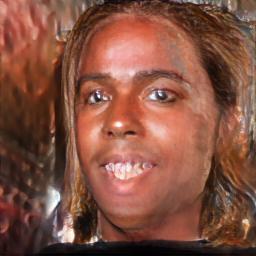}&
\interpfigt{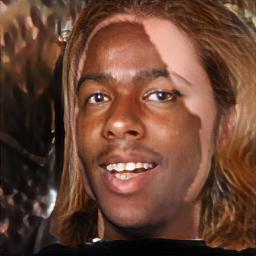}&
\interpfigt{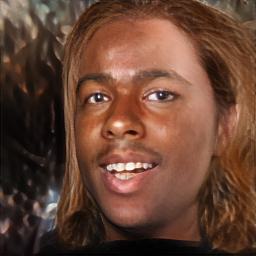}&
\interpfigt{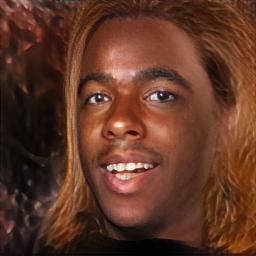} \\

\interpfigt{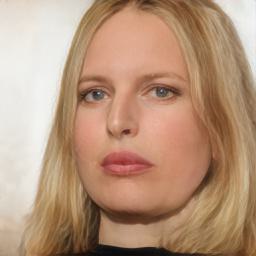}&
\interpfigt{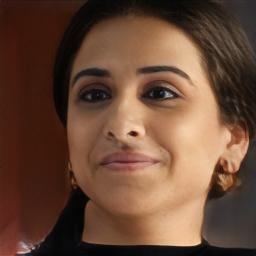}&
\interpfigt{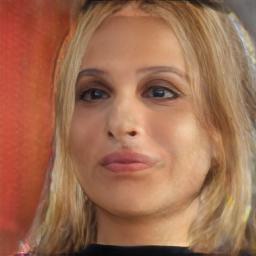}&
\interpfigt{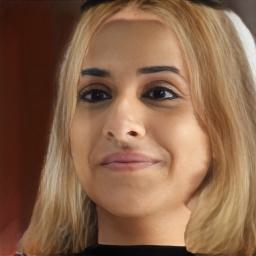}&
\interpfigt{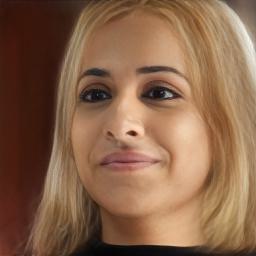}&
\interpfigt{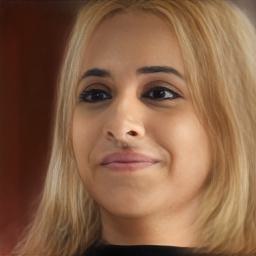}\\

\interpfigt{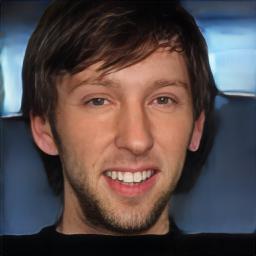}&
\interpfigt{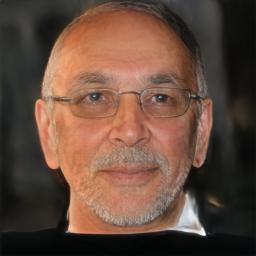}&
\interpfigt{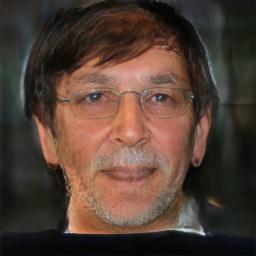}&
\interpfigt{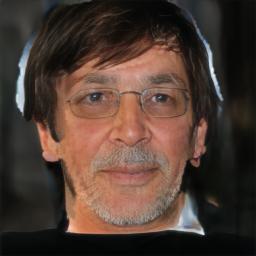}&
\interpfigt{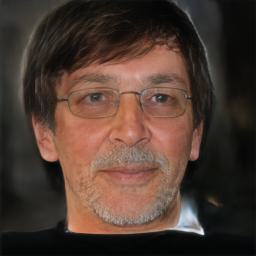}&
\interpfigt{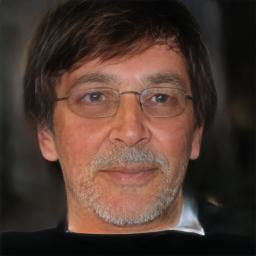}\\
Reference & Source & No V & +V1 & +V2 & +V3 \\
\end{tabular}
}
\vspace{-0.2cm}
\caption{Qualitative ablation study for glasses and hair edits, showing the effects of all fundamental stages of our pipeline.}
\vspace{-0.2cm}
\label{fig:development}
\end{figure}

\begin{figure}[t]
\centering
\setlength\tabcolsep{1pt}
\Large
\scalebox{0.55}
{
\begin{tabular}{ccccccc}
\rotatebox{90}{~~~Eyes} &
\interpfigt{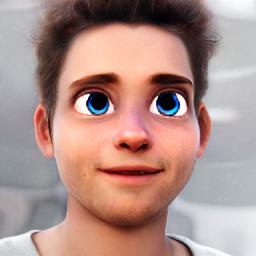}&
\interpfigt{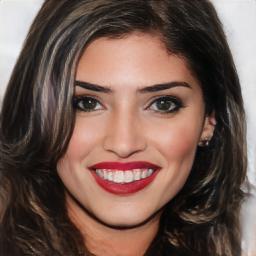}&
\interpfigt{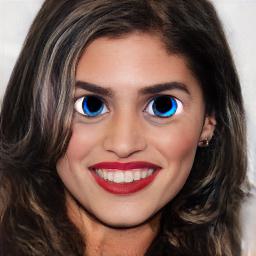}&
\interpfigt{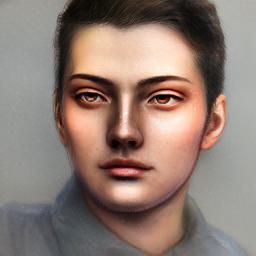}&
\interpfigt{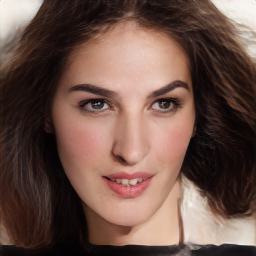}&
\interpfigt{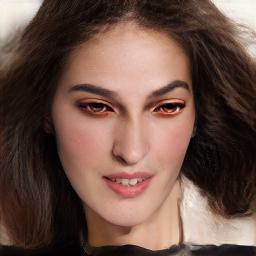} 
\\

\rotatebox{90}{~~~Mouth} &
\interpfigt{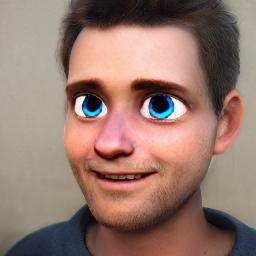}&
\interpfigt{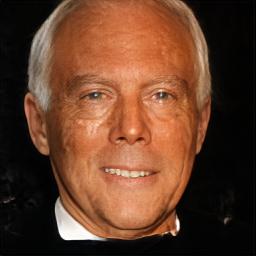}&
\interpfigt{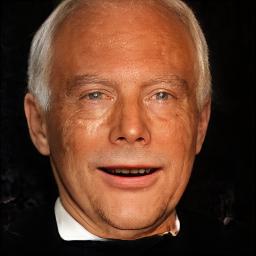}&
\interpfigt{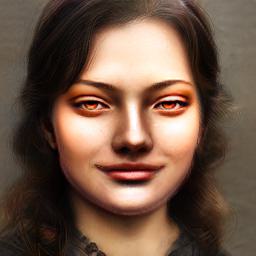}&
\interpfigt{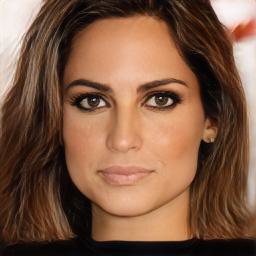}&
\interpfigt{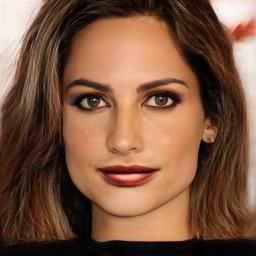} 
\\
 & Reference & Source & Output & Reference & Source & Output \\
\end{tabular}
}
\vspace{-0.2cm}
\caption{Cross-generator edits with stylizing. Our method achieves copying local parts from stylized images, such as cartoon portraits.}
\vspace{-0.5cm}
\label{fig:stylised_edits}
\end{figure}

\begin{figure}[t!]
\centering
\Large
\scalebox{0.55}{
\setlength\tabcolsep{1pt}
\begin{tabular}{ccccccc}
\rotatebox{90}{~~~Eyes} &
\interpfigt{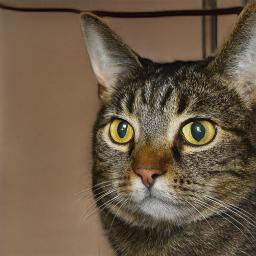}&
\interpfigt{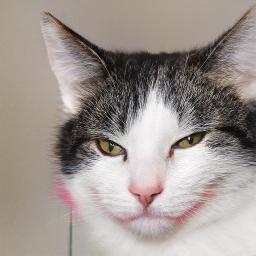}&
\interpfigt{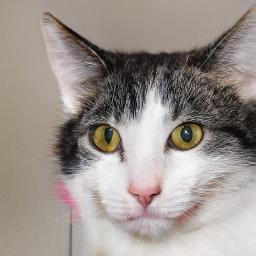}&
\interpfigt{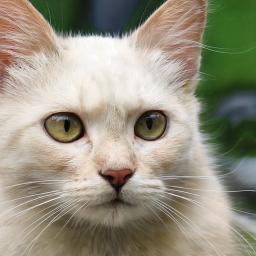}&
\interpfigt{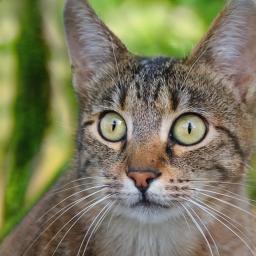}&
\interpfigt{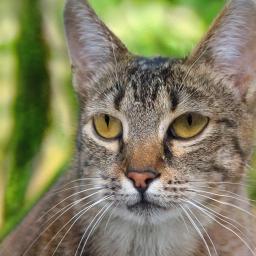}
\\
\rotatebox{90}{~~~Eyes} &
\interpfigt{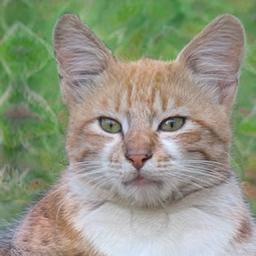}&
\interpfigt{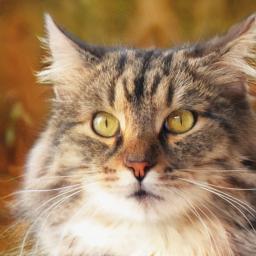}&
\interpfigt{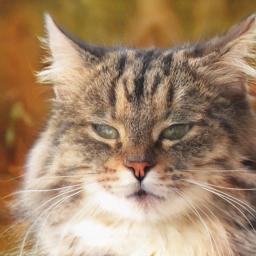}&
\interpfigt{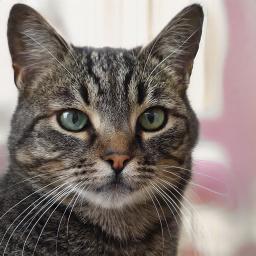}&
\interpfigt{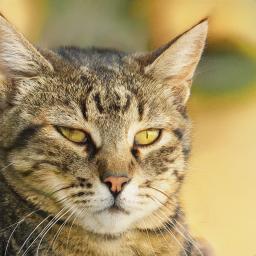}&
\interpfigt{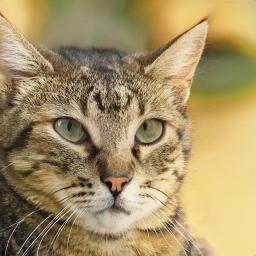}
\\
\rotatebox{90}{~~~Mouth} &
\interpfigt{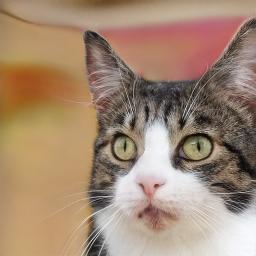}&
\interpfigt{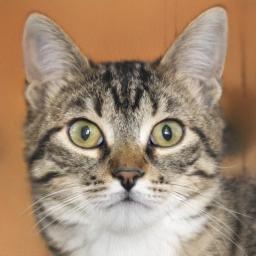}&
\interpfigt{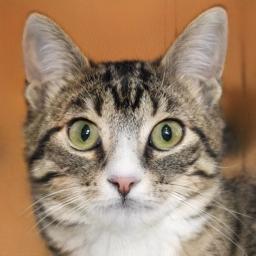}&
\interpfigt{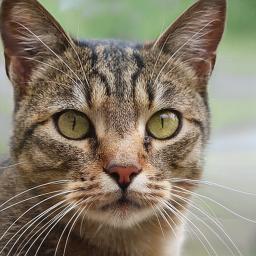}&
\interpfigt{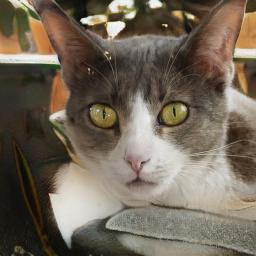}&
\interpfigt{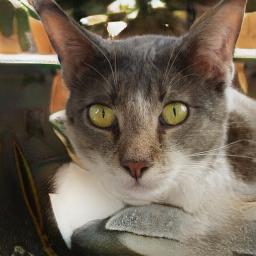}\\
\rotatebox{90}{~~~Mouth} &
\interpfigt{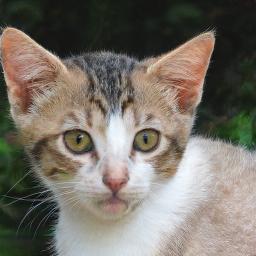}&
\interpfigt{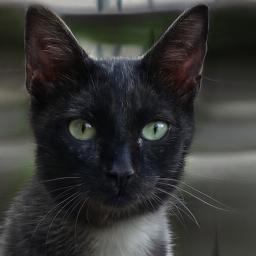}&
\interpfigt{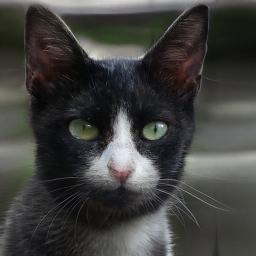}&
\interpfigt{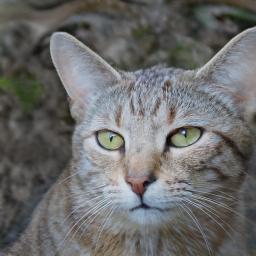}&
\interpfigt{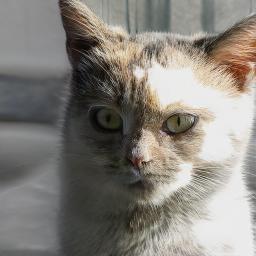}&
\interpfigt{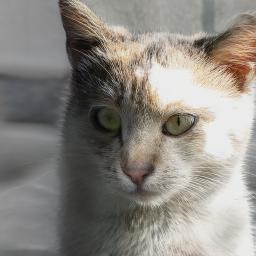}
\\
 & Reference & Source & Output &Reference & Source & Output \\
\end{tabular}
}
\vspace{-0.2cm}
\caption{Additional editing examples from AFHQ dataset. The nose and mouth are handled as a single part.}
\vspace{-0.5cm}
\label{fig:additional_cat}
\end{figure}

Next, we conduct a user study with 25 participants to evaluate our reference-based edits. Participants are shown original and edited images and asked to identify the edited ones. We utilize outputs of EG3D for both original and edited images to neutralize the influence of encoding on the results, and utilize the same angle for the source and edited images with random ordering to minimize bias. Participants could also choose ''{undecided}" if they find it difficult to distinguish. The study focus on edits to mouth, eyes, and nose on the FFHQ dataset, and eyes, nose \& mouth on the AFHQ dataset. Some participants frequently chose "undecided", while others perform near random chance, in \cref{table:user_study}.

\begin{figure*}[t!]
\centering
\Large
\setlength\tabcolsep{0.5pt}
\scalebox{0.5}{
\begin{tabular}{ccc}
     \includegraphics[scale=0.28]{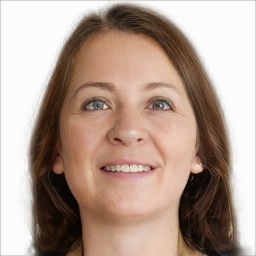}&
     \includegraphics[scale=0.28]{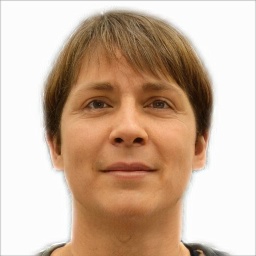}&
     \includegraphics[scale=0.28]{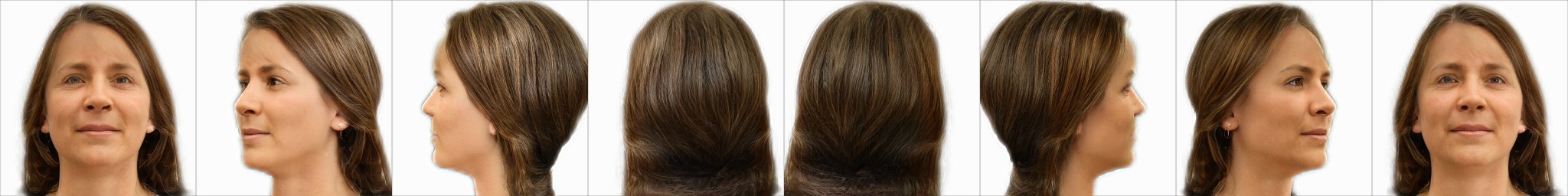}
     \\
     \includegraphics[scale=0.28]{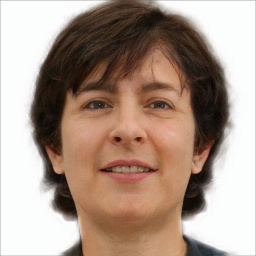}&
     \includegraphics[scale=0.28]{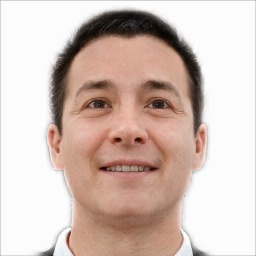}&
     \includegraphics[scale=0.28]{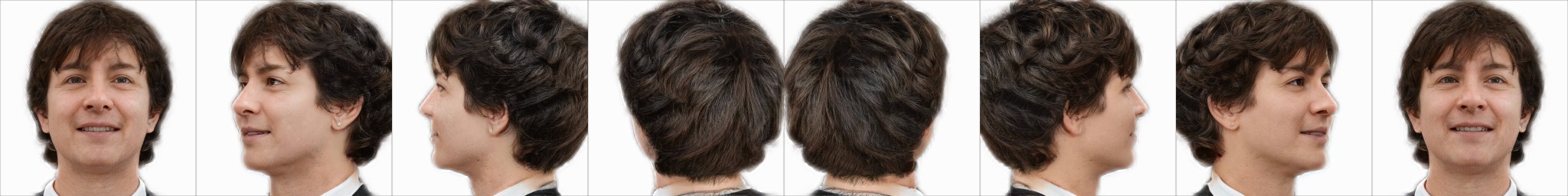}
     \\
     Reference & Source & Edited 3D outputs
\end{tabular}
}
\vspace{-0.2cm}
\caption{Extending our method on full-head hair edits on~\cite{An_2023_CVPR}.}
\label{fig:panohead_edits}
\end{figure*}

\noindent\textbf{Ablation study.} 
We demonstrate the improvements during the development of our pipeline stages, both quantitatively and qualitatively, in~\cref{table:development} and~\cref{fig:development}, respectively.

In our initial ablation study, we apply~\cref{eqn:manual_fusion} to merge the triplanes using a mask calculated via autograd function without any post-processing (No V).
Due to the intricate volumetric function affecting many pixels for each value in the triplane, the initial mask fails to stitch images effectively, resulting in blurry outputs.
Following the introduction of post-processing (+V1), as described in~\cref{sec:localizingtriplane}, we successfully achieve clear stitching, as depicted in~\cref{fig:development}. This allows us to transfer the hair of one person to our input while aligning the features using the canonical representation of the triplane.
However, the resulting output still lacks realism because it combines features from two different images with varying illuminations, identities, and skin colors (row 2).
To address the issue of smoothness at stitch boundaries, we follow~\cref{sec:implicit} and perform encoding and decoding via the pretrained encoder and decoder, respectively, on the fused triplane. Since these encoders are trained with real images, they know about real image priors. Despite the input image not being realistic, as shown in~\cref{fig:development} (V1), the encoder successfully encodes its latent to the generator's natural latent space while attempting to preserve the identity in (+V2).
However, a pretrained encoder optimized for projecting real images onto the generator's latent space is not optimal for our specific use case. Consequently, we replace the encoder with one trained specifically for this task, elaborated in~\cref{sec:fine-tune}. This specialized encoder (+V3) ensures color preservation (row 2) and enhances editing details, such as the coherence of eyeglass frames, as well as reducing background leakage (rows 1 and 4).

\noindent\textbf{Cross-generator edits.} We also provide novel edits in~\cref{fig:stylised_edits}, where the reference and source triplanes are gathered from stylized and non-stylized generators, respectively. Specifically, we utilize~\cite{song2022diffusion} and fine-tune the EG3D backbones via different text prompts~\cite{poole2023dreamfusion}. Then, we synthesize stylized triplanes ($\textbf{T}_\text{ref}$) and perform reference-based editing on the triplanes of default EG3D ($\textbf{T}_\text{src}$). The rest of the steps are the same as before, using the original EG3D and encoder we train for EG3D. The results presented in~\cref{fig:stylised_edits} demonstrate our method's independence from backbones, showcasing its capability to achieve part-based attribute stylization. This differs from~\cite{song2022diffusion}, which offers global stylization.

\begin{figure}[t!]
    \centering
    \footnotesize
    \setlength\tabcolsep{0.5pt}
    \begin{tabular}{cc}
        & References \\
        & \includegraphics[width=0.75\linewidth]{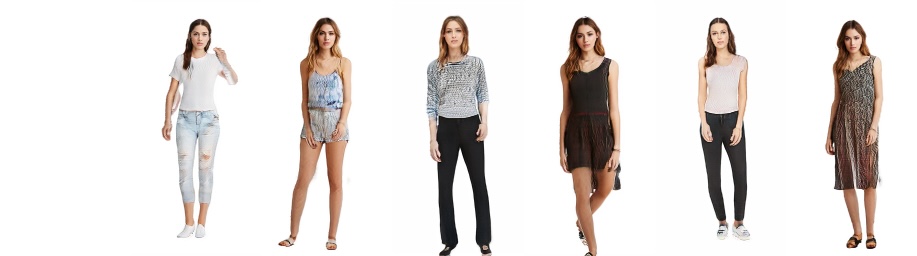}
        \\
        \rotatebox{90}{\quad \quad Top} & \includegraphics[width=0.75\linewidth]{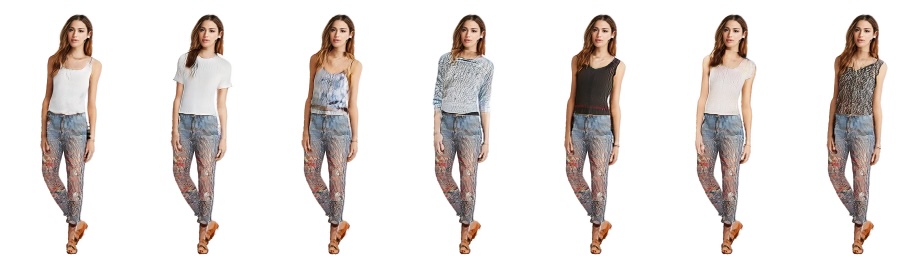}
        \\
        \rotatebox{90}{\quad \quad Pants} & \includegraphics[width=0.75\linewidth]{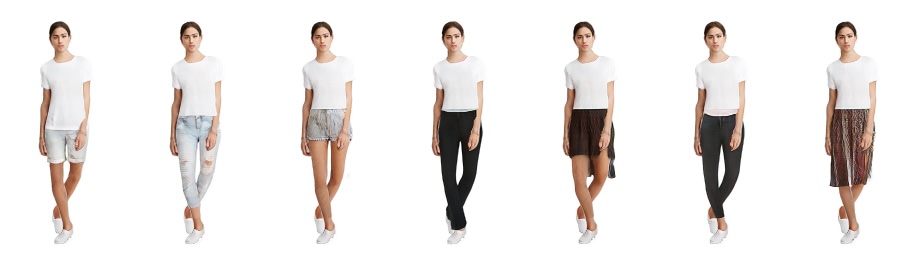}
    \end{tabular}
    \vspace{-0.2cm}
    \caption{Extending our method on try-on edits on~\cite{dong2023ag3d}.}
    \label{fig:fashion_edits}
    \vspace{-0.5cm}
\end{figure}

\noindent\textbf{Full body and 360-degree head edits.}
\cref{fig:other_editings_face} demonstrates challenging human face reference-based edits on EG3D~\cite{chan2022efficient} like transferring lips, eyeglasses, and nose from one person to another.~\cref{fig:additional_cat} shows edits on animal face parts for eyes, nose, and mouth.~\cref{fig:fashion_edits} demonstrates fashion edits on AG3D~\cite{dong2023ag3d} trained with DeepFashion~\cite{liuLQWTcvpr16DeepFashion} dataset.~\cref{fig:panohead_edits} extends human face part edits to full 360-degree hair edits on PanoHead~\cite{An_2023_CVPR}. It is evident that our approach is generalizable to different triplane generators.

While extending our method to different triplane-based generators and datasets, we only changed the 2D segmentation network and the encoder fine-tuning dataset to comply with the generator, when required. We also provide multi-view image results and more examples in Supplementary.

\noindent\textbf{Generalizing to class-agnostic edits.} Given the importance of large reconstruction models~\cite{hong2024lrm,xu2024instantmesh,lan2024ln3diff,zhang2024clay,lan2024ga,chen20243dtopiaxlscalinghighquality3d,xiang2024structured,jun2023shapegeneratingconditional3d}, we extend our method to arbitrary object edits using the triplanes of LN3Diff and InstantMesh~\cite{xu2024instantmesh,lan2024ln3diff} in~\cref{fig:teaser}, proving the potential capabilities of our method.%

\begin{figure}
\centering
\scriptsize
\setlength\tabcolsep{0.5pt}
\begin{tabular}{ccccccc}
Src & Ref (1) & Ref (2) & Ref (3) & Hair (1) & Eyes (2) & Lips (3)\\
\includegraphics[width=.065\textwidth]{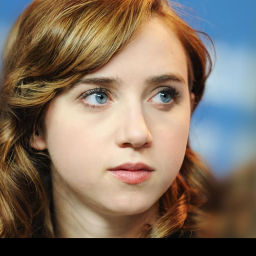}&
\includegraphics[width=.065\textwidth]{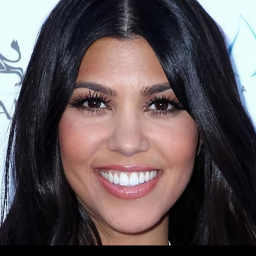}&
\includegraphics[width=.065\textwidth]{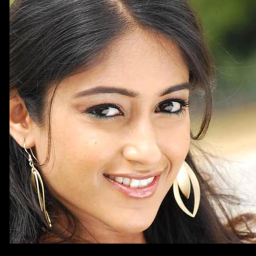}&
\includegraphics[width=.065\textwidth]{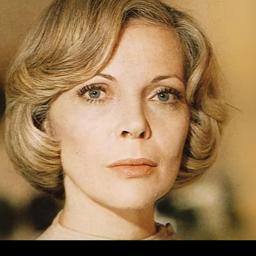}&
\includegraphics[width=.065\textwidth]{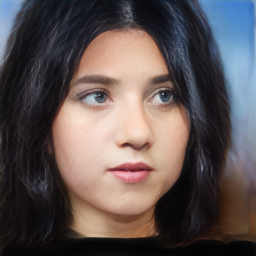}&
\includegraphics[width=.065\textwidth]{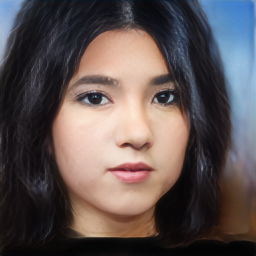}&
\includegraphics[width=.065\textwidth]{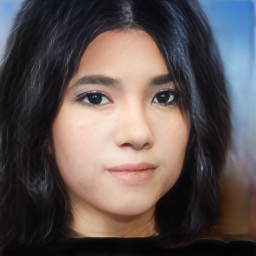}
\\
\includegraphics[width=.065\textwidth]{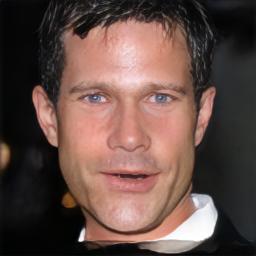}&
\includegraphics[width=.065\textwidth]{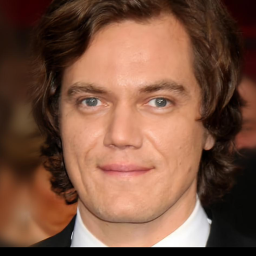}&
\includegraphics[width=.065\textwidth]{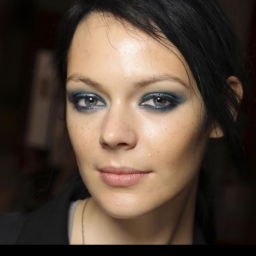}&
\includegraphics[width=.065\textwidth]{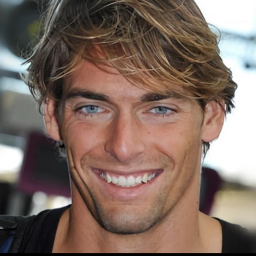}&
\includegraphics[width=.065\textwidth]{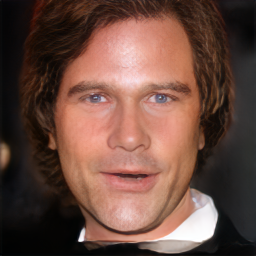}&
\includegraphics[width=.065\textwidth]{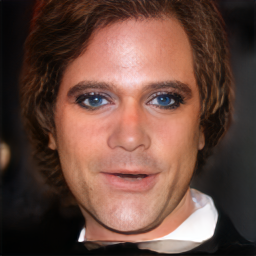}&
\includegraphics[width=.065\textwidth]{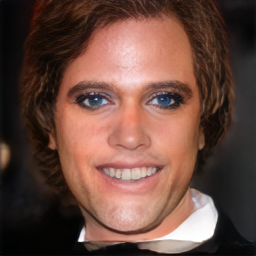}
\\
\end{tabular}
\vspace{-0.2cm}
\caption{Simultaneous editing results.}
\vspace{-0.5cm}
\label{fig:sim_editing}
\end{figure}

\section{Conclusion}

In conclusion, our work presents a comprehensive framework for reference-based, 3D-aware image editing, leveraging the unique capabilities of triplanes. Through spatial disentanglement and fusion learning, we achieve seamless integration of reference attributes while preserving the identity of the input image. We have shown our method's effectiveness through extensive qualitative and quantitative experiments. Our approach fills a crucial gap by offering a unified and generalizable solution.

\textbf{Limitations.} Our approach relies on the capabilities of EG3D, AG3D, PanoHead, LN3Diff and InstantMesh, which may struggle with background generation and high-quality reconstruction. Consequently, in some instances, rich background details may not be fully presented (row 4 on~\cref{fig:full_comparison}). However, this can be mitigated by not relying on the generator for background generation. %

\textbf{Future work.} Extending our reference-based editing approach beyond triplanes using large reconstruction models~\cite{zhang2024clay,lan2024ga,chen20243dtopiaxlscalinghighquality3d,xiang2024structured,jun2023shapegeneratingconditional3d} is an underexplored path. Specifically, on DiT-based approaches, we believe that the reference and source image tokens can be processed jointly, and the masks created via gradient accumulation can be applied onto self-attention layers for implicit fusion. We think that this can also eliminate the necessity for a canonical space where the source and reference features must be aligned, as the attention mechanism can handle misalignment easily.

\clearpage
{
    \small

}
\clearpage
\maketitlesupplementary
\setcounter{page}{1}

\usemintedstyle{tango}

\section{Hyperparameters}
We provide the mentioned hyperparameters of the main paper here.

\noindent\textbf{Triplane gradient post-processing.} We provide the code snippet for post-processing stage in~\cref{lst:post_processing}, and the default parameters for $\epsilon$, $\mu$, and $\sigma$. For the sake of simplicity, we omitted the batch dimension in \textbf{t}.

\begin{listing*}[ht]
\begin{minted}[mathescape, linenos, fontsize=\footnotesize]{python}
import torchvision.transforms as T
def postprocess_tp_mask(t, epsilons=(0.9, 1.1), mu=7, sigmas=(0.1, 2.0)):
    blur_func = T.GaussianBlur(mu, sigma=sigmas)

    3,C_N,H,W = t.shape
    t = t.view(3*C_N,H,W) # 96 256 256 for EG3D and AG3D, 288 256 256 for PanoHead
    C = t.shape[0]
    for i in range(C):
        t[i] -= torch.min(t[i])
        t[i] /= torch.max(t[i])
    for i in range(C):
        mu = t[i].mean()
        # take around mean region
        t[i] = (t[i]<epsilons[1]*mu) * (t[i]>epsilons[0]*mu)
        t[i] = blur_func(t[i].unsqueeze(0)).squeeze(0)
    # get the inverse, we are interested in outside of the mean
    t = 1-t
    # binarize
    mean_grad = torch.mean(t, dim=0, keepdim=True).repeat(C,1,1)
    b_mask = torch.zeros_like(mean_grad)
    b_mask[mean_grad>mean_grad.mean()] = 1.0
    b_mask = blur_func(b_mask.unsqueeze(0)).squeeze(0)
    t = b_mask
    t = t.view(3,C_N,H,W)
    return t
\end{minted}
\caption{Post-processing the triplane gradients.}
\label{lst:post_processing}
\end{listing*}

\noindent\textbf{Fusion of the triplanes.}~\cref{eqn:final_fusion_ablation} describes the fusion operation, where $\textbf{T}_\text{imp}$ is the implicitly fused triplane, and $\mathcal{E}$ is the morphological erosion operation applied on the post-processed triplane masks $\textbf{M}$.
\begin{equation}
   \textbf{T}_\text{f} = \mathcal{E}( \textbf{M}_\text{ref})*\textbf{T}_\text{ref} + \mathcal{E}(\textbf{M}_\text{src})*\textbf{T}_\text{src} + (\mathcal{E}(\textbf{M}_\text{src})-\mathcal{E}( \textbf{M}_\text{ref}))*\textbf{T}_\text{imp}
   \label{eqn:final_fusion_ablation}
\end{equation}

Snippet for the masking operations are given in~\cref{lst:fusion}. Morphological operations are realized with \texttt{max\_pool2d} and Gaussian blurring. Note that \texttt{dilated\_tp\_mask} can be utilized when $\textbf{M}_\text{ref} = 1 - \textbf{M}_\text{src}$ is set. Batch size is set to 1.

\begin{listing*}[ht]
\begin{minted}[mathescape, linenos, fontsize=\footnotesize]{python}
import torch.nn.functional as F
import torchvision.transforms as T
def create_dilated_eroded_tp_masks(self, tp_mask, 
                                   blur_k_size=9, std_devs=(2.0,2.0), morph_k_size=11):
    blur = T.GaussianBlur(blur_k_size, sigma=std_devs)
    1,3,N_C,H,W = tp_mask.shape # N_C = 32 for EG3D and AG3D, 96 for PanoHead
    dilated_tp_mask = blur(F.max_pool2d(tp_mask.view(1,3*N_C,H,W), kernel_size=morph_k_size,
                    stride=1, padding=(morph_k_size - 1) // 2)).view(1,3,N_C,H,W)
    eroded_tp_mask = blur((1 - F.max_pool2d((1 - tp_mask.view(1,3*N_C,H,W)), kernel_size=morph_k_size, 
                    stride=1, padding=(morph_k_size - 1) // 2))).view(1,3,N_C,H,W)
    return eroded_tp_mask, dilated_tp_mask
\end{minted}
\caption{Dilating and eroding triplane masks.}
\label{lst:fusion}
\end{listing*}

\section{Additional visual results} 
\begin{figure*}[ht!]
\centering
\setlength\tabcolsep{1pt}

\scalebox{0.95}{
\begin{tabular}{cccccc}
\interpfigt{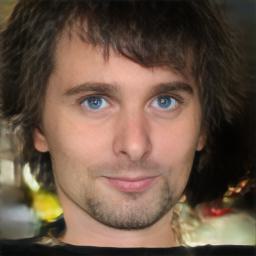}&
\interpfigt{Figures/hair/ours/0000_rec_src_origpose.jpg}&
\interpfigt{Figures/hair/ours/0000_rec_edited_origpose.jpg}&
\interpfigt{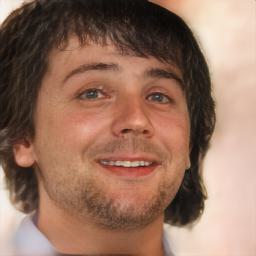}&
\interpfigt{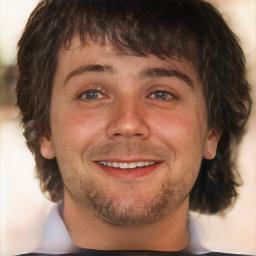}&
\interpfigt{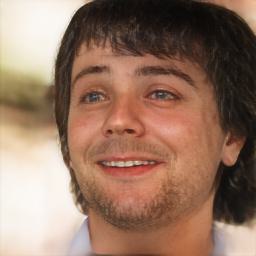} \\

\interpfigt{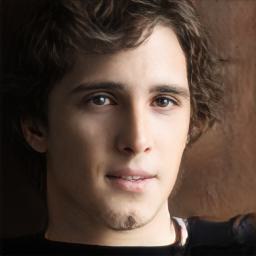}&
\interpfigt{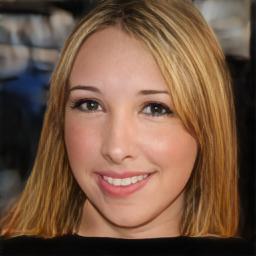}&
\interpfigt{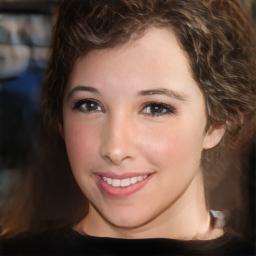}&
\interpfigt{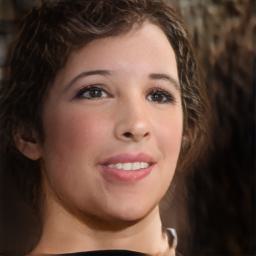}&
\interpfigt{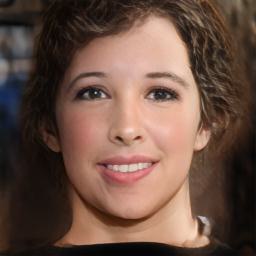}&
\interpfigt{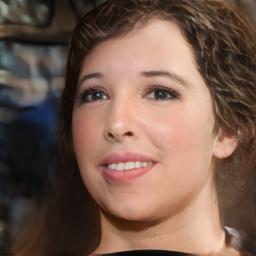} \\

\interpfigt{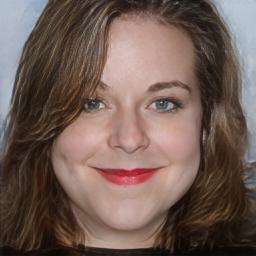}&
\interpfigt{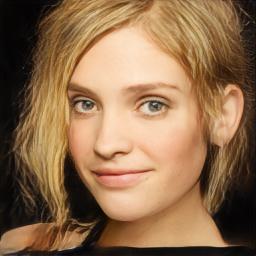}&
\interpfigt{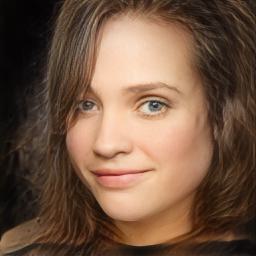}&
\interpfigt{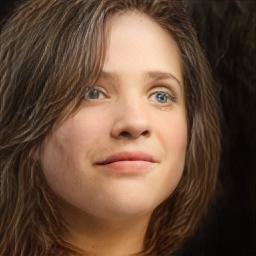}&
\interpfigt{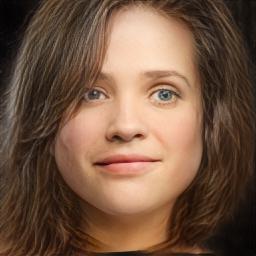}&
\interpfigt{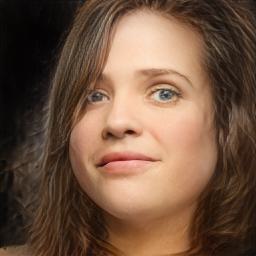} \\

\interpfigt{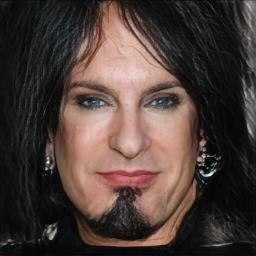}&
\interpfigt{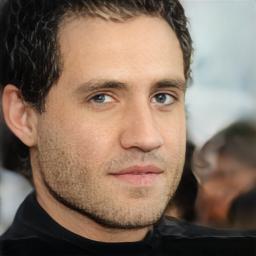}&
\interpfigt{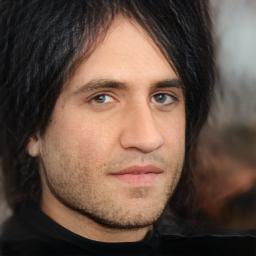}&
\interpfigt{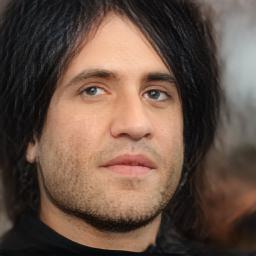}&
\interpfigt{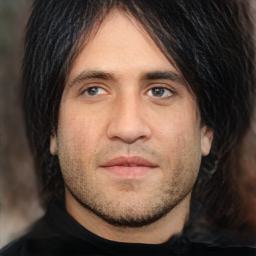}&
\interpfigt{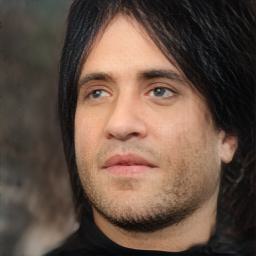} \\

\interpfigt{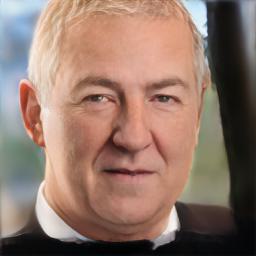}&
\interpfigt{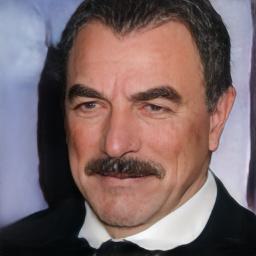}&
\interpfigt{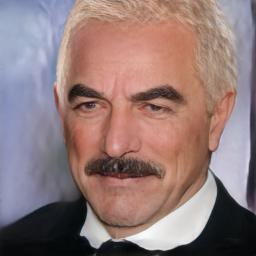}&
\interpfigt{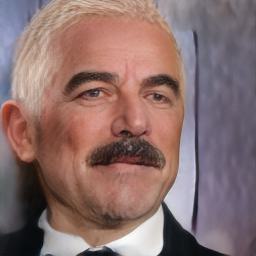}&
\interpfigt{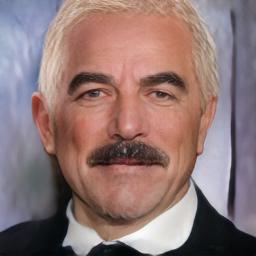}&
\interpfigt{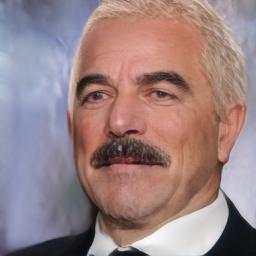} \\

\interpfigt{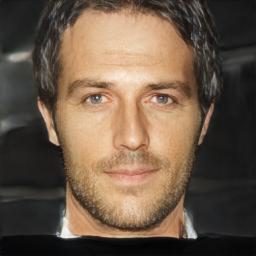}&
\interpfigt{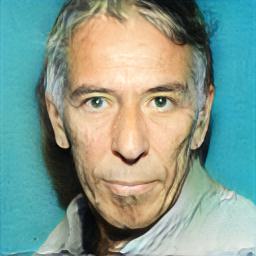}&
\interpfigt{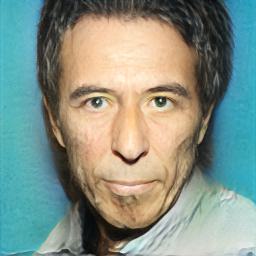}&
\interpfigt{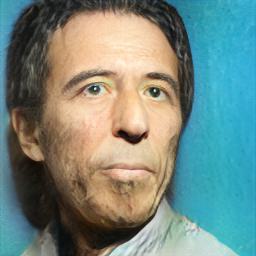}&
\interpfigt{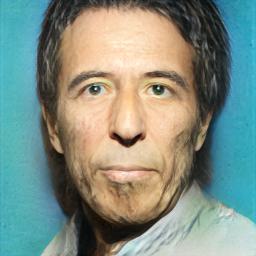}&
\interpfigt{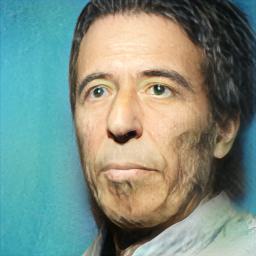} \\

\interpfigt{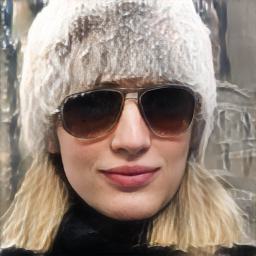}&
\interpfigt{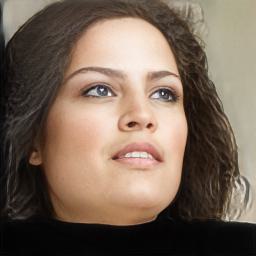}&
\interpfigt{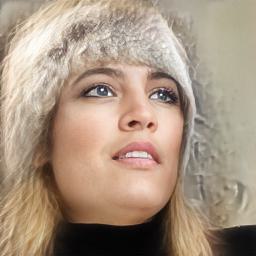}&
\interpfigt{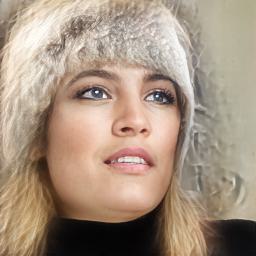}&
\interpfigt{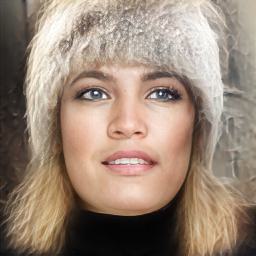}&
\interpfigt{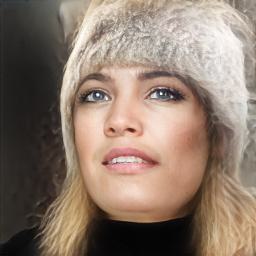} \\

\interpfigt{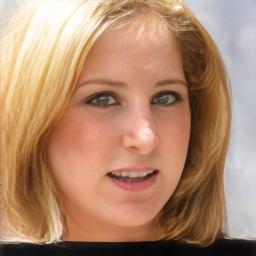}&
\interpfigt{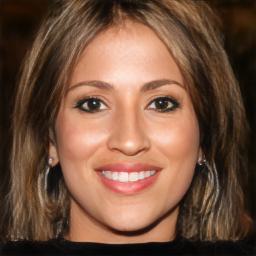}&
\interpfigt{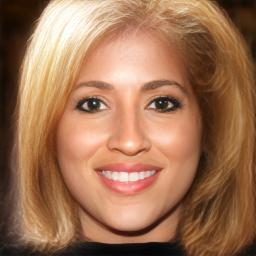}&
\interpfigt{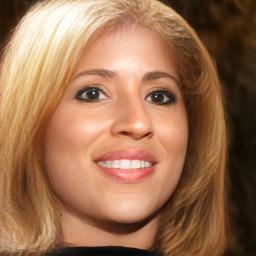}&
\interpfigt{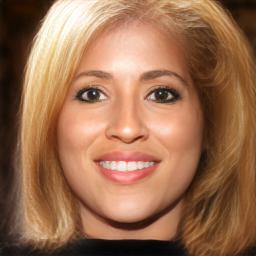}&
\interpfigt{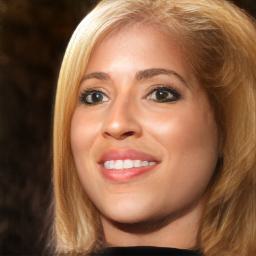} 
\\

\interpfigt{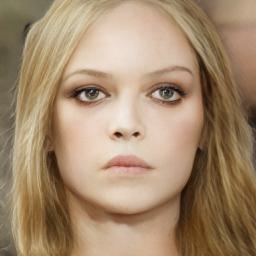}&
\interpfigt{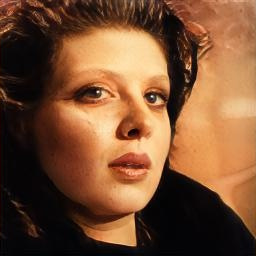}&
\interpfigt{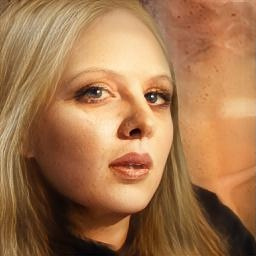}&
\interpfigt{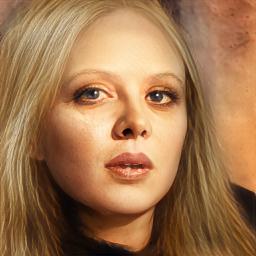}&
\interpfigt{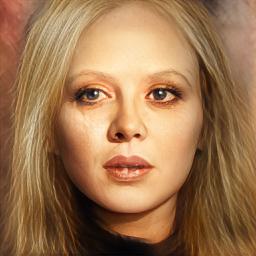}&
\interpfigt{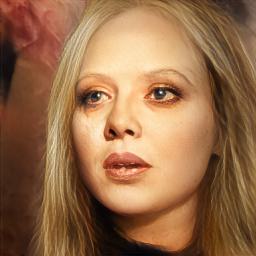} 
\\
Reference & Source & \multicolumn{4}{c}{Multi-view Outputs}\\
\end{tabular}
}
\caption{Hair edits and 3D visualizations.}
\label{fig:suppl_hair_edits}
\end{figure*}

\begin{figure*}[ht!]
\centering
\setlength\tabcolsep{1pt}
\scalebox{1.0}{
\begin{tabular}{ccccccccc}
\interpfigt{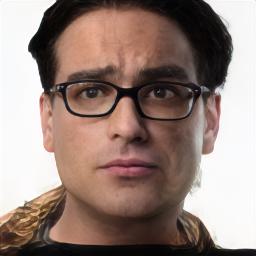}&
\interpfigt{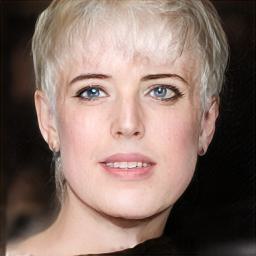}&
\interpfigt{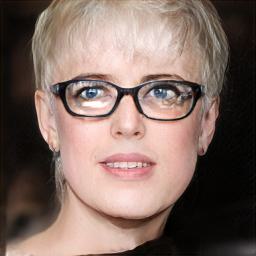}&
\interpfigt{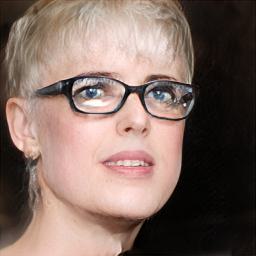}&
\interpfigt{Figures/glasses/ours/30/0030_rec_edited_1.jpg}&
\interpfigt{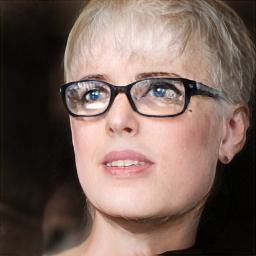}
\\

\interpfigt{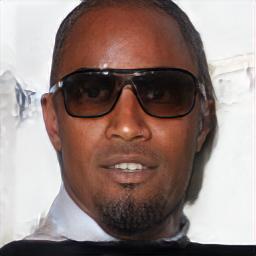}&
\interpfigt{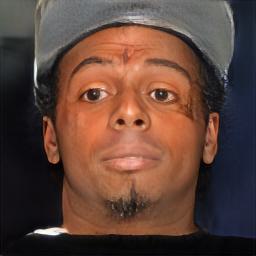}&
\interpfigt{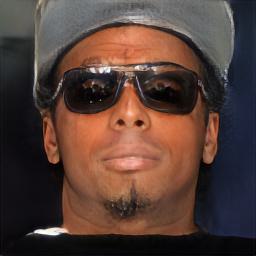}&
\interpfigt{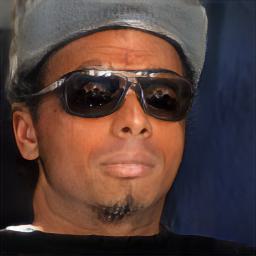}&
\interpfigt{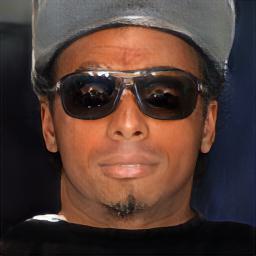}&
\interpfigt{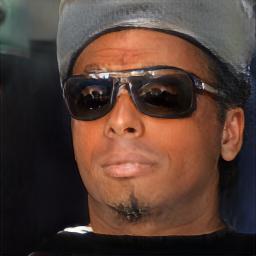}
\\

\interpfigt{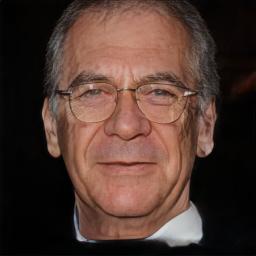}&
\interpfigt{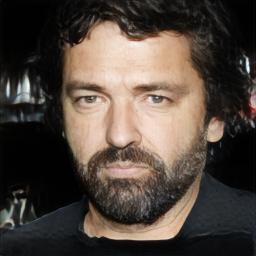}&
\interpfigt{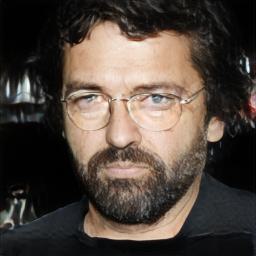}&
\interpfigt{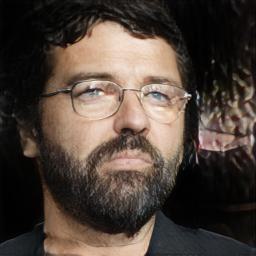}&
\interpfigt{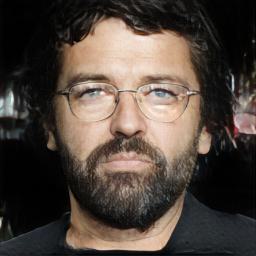}&
\interpfigt{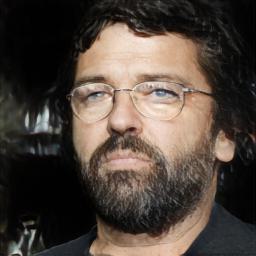}
\\

\interpfigt{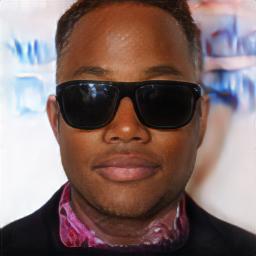}&
\interpfigt{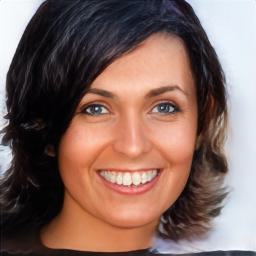}&
\interpfigt{Figures/glasses/ours/33/0033_rec_edited_origpose.jpg}&
\interpfigt{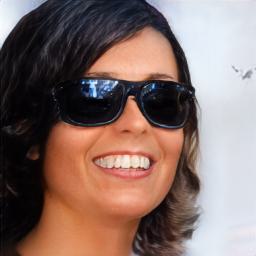}&
\interpfigt{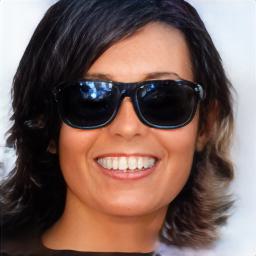}&
\interpfigt{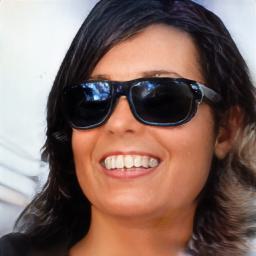}
\\

\interpfigt{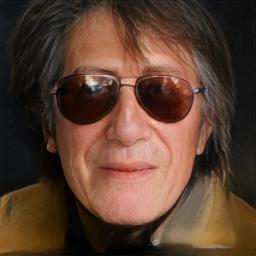}&
\interpfigt{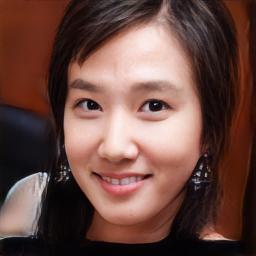}&
\interpfigt{Figures/glasses/ours/67/0067_rec_edited_origpose.jpg}&
\interpfigt{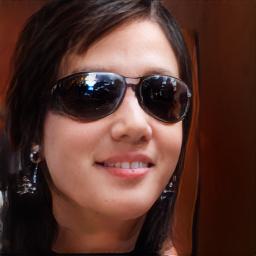}&
\interpfigt{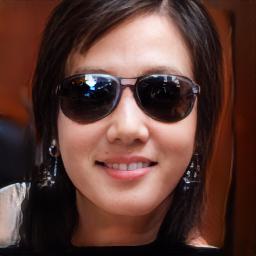}&
\interpfigt{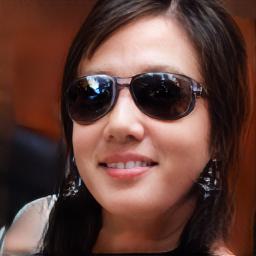}
\\

\interpfigt{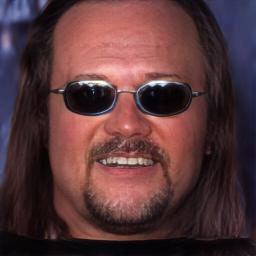}&
\interpfigt{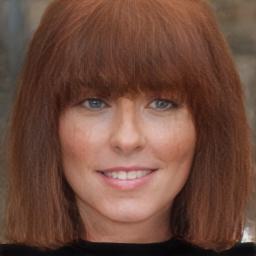}&
\interpfigt{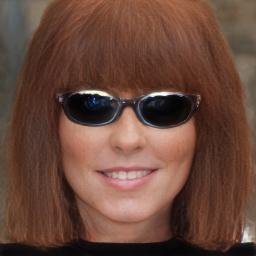}&
\interpfigt{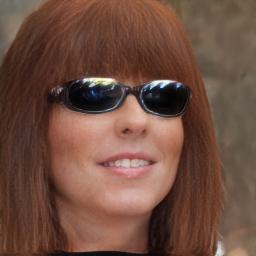}&
\interpfigt{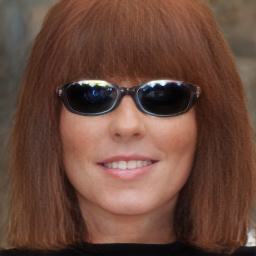}&
\interpfigt{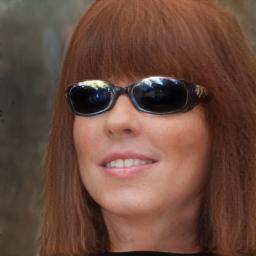}
\\

\interpfigt{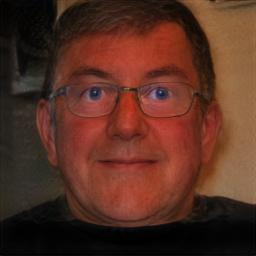}&
\interpfigt{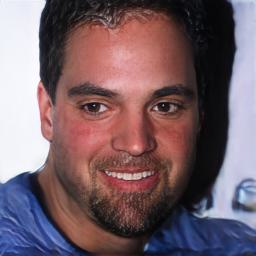}&
\interpfigt{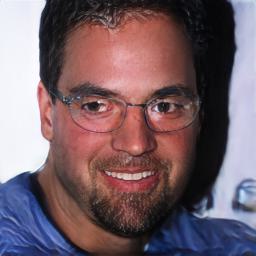}&
\interpfigt{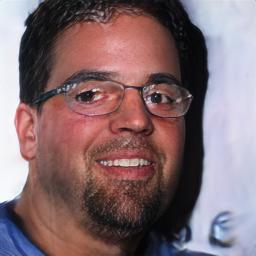}&
\interpfigt{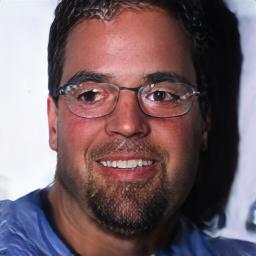}&
\interpfigt{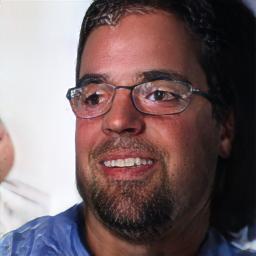}
\\

\interpfigt{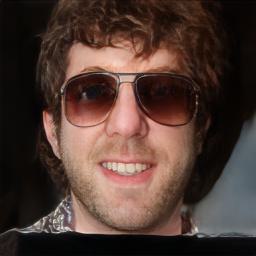}&
\interpfigt{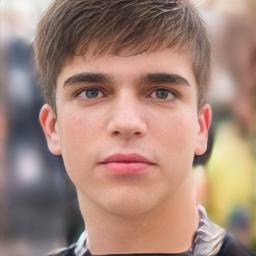}&
\interpfigt{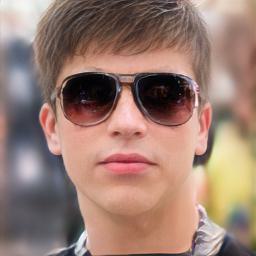}&
\interpfigt{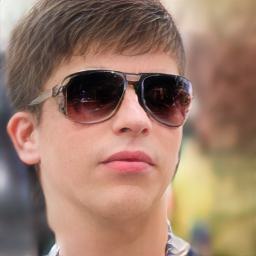}&
\interpfigt{Figures/glasses/ours/3/0003_rec_edited_1.jpg}&
\interpfigt{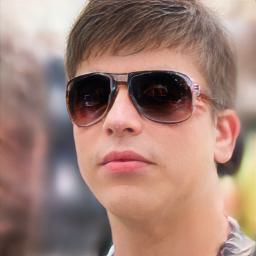}
\\
{Reference} & {Source} & \multicolumn{4}{c}{Multi-view Outputs}
\end{tabular}
}
\caption{Glasses edits and 3D visualizations.}
\label{fig:suppl_glasses_edits}
\end{figure*}

\begin{figure*}[ht!]
\centering
\setlength\tabcolsep{1pt}
\scalebox{1.0}{
\begin{tabular}{ccccccccc}
\interpfigt{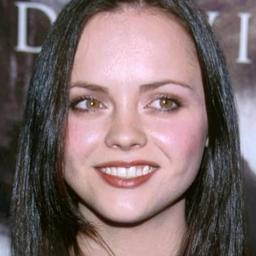}&
\interpfigt{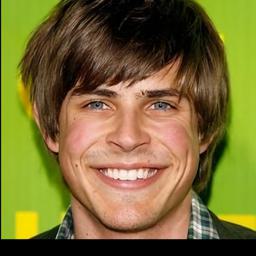}&
\interpfigt{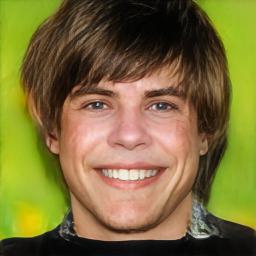}&
\interpfigt{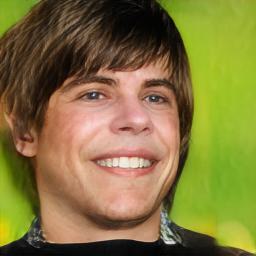}&
\interpfigt{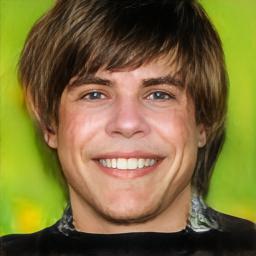}&
\interpfigt{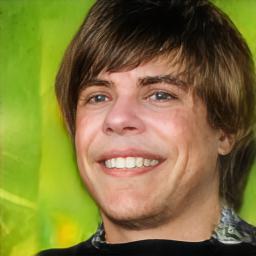}
\\
\interpfigt{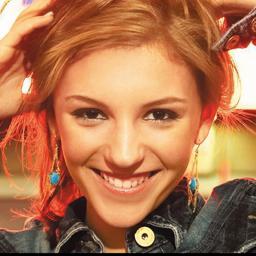}&
\interpfigt{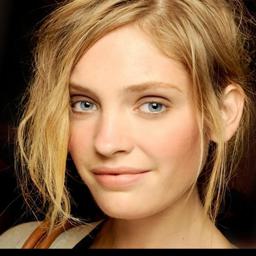}&
\interpfigt{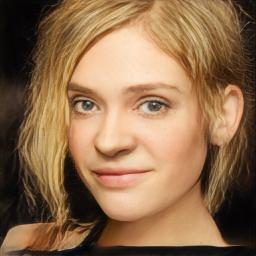}&
\interpfigt{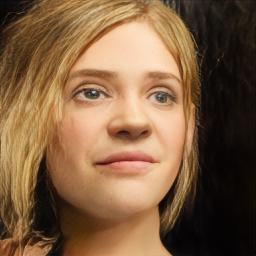}&
\interpfigt{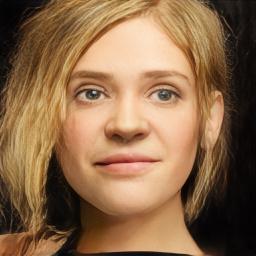}&
\interpfigt{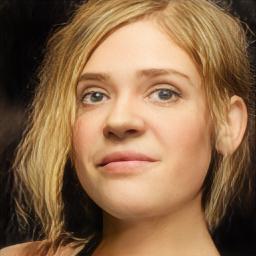}
\\
\interpfigt{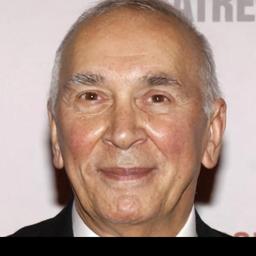}&
\interpfigt{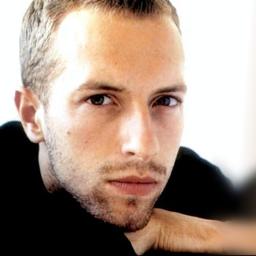}&
\interpfigt{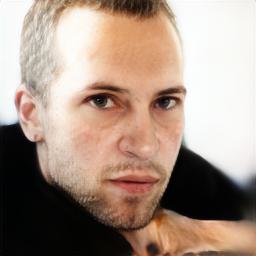}&
\interpfigt{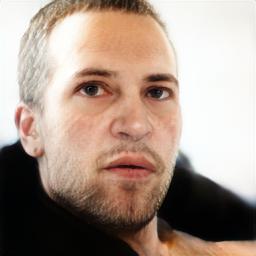}&
\interpfigt{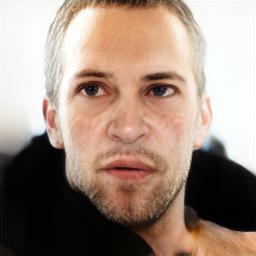}&
\interpfigt{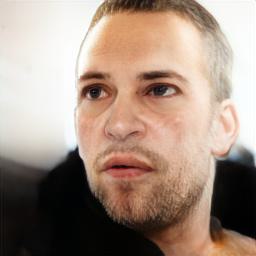}
\\
\interpfigt{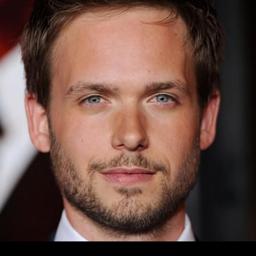}&
\interpfigt{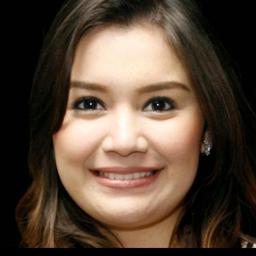}&
\interpfigt{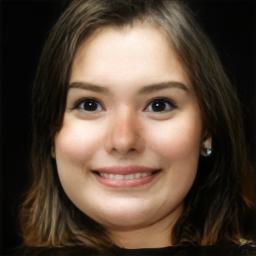}&
\interpfigt{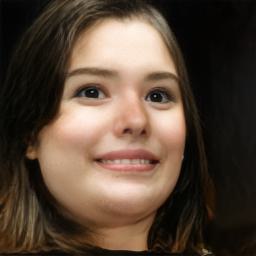}&
\interpfigt{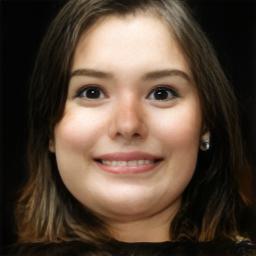}&
\interpfigt{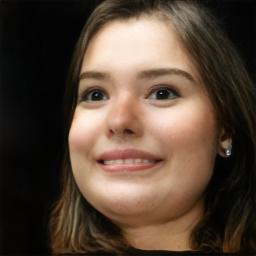}
\\
\interpfigt{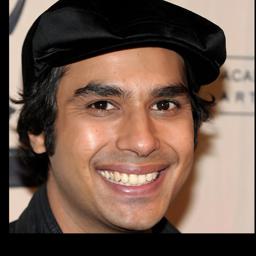}&
\interpfigt{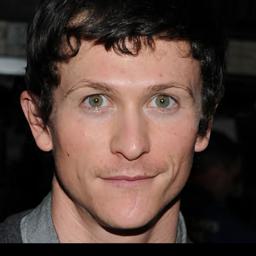}&
\interpfigt{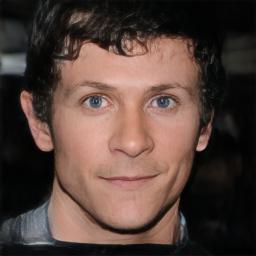}&
\interpfigt{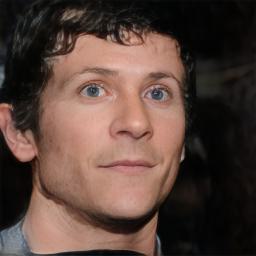}&
\interpfigt{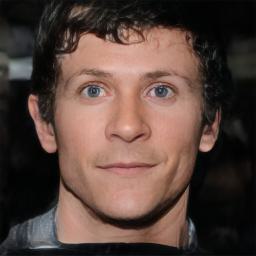}&
\interpfigt{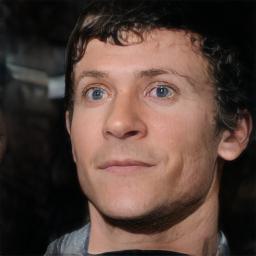}
\\
\interpfigt{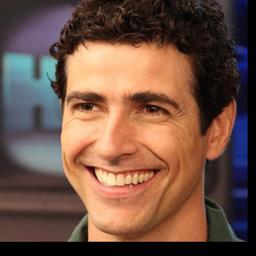}&
\interpfigt{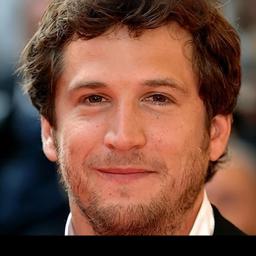}&
\interpfigt{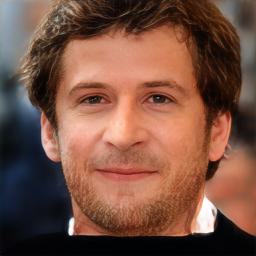}&
\interpfigt{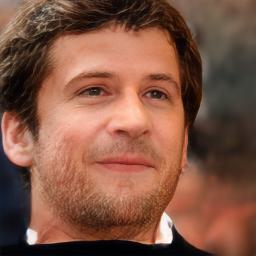}&
\interpfigt{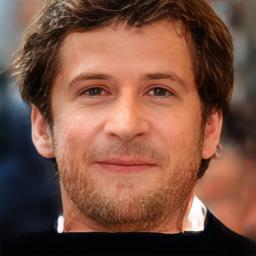}&
\interpfigt{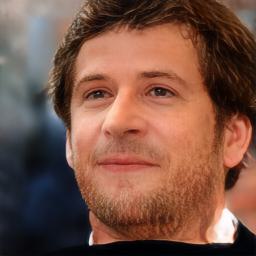}
\\
\interpfigt{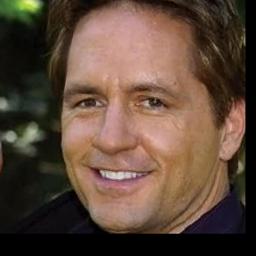}&
\interpfigt{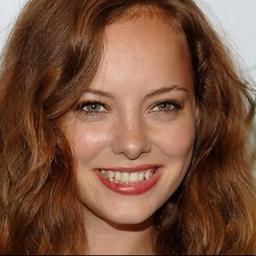}&
\interpfigt{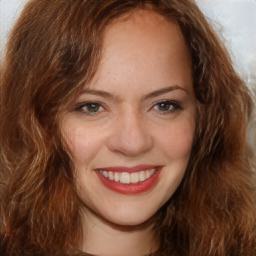}&
\interpfigt{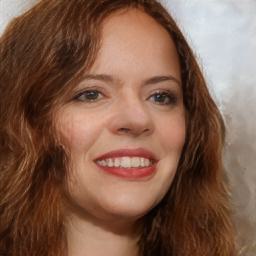}&
\interpfigt{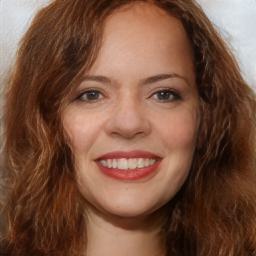}&
\interpfigt{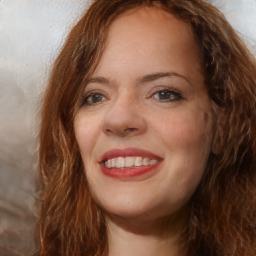}
\\
\interpfigt{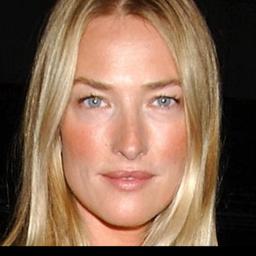}&
\interpfigt{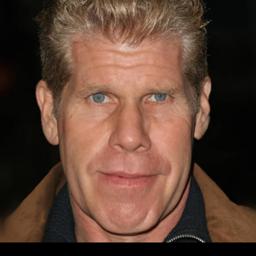}&
\interpfigt{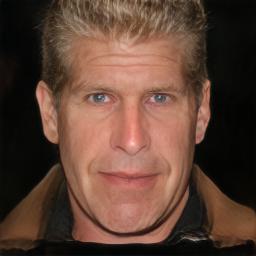}&
\interpfigt{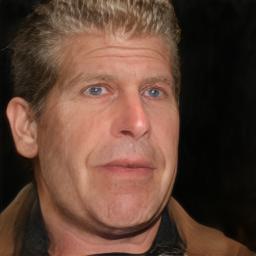}&
\interpfigt{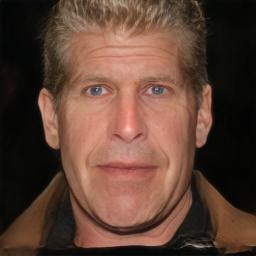}&
\interpfigt{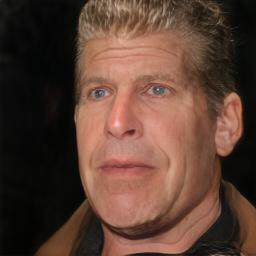}
\\
Reference & Source & \multicolumn{4}{c}{Multi-view Outputs}\\
\end{tabular}
}
\caption{Nose edits and 3D visualizations.}
\label{fig:suppl_nose_edits}
\end{figure*}

\begin{figure*}[ht!]
\centering
\setlength\tabcolsep{1pt}
\scalebox{1.0}{
\begin{tabular}{ccccccccc}
\interpfigt{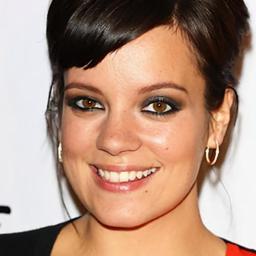}&
\interpfigt{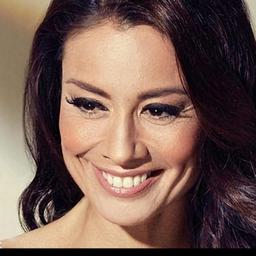}&
\interpfigt{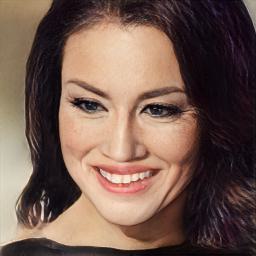}&
\interpfigt{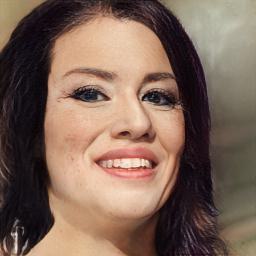}&
\interpfigt{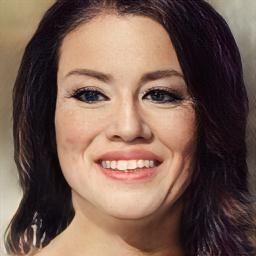}&
\interpfigt{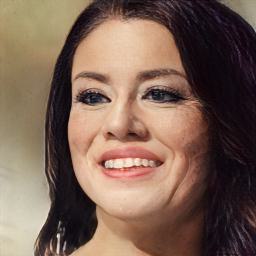}
\\
\interpfigt{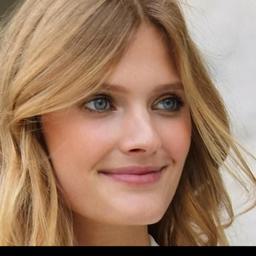}&
\interpfigt{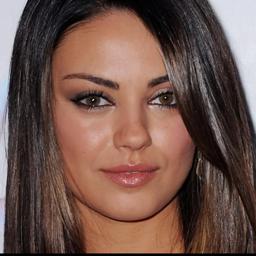}&
\interpfigt{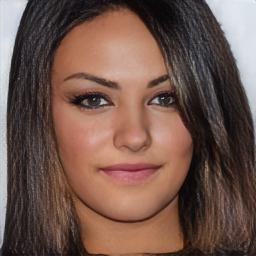}&
\interpfigt{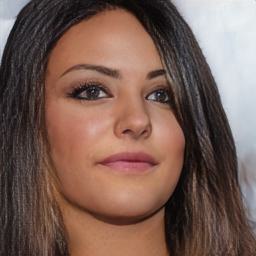}&
\interpfigt{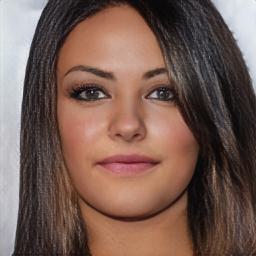}&
\interpfigt{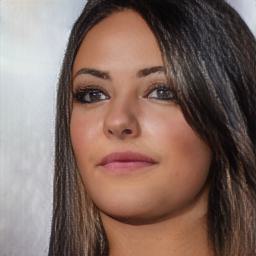}
\\
\interpfigt{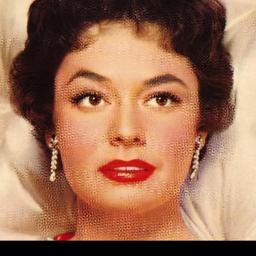}&
\interpfigt{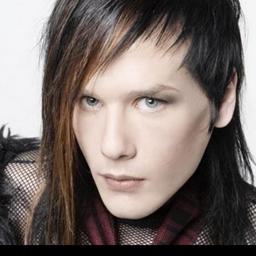}&
\interpfigt{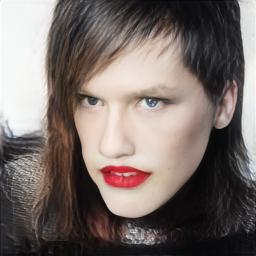}&
\interpfigt{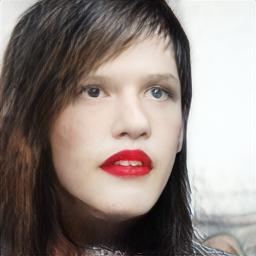}&
\interpfigt{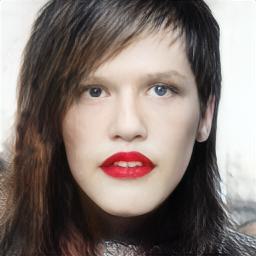}&
\interpfigt{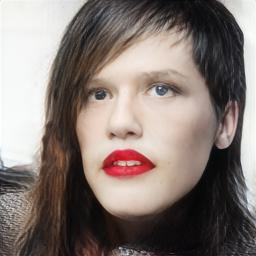}
\\
\interpfigt{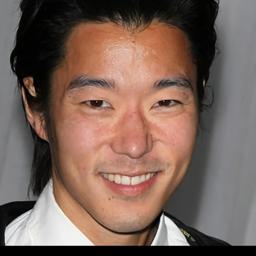}&
\interpfigt{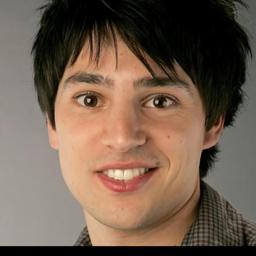}&
\interpfigt{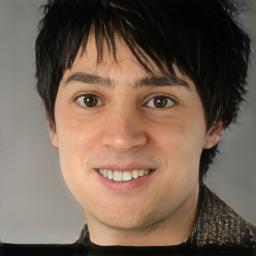}&
\interpfigt{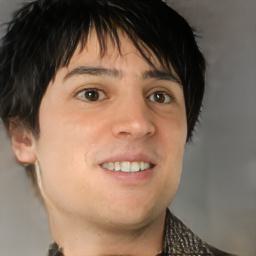}&
\interpfigt{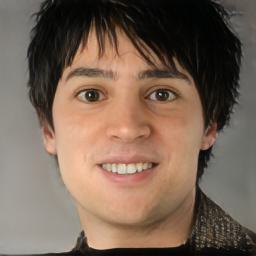}&
\interpfigt{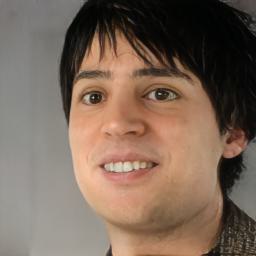}
\\
\interpfigt{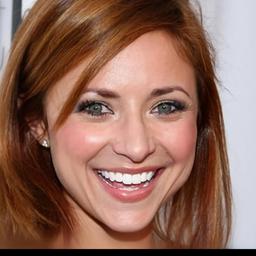}&
\interpfigt{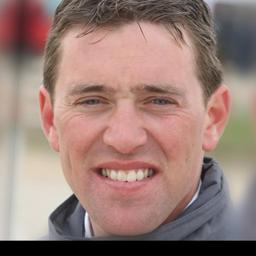}&
\interpfigt{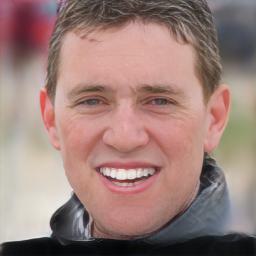}&
\interpfigt{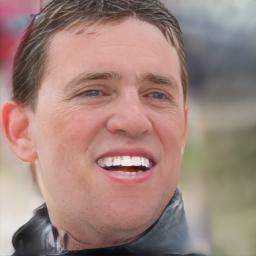}&
\interpfigt{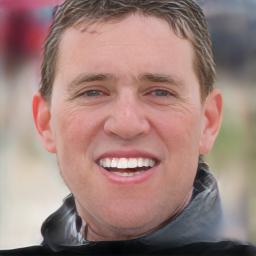}&
\interpfigt{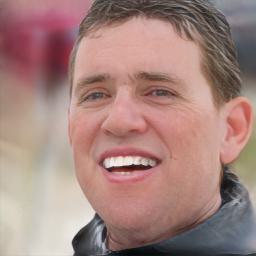}
\\
\interpfigt{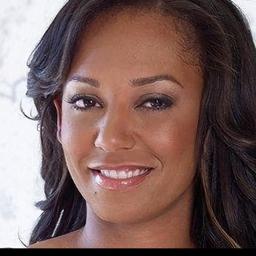}&
\interpfigt{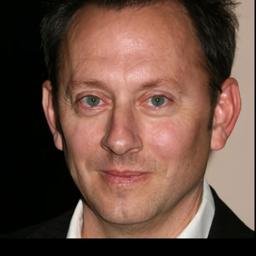}&
\interpfigt{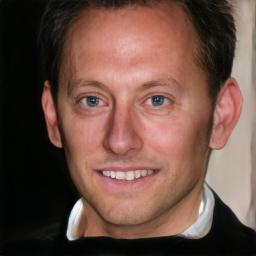}&
\interpfigt{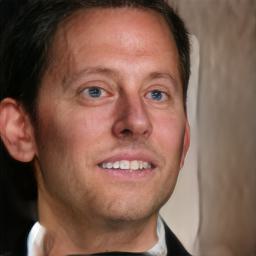}&
\interpfigt{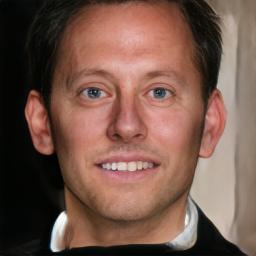}&
\interpfigt{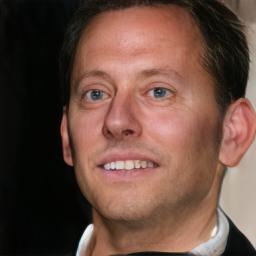}
\\
\interpfigt{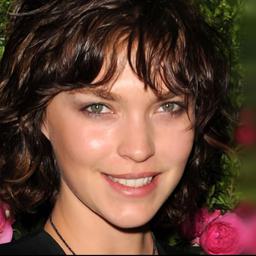}&
\interpfigt{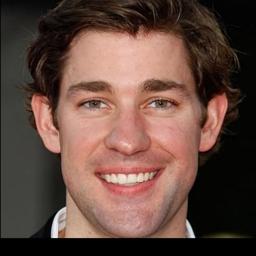}&
\interpfigt{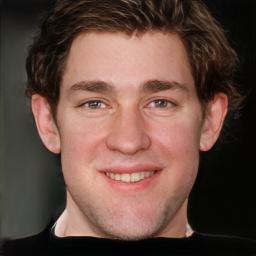}&
\interpfigt{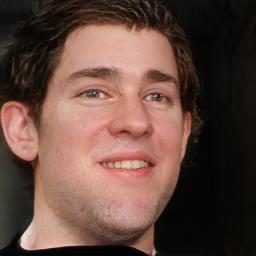}&
\interpfigt{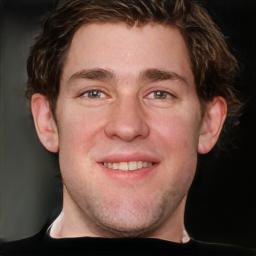}&
\interpfigt{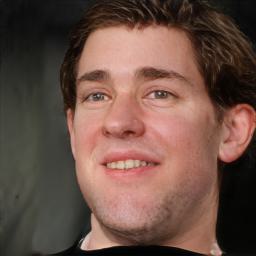}
\\
\interpfigt{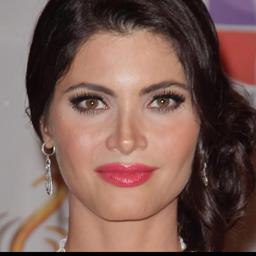}&
\interpfigt{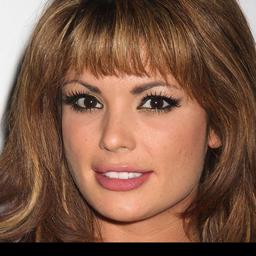}&
\interpfigt{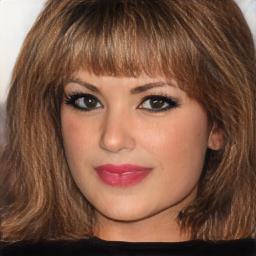}&
\interpfigt{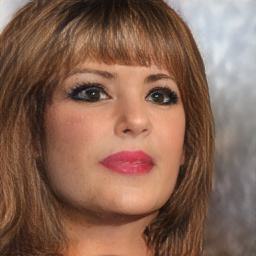}&
\interpfigt{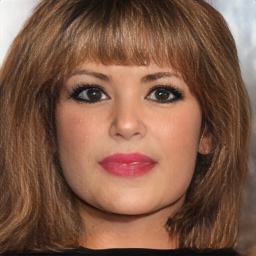}&
\interpfigt{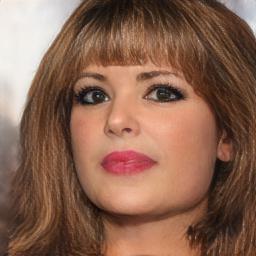}
\\
Reference & Source & \multicolumn{4}{c}{Multi-view Outputs}\\
\end{tabular}
}
\caption{Mouth edits and 3D visualizations.}
\label{fig:suppl_mouth_edits}
\end{figure*}

\begin{figure*}[ht!]
\centering
\setlength\tabcolsep{1pt}
\scalebox{1.0}{
\begin{tabular}{ccccccccc}
\interpfigt{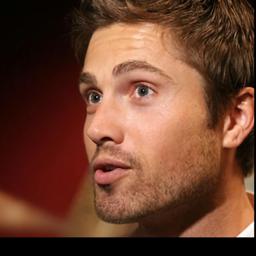}&
\interpfigt{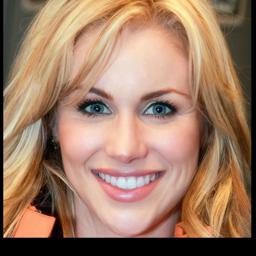}&
\interpfigt{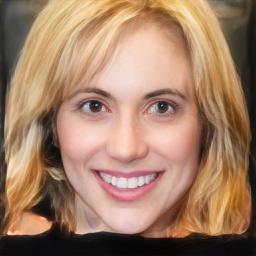}&
\interpfigt{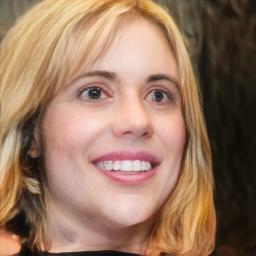}&
\interpfigt{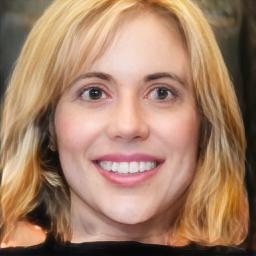}&
\interpfigt{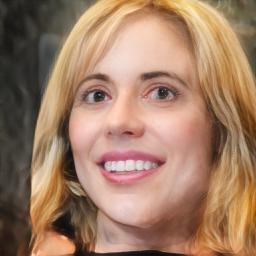}
\\
\interpfigt{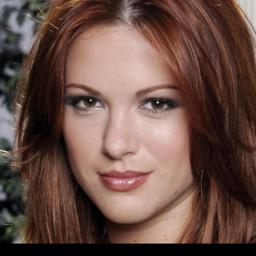}&
\interpfigt{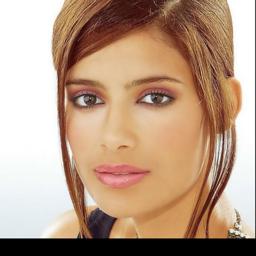}&
\interpfigt{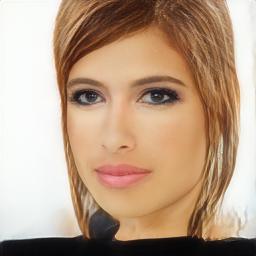}&
\interpfigt{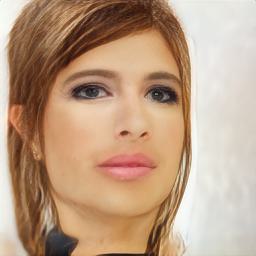}&
\interpfigt{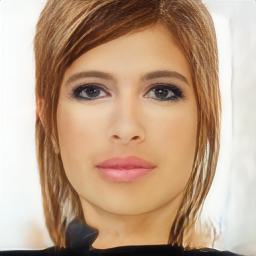}&
\interpfigt{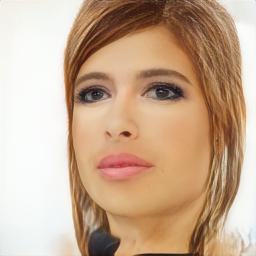}
\\
\interpfigt{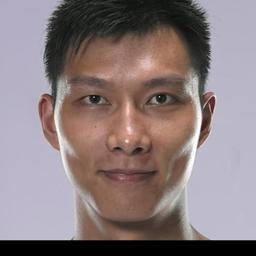}&
\interpfigt{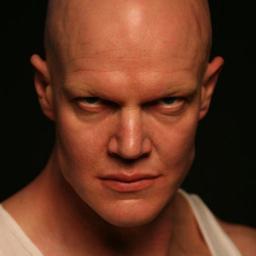}&
\interpfigt{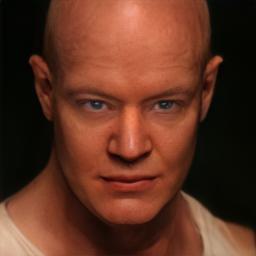}&
\interpfigt{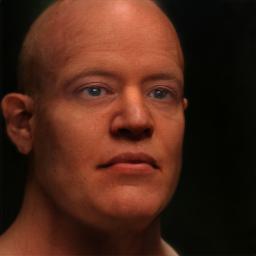}&
\interpfigt{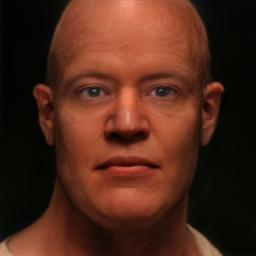}&
\interpfigt{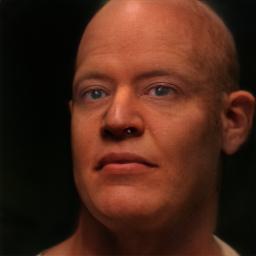}
\\
\interpfigt{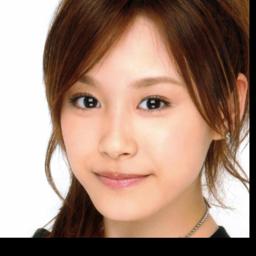}&
\interpfigt{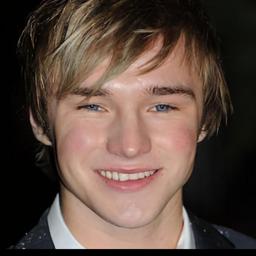}&
\interpfigt{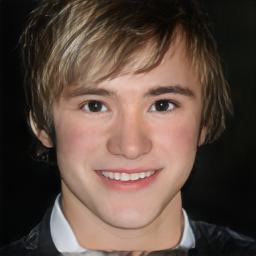}&
\interpfigt{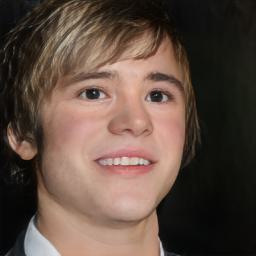}&
\interpfigt{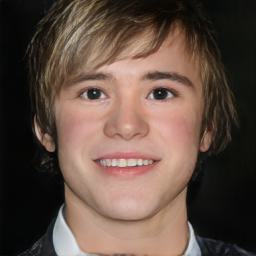}&
\interpfigt{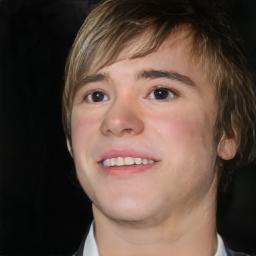}
\\
\interpfigt{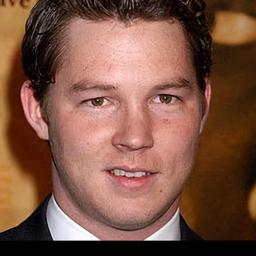}&
\interpfigt{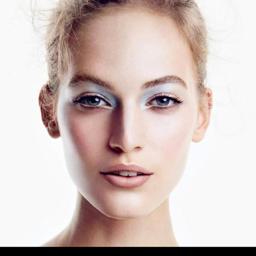}&
\interpfigt{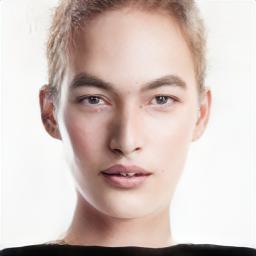}&
\interpfigt{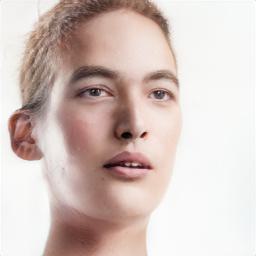}&
\interpfigt{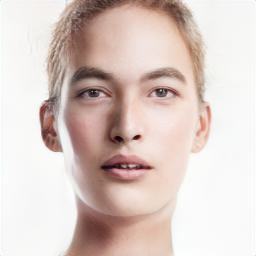}&
\interpfigt{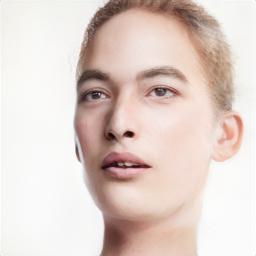}
\\
\interpfigt{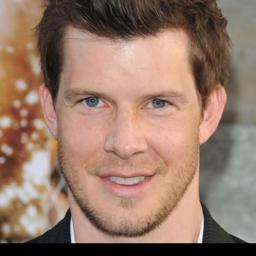}&
\interpfigt{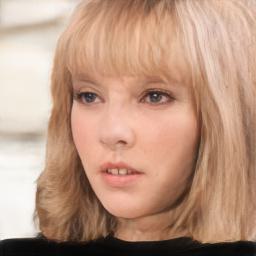}&
\interpfigt{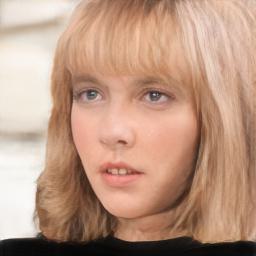}&
\interpfigt{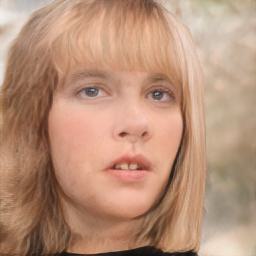}&
\interpfigt{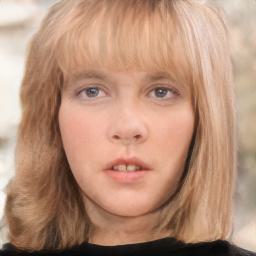}&
\interpfigt{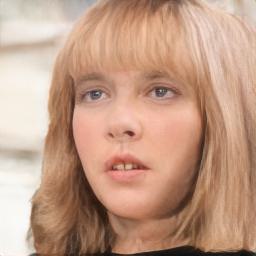}
\\
\interpfigt{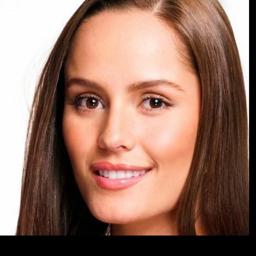}&
\interpfigt{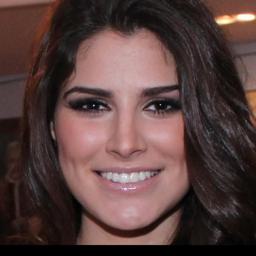}&
\interpfigt{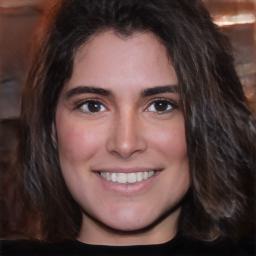}&
\interpfigt{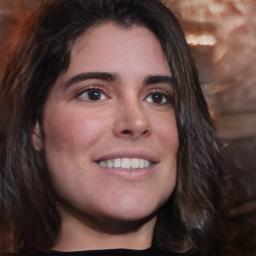}&
\interpfigt{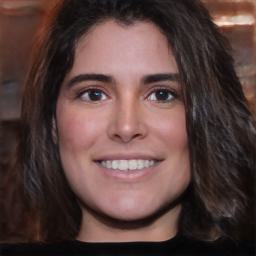}&
\interpfigt{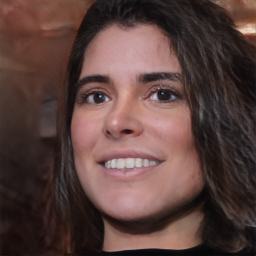}
\\
\interpfigt{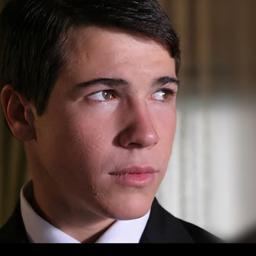}&
\interpfigt{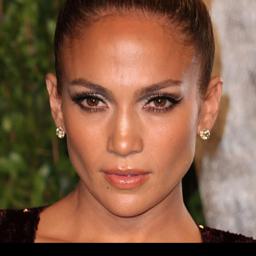}&
\interpfigt{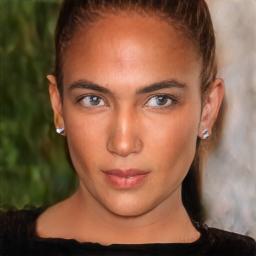}&
\interpfigt{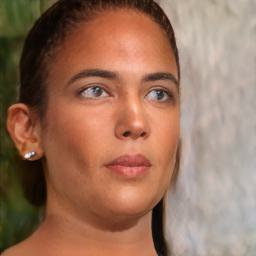}&
\interpfigt{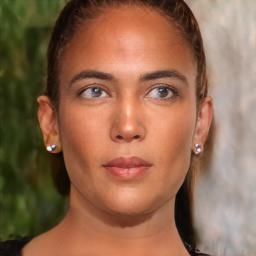}&
\interpfigt{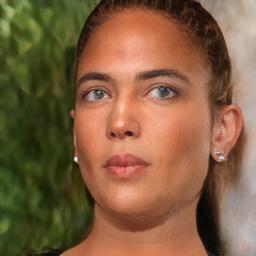}
\\
Reference & Source & \multicolumn{4}{c}{Multi-view Outputs}\\
\end{tabular}
}
\caption{Eyes edits and 3D visualizations.}
\label{fig:suppl_eyes_edits}
\end{figure*}

\begin{figure*}[ht!]
\centering
\setlength\tabcolsep{1pt}
\scalebox{1.0}{
\begin{tabular}{ccccccccc}
\interpfigt{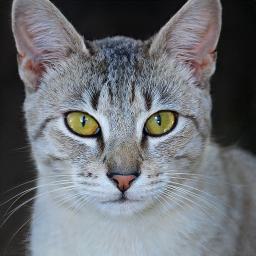}&
\interpfigt{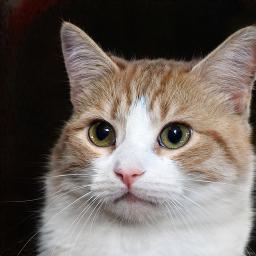}&
\interpfigt{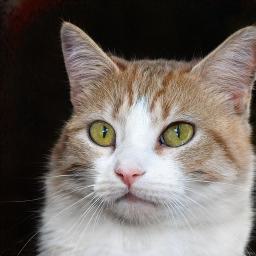}&
\interpfigt{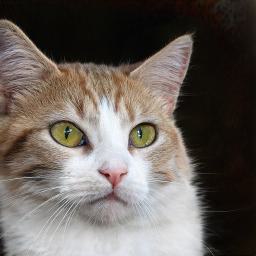}&
\interpfigt{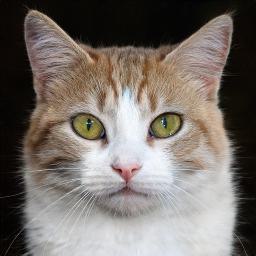}&
\interpfigt{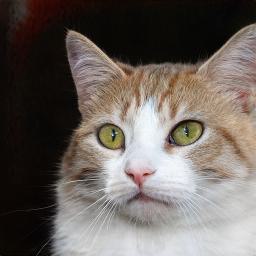}
\\
\interpfigt{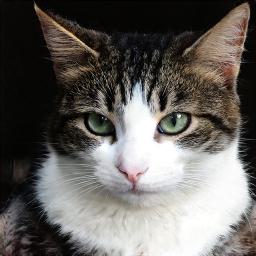}&
\interpfigt{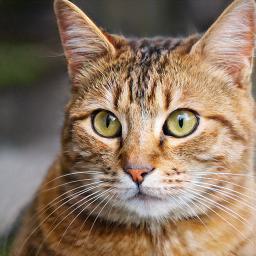}&
\interpfigt{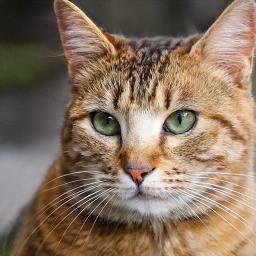}&
\interpfigt{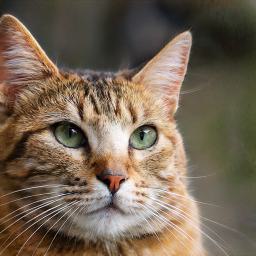}&
\interpfigt{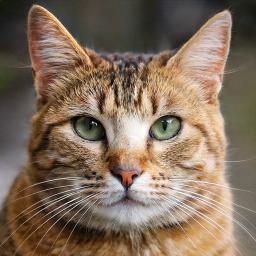}&
\interpfigt{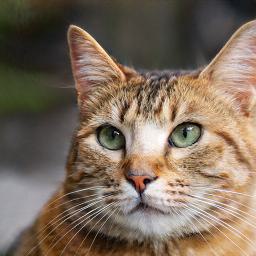}
\\
\interpfigt{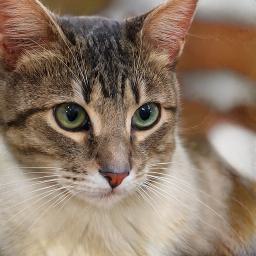}&
\interpfigt{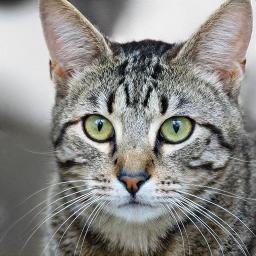}&
\interpfigt{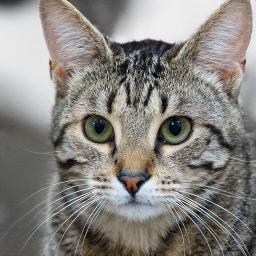}&
\interpfigt{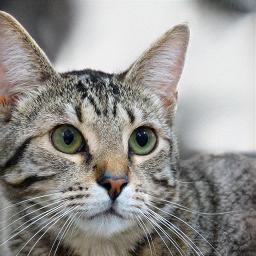}&
\interpfigt{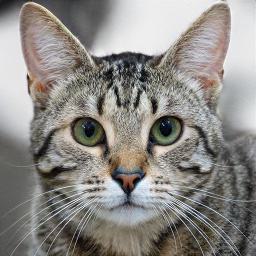}&
\interpfigt{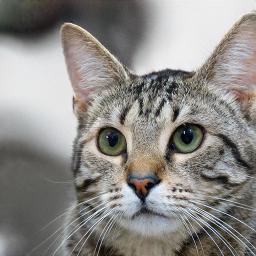}
\\
\interpfigt{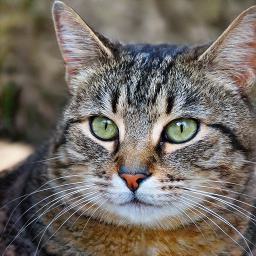}&
\interpfigt{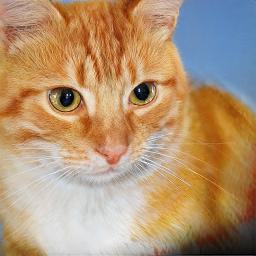}&
\interpfigt{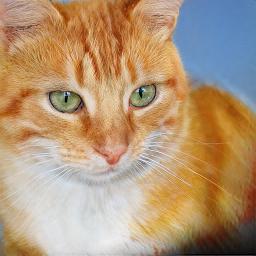}&
\interpfigt{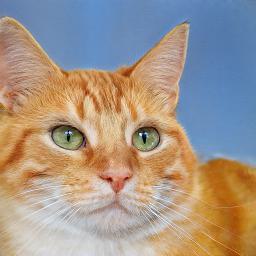}&
\interpfigt{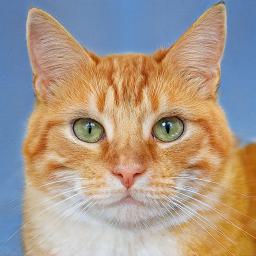}&
\interpfigt{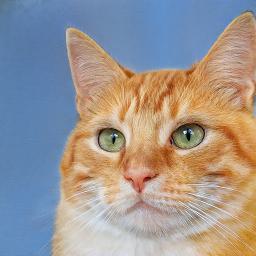}
\\
\interpfigt{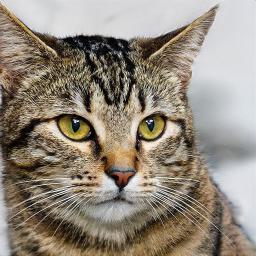}&
\interpfigt{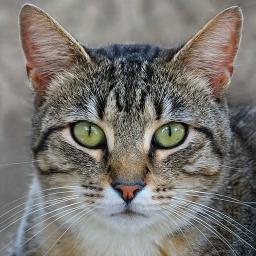}&
\interpfigt{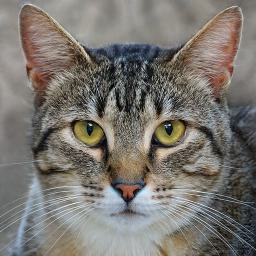}&
\interpfigt{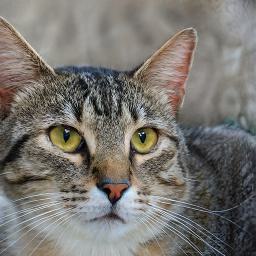}&
\interpfigt{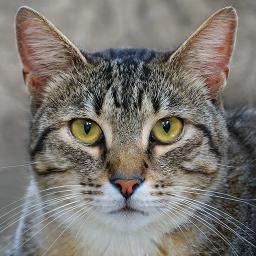}&
\interpfigt{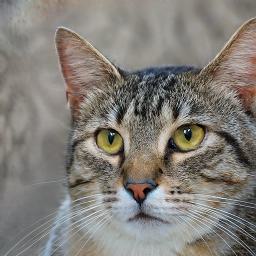}
\\

Reference & Source & \multicolumn{4}{c}{Multi-view Outputs}\\
\end{tabular}
}
\caption{AFHQ eyes edits and 3D visualizations.}
\label{fig:suppl_afhq_eyes_edits}
\end{figure*}

\begin{figure*}[ht!]
\centering
\setlength\tabcolsep{1pt}
\scalebox{1.0}{
\begin{tabular}{ccccccccc}
\interpfigt{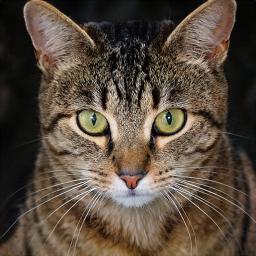}&
\interpfigt{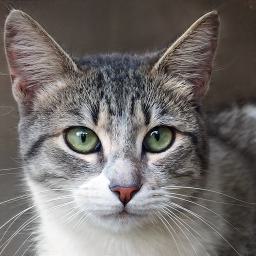}&
\interpfigt{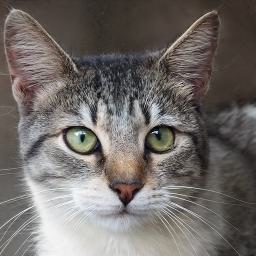}&
\interpfigt{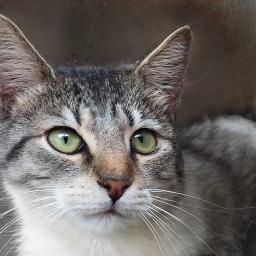}&
\interpfigt{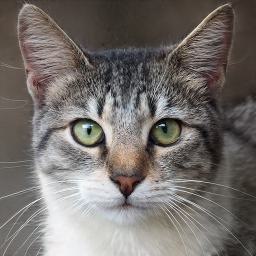}&
\interpfigt{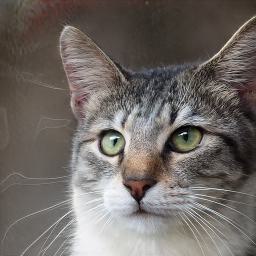}
\\
\interpfigt{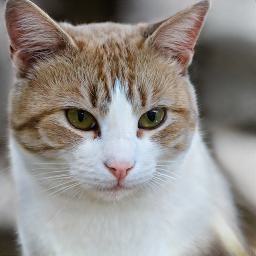}&
\interpfigt{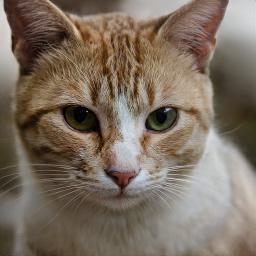}&
\interpfigt{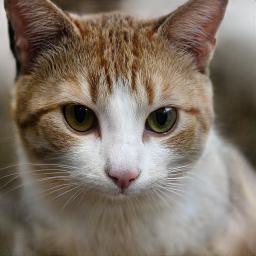}&
\interpfigt{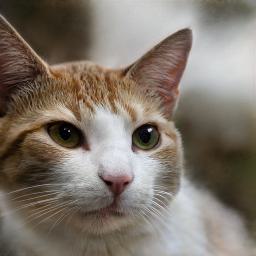}&
\interpfigt{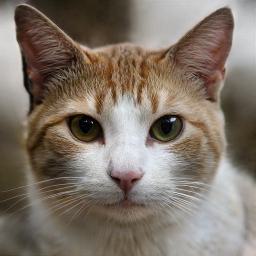}&
\interpfigt{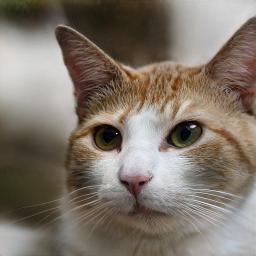}
\\
\interpfigt{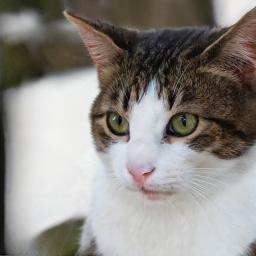}&
\interpfigt{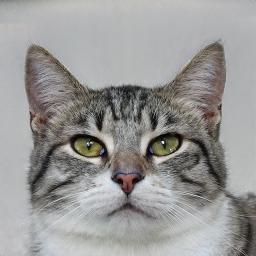}&
\interpfigt{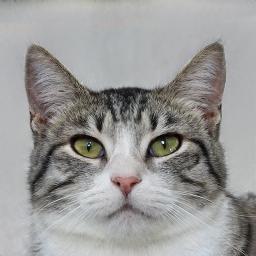}&
\interpfigt{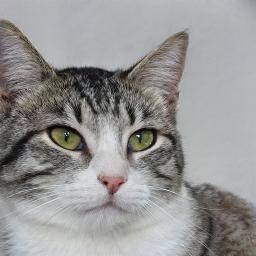}&
\interpfigt{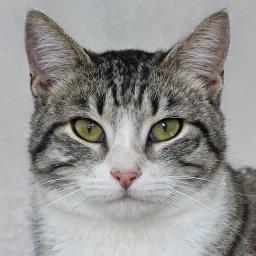}&
\interpfigt{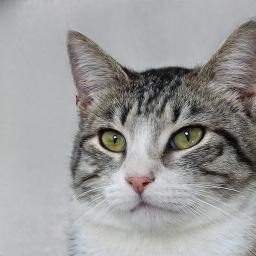}
\\
\interpfigt{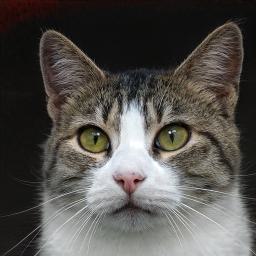}&
\interpfigt{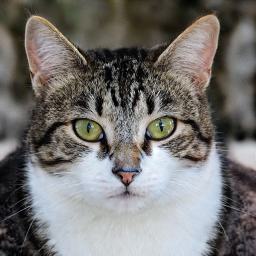}&
\interpfigt{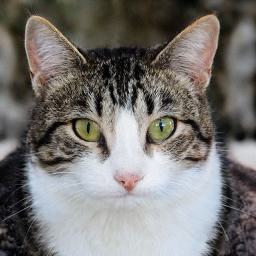}&
\interpfigt{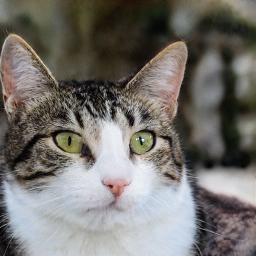}&
\interpfigt{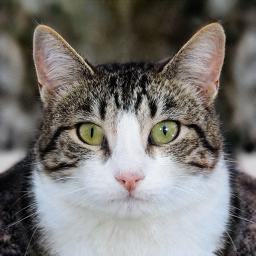}&
\interpfigt{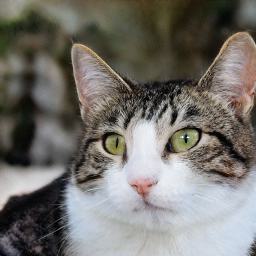}
\\
\interpfigt{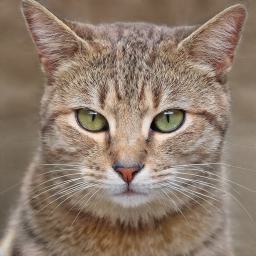}&
\interpfigt{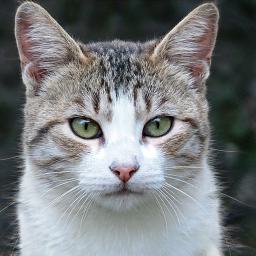}&
\interpfigt{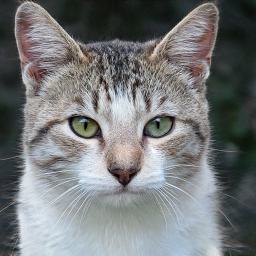}&
\interpfigt{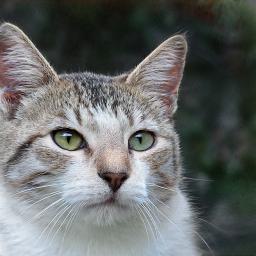}&
\interpfigt{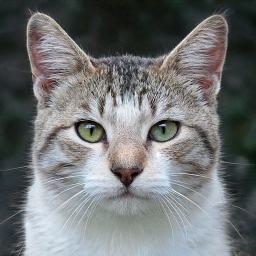}&
\interpfigt{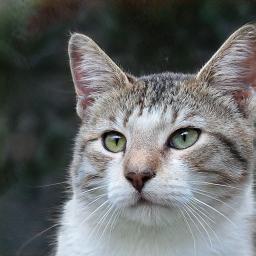}
\\
Reference & Source & \multicolumn{4}{c}{Multi-view Outputs}\\
\end{tabular}
}
\caption{AFHQ mouth \& nose edits and 3D visualizations.}
\label{fig:suppl_afhq_nose_edits}
\end{figure*}

We showcase additional reference-based editing results in~\cref{fig:suppl_hair_edits,fig:suppl_glasses_edits,fig:suppl_mouth_edits,fig:suppl_nose_edits,fig:suppl_eyes_edits} for glasses, hairstyle, mouth, nose, and eyes on CelebA, and~\cref{fig:suppl_afhq_eyes_edits,fig:suppl_afhq_nose_edits} for eyes and nose on AFHQ dataset. Reference and input images are encoded and reconstructed via EG3D-GOAE for human portraits, and synthesised for animal faces.

\section{User study}
\begin{figure*}
\centering
\includegraphics[width=0.9\linewidth]{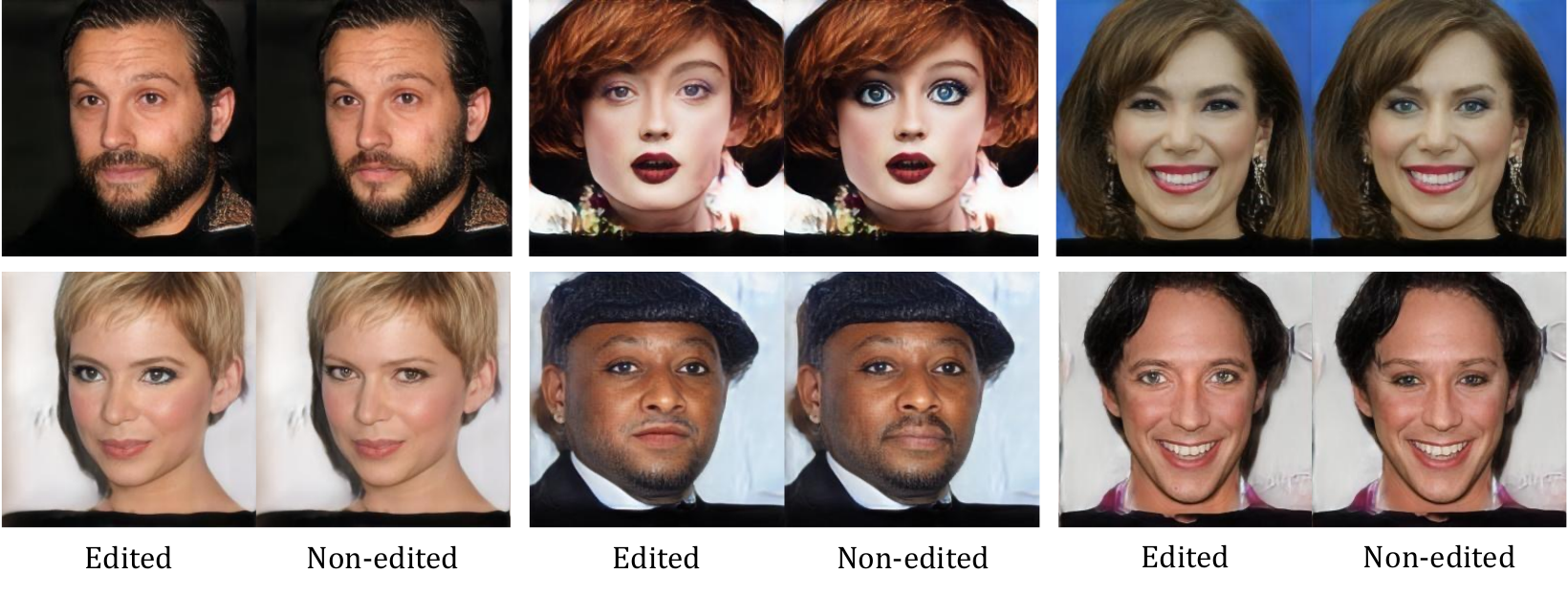}
\caption{Samples for top 3 most incorrect (first row) and correct (second row) responses on the user study. Incorrect responses correspond to the user not being able to distinguish between edited and non-edited samples, and vice versa for the correct responses.}
\label{fig:supp_user_study}
\end{figure*}

We provide additional samples for the top 3 incorrect and correct responses in~\cref{fig:supp_user_study}. For the correct responses, even though the users were able to distinguish between edited and non-edited samples, the editing is still faithful to the source images as there are no visible artifacts.


\newcommand{\interpfigu}[1]{\includegraphics[trim=0 0 0cm 0, clip, width=7cm]{#1}}

\end{document}